\RequirePackage{fix-cm}
\documentclass[twocolumn]{svjour3}          % twocolumn
\smartqed  % flush right qed marks, e.g. at end of proof

\usepackage{graphics}
\usepackage{amsmath}
\usepackage{amssymb}
\usepackage{algorithm}
\usepackage{algpseudocode}
\usepackage{subcaption}
\usepackage{paralist}
\usepackage{color,soul}
\usepackage{multirow}
\usepackage{xfrac}
\usepackage{hyperref}
\usepackage{cite}

\begin{document}

\title{2D Map Alignment With Region Decomposition%\thanks{Grants or other notes}
}
%% \subtitle{Do you have a subtitle?\\ If so, write it here}
%% \titlerunning{Short form of title}        % if too long for running head

\author{Saeed Gholami Shahbandi \and Martin Magnusson}
%\authorrunning{S. Gholami Shahbandi \and M. Magnusson} % if too long for running head

\institute{
  Saeed Gholami Shahbandi \at
  Center for Applied Intelligent Systems Research,\\
  Intelligent Systems Lab, Halmstad University, Sweden\\
  \email{saesha@hh.se}           %  \\
  \and
  Martin Magnusson \at
  Mobile Robotics and Olfaction Lab,\\
  Centre for Applied Autonomous Sensor Systems (AASS),\\
  \"{O}rebro University, Sweden\\
  \email{martin.magnusson@oru.se}           %  \\
}

\date{} %% \date{Received: 19 June 2017 / Accepted: 18 April 2018}

\maketitle
%% \thispagestyle{empty}
%% \pagestyle{empty}

%%%%%%%%%%%%%%%%%%%%%%%%%%%%%%%%%%%%%%%%%%%%%%%%%%%%%%%%%%%%%%%%%%%%%%%%%%%%%%%%
%%%%%%%%%%%%%%%%%%%%%%%%%%%%%%%%%%%%%%%%%%%%%%%%%%%%%%%%%%%%%%%%%%%%%%%%%%%%%%%%
%%%%%%%%%%%%%%%%%%%%%%%%%%%%%%%%%%%%%%%%%%%%%%%%%%%%%%%%%%%%%%%%%%%%%%%%%%%%%%%%
\begin{abstract}
In many applications of autonomous mobile robots the following problem is encountered.
Two maps of the same environment are available, one a prior map and the other a sensor map built by the robot.
To benefit from all available information in both maps, the robot must find the correct alignment between the two maps.
There exist many approaches to address this challenge, however, most of the previous methods rely on assumptions such as similar modalities of the maps, same scale, or existence of an initial guess for the alignment.
In this work we propose a decomposition-based method for 2D spatial map alignment which does not rely on those assumptions.
Our proposed method is validated and compared with other approaches, including generic data association approaches and map alignment algorithms.
Real world examples of four different environments with thirty six sensor maps and four layout maps are used for this analysis.
The maps, along with an implementation of the method, are made publicly available online.
\end{abstract}

%%%%%%%%%%%%%%%%%%%%%%%%%%%%%%%%%%%%%%%%%%%%%%%%%%%%%%%%%%%%%%%%%%%%%%%%%%%%%%%%
%%%%%%%%%%%%%%%%%%%%%%%%%%%%%%%%%%%%%%%%%%%%%%%%%%%%%%%%%%%%%%%%%%%%%%%%%%%%%%%%
%%%%%%%%%%%%%%%%%%%%%%%%%%%%%%%%%%%%%%%%%%%%%%%%%%%%%%%%%%%%%%%%%%%%%%%%%%%%%%%%
\section{Introduction} \label{sec:introduction}
There are many applications in which it is beneficial for a robot to merge its map with any of a number of existing maps.
For example, in environment surveying, a blueprint layout map could be introduced to give the robot a head start in terms of exploration.
Such a prior map could also improve global consistency of SLAM algorithms by exploiting the global consistency of the prior map.
Another example is to integrate semantic information or traffic flow data into a central map that a single agent could not obtain alone.
Furthermore, a hybrid map constructed from merging maps of different modalities, enables access to all included modalities through each individual map.
For instance assume that the semantic labels are provided by one map and the robot can localize itself using another sensor modality.
The robot can become aware of each region's semantic label merely by localizing itself in one map and exploiting the association between maps.
All the aforementioned applications share the need for a map alignment procedure.
Solving the \emph{autonomous} map alignment problem has interesting upshots.
A seamless map alignment procedure improves the autonomy of robotic services by reducing the demand for human intervention.

A scenario in which the map alignment is of particular interest is where a robot is expected to employ a prior map of the environment in addition to its own capacity to create maps.
In this example, an important challenge is the difference in map formats.
In such cases the prior map of the environment is a blueprint, and the robot maps are often discrete such as Occupancy Grid Maps.
\emph{This paper addresses 2D map alignment where the maps share no frame of reference, overlap only partially, have different amounts of clutter, and have different modalities.}

\begin{figure}
  \centering
    \includegraphics[width=\linewidth]{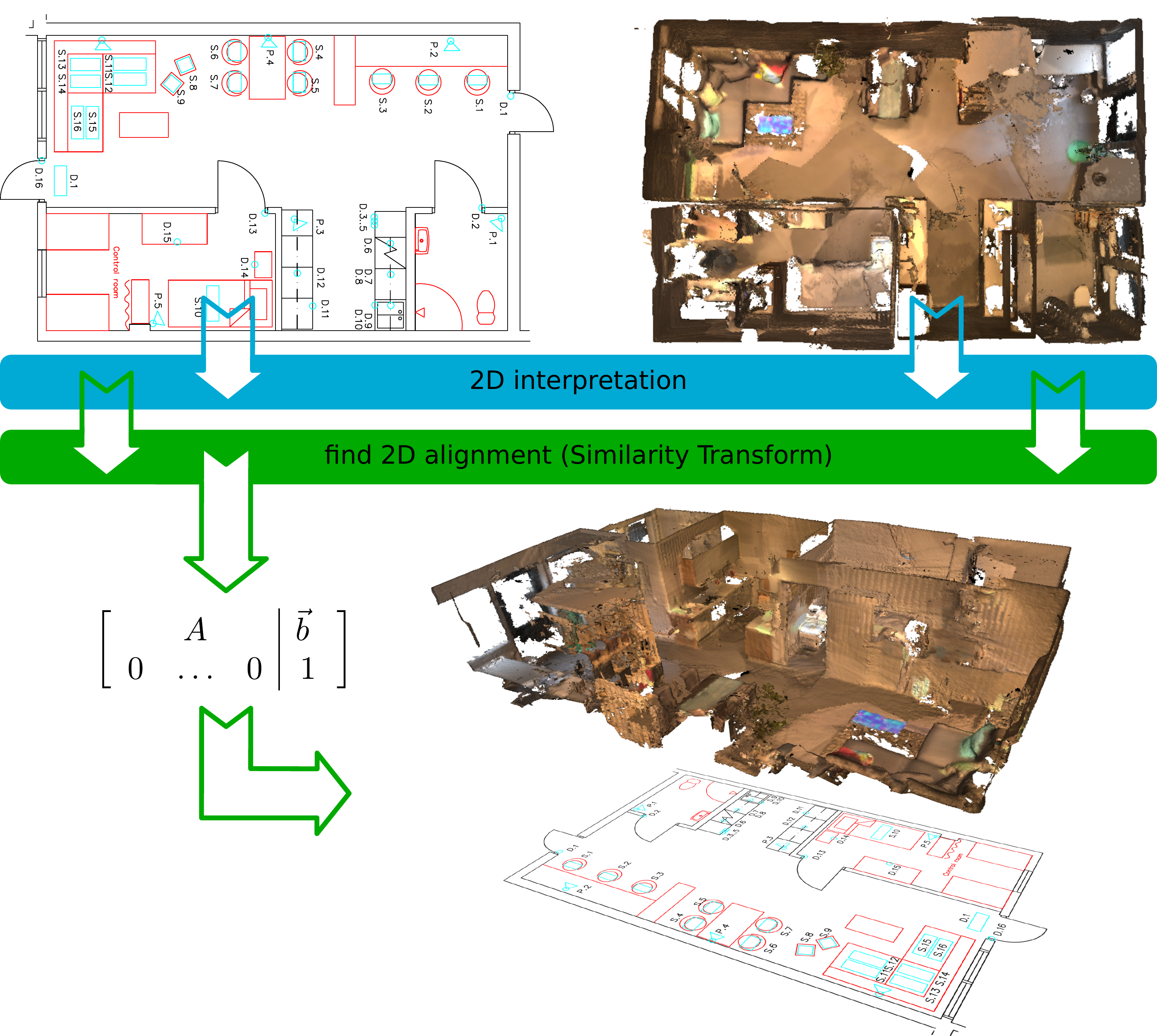}
    \caption[xxx]{
      An example of a pair of layout and sensor maps.
      Sensor maps are acquired with a Google Tango tablet as 3D meshes, and later are converted to 2D occupancy-like maps.
      This example is from the Halmstad Intelligent Home~\cite{lundstrom2016halmstad}.
      Other maps also include office environments (see Appendix~\ref{app:datasets}).
    }\label{fig:HIH_example}
\end{figure}

%%%%%%%%%%%%%%%%%%%%%%%%%%%%%%%%%%%%%%%%%%%%%%%%%%%%%%%%%%%%%%%%%%%%%%%%%%%%%%%%
\subsection{Problem description} \label{subsec:problem}

We define the map alignment as a data association problem across two representations of the same environment (two maps).
The solution to this problem is a transformation function between the coordinate frames of the input maps.
The objective is then to find the optimal transformation under which the distances between corresponding elements in the two representations are minimized.
The challenges of map alignment we intend to address in this work are:
\begin{itemize}
\item multi-modality and representation discrepancy between the maps (e.g. blueprint vs. sensor map),
\item scale mismatch between input maps (e.g. due to different modalities),
\item repeating patterns and the associated problem of local minima in the alignment objective function (auto-isomorphic graphs in a topological sense, and shape correspondence problem in a geometric sense).
\end{itemize}

By the following set of assumptions we contain the problem description to a more specific application domain:

\begin{itemize}
\item the environments are:
  \begin{itemize}
  \item well-structured, that is to say their maps could be modeled (abstracted) with a set of 2D geometric curves,
  \item composed of meaningful regions that can be segmented (e.g. room, corridor, etc.).
  \end{itemize}
\item the maps are:
  \begin{itemize}
  \item spatial, geometric and 2-dimensional (or could be represented in a 2D plane),
  \item globally consistent (not ``broken'') and uniformly scaled.
  \end{itemize}
\end{itemize}

The restrictions and limitations caused by these assumptions are further discussed in Section~\ref{sec:conclusion}.

%%%%%%%%%%%%%%%%%%%%%%%%%%%%%%%%%%%%%%%%%%%%%%%%%%%%%%%%%%%%%%%%%%%%%%%%%%%%%%%%
\subsection{Approaches} \label{subsec:approaches}

In this general formulation, the map alignment problem proves to be more challenging than the relaxed versions such as scan matching or image alignment problems.
When the displacement between two frames (maps or scans) are small, optimization based algorithms such as point set registration~\cite{besl1992method}, \cite{Gold19981019}, \cite{Tsin2004}, \cite{NIPS2006_2962}, \cite{5432191} or image registration~\cite{Baker2004} are suitable solutions.
However, these algorithms are vulnerable to local minima, especially in the absence of an initial guess.
This pitfall is exacerbated when the input maps contain repetitive patterns that increase the number of local minima.

Another approach is to employ the Hough transform to structure the search space and decompose the transformation into separate operations of rotation and translation~\cite{carpin2008fast}, \cite{saeedi2012efficient}.
However, these approaches are limited to rigid transformations (i.e. Euclidean transformation that includes translation, rotation and no scaling) and expect homogeneity of the input signal (same modality).

A more common approach to the map alignment problem is to interpret the input maps with an abstract representation that enables a search on the similarity of instances.
For example, graphs capture the canonical points of the open space as vertices, and the connectivity of the open space is represented by the edges between vertices~\cite{huang2005topological}, \cite{schwertfeger2013evaluation}, \cite{kakuma2017alignment}.
Consequently, geometric and/or topological similarities of the vertices and/or edges are used to find a match between two maps.
When maps are of different types, such an interpretation plays an important role.
This interpretation abstracts the input maps into a \emph{shared instantiated representation}, which makes the search for similarities between maps feasible.

A more thorough review of related work is presented in Section~\ref{sec:related}, after which detailing our method and the experimental validation of it are presented in sections~\ref{sec:method} and \ref{sec:results}.

%%%%%%%%%%%%%%%%%%%%%%%%%%%%%%%%%%%%%%%%%%%%%%%%%%%%%%%%%%%%%%%%%%%%%%%%%%%%%%%%
\subsection{Our approach} \label{subsec:our_approach}
In this work, we propose a method based on the aforementioned concept of \emph{shared instantiated representation}.
The underlying representation of our method is a geometric decomposition that outlines the segmentation of regions, namely ``arrangement'' \cite{agarwal2000arrangements} in 2D.
When modeling the occupied regions of the maps (corresponding to the physical elements in the environment), the 2D arrangement explicitly represents both the \emph{boundaries} and the \emph{regions} of the open-spaces.
In addition, it implicitly captures the connectivity of the open-spaces through the regions' adjacency.
Our proposal is to 
\begin{inparaenum}[i)]
\item abstract input maps via region decomposition into a shared geometric representation, i.e. 2D arrangement,
\item search for all potential alignments that match the regions of the maps, %candidates suggested by matching regions of the two maps, and 
\item select the best alignment according to a proposed quality measure.
\end{inparaenum}
The details of the method, the 2D arrangement, and further details are provided in Section~\ref{sec:method}.
The differentiating characteristic of our method is its representation and the consequent search approach.
Most approaches generate hypotheses from some initial cues and follow along the progress of those cues.
In contrast, we exhaust the search space to avoid missing the correct answer.
This is crucial in the case of noisy maps and maps of different modalities.
To our knowledge, no algorithm has been developed for solving the alignment problem in this manner.
The main contributions of this work are as follows:
\begin{itemize}
\item
  An algorithm for map alignment that does not rely on rigid transformations, available initial guess or similar representation between maps.

\item
  An abstract representation (2D arrangement), which is capable of interpreting maps of different modalities.
  We use this interpretation for region segmentation, and for solving the alignment problem.
  This interpretation results in a hierarchical representation of maps, where the abstract models on the top level are readily available for other geometric processing and manipulations after alignment.

\item
  A publicly available collection of maps, containing forty maps of four different environments.
  The source code to our implementation of the proposed method is also made available online.
  For the links to these online repositories please see Section~\ref{sec:results}.
\end{itemize}

%%%%%%%%%%%%%%%%%%%%%%%%%%%%%%%%%%%%%%%%%%%%%%%%%%%%%%%%%%%%%%%%%%%%%%%%%%%%%%%%
%%%%%%%%%%%%%%%%%%%%%%%%%%%%%%%%%%%%%%%%%%%%%%%%%%%%%%%%%%%%%%%%%%%%%%%%%%%%%%%%
%%%%%%%%%%%%%%%%%%%%%%%%%%%%%%%%%%%%%%%%%%%%%%%%%%%%%%%%%%%%%%%%%%%%%%%%%%%%%%%%
\section{Related Work} \label{sec:related}
The main underlying problem in map alignment is \emph{data association}, which manifests in a variety of forms according to the application context.
Few examples are image registration (e.g. stereo vision correspondence, optical flow, and visual odometry), laser scan matching and point cloud registration in Simultaneous Localization and Mapping (SLAM), and the correspondence problems in SLAM such as loop closure and partial map merging.
While above-mentioned problems share the underlying challenge of data association, different methods formulate their underlying problem differently depending on the context of their application, data type, and prior assumptions.
While we try to point out some of the seminal works with formulations other than those related to our work, we turn the focus of the literature review to those closest to ours, i.e. map alignment.

%%%%%%%%%%%%%%%%%%%%%%%%%%%%%%%%%%%%%%%%
\paragraph{Motivations of sensor to prior map alignment}
There are different motivations for fusing prior maps and sensor maps.
%%%%%%%%%% prior map -  a source of semantics and a common ground for communication
For instance, Sanchez and Branaghan argue that abstract maps are easier to learn \cite{sanchez2009interaction}, and accordingly, Georgiou et al.~\cite{georgiou2017constructing} state that a correspondence between an abstract human readable map and robot's sensor map is desired to facilitate collaborative tasks between humans and robots.
Bowen-Biggs et al. claim that sensor maps are not ``natural'' for many high level tasks~\cite{bowen2016sketched}, especially those including semantics or with human in the loop.
In their work, they present a method of fusing two sensor and floor maps, and using the combination for accomplishing elaborate tasks.
However, in their work the map to map correspondence is established manually.
%%%%%%%%%% prior map - slam improve
In other examples, the prior map is exploited towards improving the performance of SLAM algorithms, either through exploiting the structure of the prior map, or by aligning local maps to build a global map.
Georgiou et al.~\cite{georgiou2017constructing} formulated the ``structural information from architectural drawings'' as ``informative Bayesian mapping priors'' in order to improve the performance of the SLAM algorithm.
Although, this work does not address the map alignment problem per se.
Instead the SLAM output is structured according to the prior information embedded into the SLAM algorithm.
Vysotska and Stachniss~\cite{vysotska2017improving} proposed an approach to improve SLAM performance by generating constraints from the correspondence between the building information from \emph{OpenSteetMap} and the robot's perception of its surroundings.
They also benefit from the ``localizability'' information available in the OpenSteetMap.
Mielle et al.~\cite{mielle2017slam} proposed a method for applications with extreme conditions (e.g. with dust or smoke) where the information from a ``rough prior'' is incorporated in order to improve the SLAM performance, and enhance the quality of the rough prior map by fusing it with sensor map.

%%%%%%%%%%%%%%%%%%%%%%%%%%%%%%%%%%%%%%%%
\subsection{Graph matching approaches}
Topological structure of the open spaces is one of the most salient information in the maps, and it is natural that the graph representation of the aforementioned structure draws much attention as a fitting representation.
Two of the sub-problems in graph theory that are most relevant to map alignment are the Maximal Common Sub-graph (MCS) problem, and the error-tolerant sub-graph isomorphism~\cite{huang2005topological}.

%%%%%%%%%%
Huang and Beevers~\cite{huang2005topological} proposed a method for merging partial maps based on the embedded topological maps.
Their approach is based on a graph matching process inspired by maximal common sub-graph (MCS) and image registration, followed by a second stage in which the geometric consistencies of the match hypotheses are evaluated.
The vertices of the topological map are embedded in a metric space, along with a minimal amount of metric information (e.g., orientation of edges at each vertex and path length for each edge).
Therefore, their method benefits from both the geometric and topological information of the open spaces.

%%%%%%%%%%
In another work with a similar approach, Wallgr{\"u}n \cite{wallgrun2010voronoi} proposes a map alignment technique with a graph matching method based on the Voronoi graph of the maps.
The objective of his work is localization and mapping, and the underlying data association model of his method is based on an inexact graph matching with graph edit distance, over annotated graphs generated from the Voronoi graphs.
Nodes are annotated with the radii of the maximal inscribed circles used to generate the Voronoi graph, and the edges are annotated with their relative length, the shape of the Voronoi curve beneath the edge, and the edge's traversability.
By assigning such attributes to the elements of the graph, he incorporates geometric constraints into the matching process.

%%%%%%%%%%
In order to develop an automated process for map quality assessment, Schwertfeger and Birk have developed an interesting method for map alignment~\cite{schwertfeger2013evaluation}.
Their method captures the high-level spatial structures of the maps through Voronoi graphs, and represents with topological graphs that contain the angles between edges and the length of edges.
The map alignment is done by finding similar vertices of the graphs and ``identification of sub-graph isomorphisms through wave-front propagation''~\cite{schwertfeger2013evaluation}.
With experimental results, they show the robustness of their method by detecting brokenness in sensor maps.

%%%%%%%%%%
In another intriguing work, Mielle et al.~\cite{mielle2016using} proposed a map alignment method based on graph matching, which enables robots to follow navigation orders specified in sketch maps.
Their method converts the Voronoi skeleton to a graph, where vertices are the bifurcation and ending points of the skeleton.
Vertex type (dead-end or junction) and an ordered list of edges are attributed to the graph's vertices in the matching process.
To find the error-tolerant maximal common sub-graph (ETMCS), they developed a modified version of Neuhaus and Bunke's~\cite{neuhaus2004error} graph matching algorithm based on the normalized Levenshtein edit-distance (LED)~\cite{yujian2007normalized}.
By skipping the absolute position values, the interpretation becomes insensitive to noise and inconsistency of the map.
Consequently their method doesn't require global consistency and uniform scaling of the maps. % to handle sketch maps.

%%%%%%%%%%
In order to benefit from semantic information available in floor maps for high level task execution, Kakuma et al. \cite{kakuma2017alignment} proposed a graph matching based method for the alignment of sensor maps to floor plans of the buildings.
Their method constructs a graph from segmented regions of the occupancy map.
Graph matching is carried out with minimizing a matching cost function based on a variation of Graph Edit Distance (GED)~\cite{sanfeliu1983distance} and Hu-Moments~\cite{hu1962visual}.

%%%%%%%%%%%%%%%%%%%%%%%%%%%%%%%%%%%%%%%%
\subsection{Hough/Radon transform approaches}
Hough (/Radon) transform maps the input signal from the Cartesian to a \emph{parametric} space.
This parametric space has the advantage of capturing the salient, thought maybe latent, structure of the maps.
The core of those methods based on Hough transform is to decompose the alignment problem into rotation and translation estimation.
Such approaches are often deterministic, non-iterative, and fast, thanks to this decomposition.
However, methods in this category are limited to rigid transformation, and work best on maps with same modalities.

%%%%%%%%%%
For merging partial maps in a multi-robots application, Carpin proposed a method \cite{carpin2008fast} that first finds the rotation alignment via a correlation between the Hough spectra of the two maps.
After the orientation alignment, the translation parameters are estimated from a x-y projection of the maps.
One of the interesting features of this method is that the estimated transformations are weighted and such weights could be treated as uncertainties.

%%%%%%%%%%
With a conceptually similar approach, Bosse and Zlot~\cite{bosse2008map} tackle the problem of global mapping by merging local maps.
Their method also decouples the rotation and translation estimations, but with some twists in their transformation techniques.
They use an ``orientation histogram of the scan normals'' (yields an output similar to a Hough transform) to determine the orientation alignment.
Then a ``weighted projection histograms created from the orthogonal projections'' (somewhat equivalent to radiography) is used for estimating the translation between the orientationally aligned data.

%%%%%%%%%%
Saeedi et al.~\cite{saeedi2012efficient}, \cite{saeedi2014map}, \cite{saeedi2014group} proposed a novel technique to represent the topology of the open space with a probabilistic Voronoi graph.
Even though they employ a graph representation, they do not solve the matching problem by graph matching techniques.
First a Radon transform is employed to find the relative orientation between maps, followed by an edge matching technique based on a 2D cross correlation over graphs' edges to find the translation.
One of the very interesting features of their method is the propagation of the uncertainty from input map to the Voronoi graph, and accounting for this uncertainty in the fusion process.

%%%%%%%%%%%%%%%%%%%%%%%%%%%%%%%%%%%%%%%%%%%%%%%%%%%%%%%%%%%%%%%%%%%%%%%%%%%%%%%%
\subsection{Optimization approaches}
One of the most popular categories of techniques for data associations in robot mapping is based on optimization.
A famous example is the Iterative Closest Point (ICP)~\cite{besl1992method} which is a \emph{point set registration} and finds a rigid transformation between two point sets.
Such an approach is inherently susceptible to the problem of local minima.
They are only suitable to problems where a [rough] initial estimate of the displacement between input data is available.
While this is a reliable assumption in incremental mapping, it is not a valid assumption in map alignment.
Furthermore, such methods work on same modality input data.
%%%%%%%%%%
Other similar approaches in image alignment, such as Lucas-Kanade algorithm~\cite{lucas1981iterative},~\cite{Baker2004} and Enhanced Correlation Coefficient (ECC) Maximization~\cite{evangelidis2008parametric}, also work under similar assumptions and consequently they are prone to similar pitfalls.

%%%%%%%%%%
One example of optimization based method applied to the map merging problem is presented by Carpin and Birk~\cite{carpin2005stochastic}, \cite{carpin2005map}, \cite{birk2006merging}.
Their approach minimizes a dissimilarity function (overlapping quality index) over the transformation parameters, with a stochastic process (random walk), used for the optimization.
An interesting feature of this method is its ability to robustly handling unstructured environments.

%%%%%%%%%%%%%%%%%%%%%%%%%%%%%%%%%%%%%%%%
\paragraph{What else?}
It is good to mention some other interesting approaches, even though we did not find them particularly relevant in order to investigate them in detail and experiment with them.
%%%%%%%%%%
Among those are methods from the multi-robot mapping applications where the alignment of individual maps are determined by localizing each robot in the partial maps of other robots.
Works by Thrun~\cite{thrun2001probabilistic}, Dedeoglu and Sukhatme~\cite{dedeoglu2000landmark}, and Williams et al.~\cite{williams2002towards} are good examples in this category.
These methods are based on the assumption that the input maps are from the same modality.
%%%%%%%%%%
With a similar application, i.e. multi-robot exploration, some researchers have developed methods to determine the relative transformation between robots' partial maps when the robots can physically meet in the environment.
Examples of the methods based on ``rendezvous'' or ``mutual observation'' are proposed by Howard et al.~\cite{howard2006experiments}, Howard~\cite{howard2004multi}, Fox et al.~\cite{fox2006distributed}, Zhou and Roumeliotis~\cite{zhou2006multi}, and Konolige et al.~\cite{konolige2003map}.
These methods are based on the robots' ability to meet and generate transformation hypotheses from a rendezvous, which is unfeasible for off-line methods.
%%%%%%%%%%
Erinc et al. proposed a method~\cite{erinc2013heterogeneous} to annotate heterogeneous maps with WIFI signal that provides cues for data association between maps.
This means two essentially different maps are annotated by a shared landmark, which provides a seamless data association cue.
%%%%%%%%%%
Boniardi et al.~\cite{boniardi2015robot} developed a method for localizing and navigating directly in a sketch map, without the map alignment.
%%%%%%%%%%
Partial map alignment is an essential component of map merging.
Saeedi et al.~\cite{saeedi2016multiple} provided a thorough review of the multi-robot SLAM field which covers a broad range of such methods.
However, most of these methods, being specifically developed to improve multi-robot mapping applications, are dependent on sources of information that are specific to robot map and not accessible for any arbitrary map (e.g. layout maps).
The work by Bonanni et al.~\cite{bonanni2017map} on merging 3-D maps, and its earlier version on 2D maps targeting the problem of merging ``partially consistent maps''~\cite{bonanni2014merging}, require the pose graph to be available (or computable) for both maps.
%%%%%%%%%%
Jiang et al.~\cite{jiang2017simultaneous} proposed a method based on ``motion averaging'' for merging multiple local maps.
Their approach is to find transformation between all local map, construct a graph of the inter-map ``motion'' and optimize such motions for optimal alignment.
However, the core of this method is the optimization of alignment between several local maps, and therefore not suitable for aligning a pair of maps.

%%%%%%%%%%%%%%%%%%%%%%%%%%%%%%%%%%%%%%%%%%%%%%%%%%%%%%%%%%%%%%%%%%%%%%%%%%%%%%%%
%%%%%%%%%%%%%%%%%%%%%%%%%%%%%%%%%%%%%%%%%%%%%%%%%%%%%%%%%%%%%%%%%%%%%%%%%%%%%%%%
%%%%%%%%%%%%%%%%%%%%%%%%%%%%%%%%%%%%%%%%%%%%%%%%%%%%%%%%%%%%%%%%%%%%%%%%%%%%%%%%
\section{Method} \label{sec:method}
\begin{figure}%[!ht]
  \centering
  \begin{subfigure}{\linewidth}
  \centering
    \includegraphics[width=\linewidth]{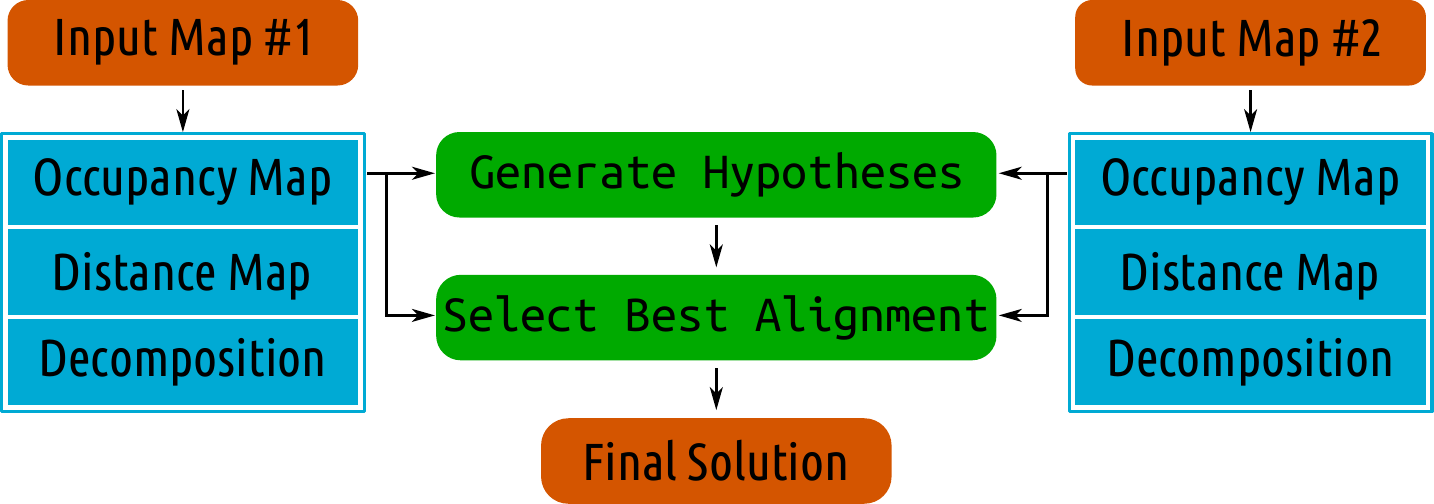}
    \caption[xxx]
            {The orange blocks represent inputs and output, blue blocks are the intermediate representations, and the green blocks are the alignment processes.}
  \label{subfig:method_core}
  \end{subfigure}%

  \begin{subfigure}{\linewidth}
    \includegraphics[width=\linewidth]{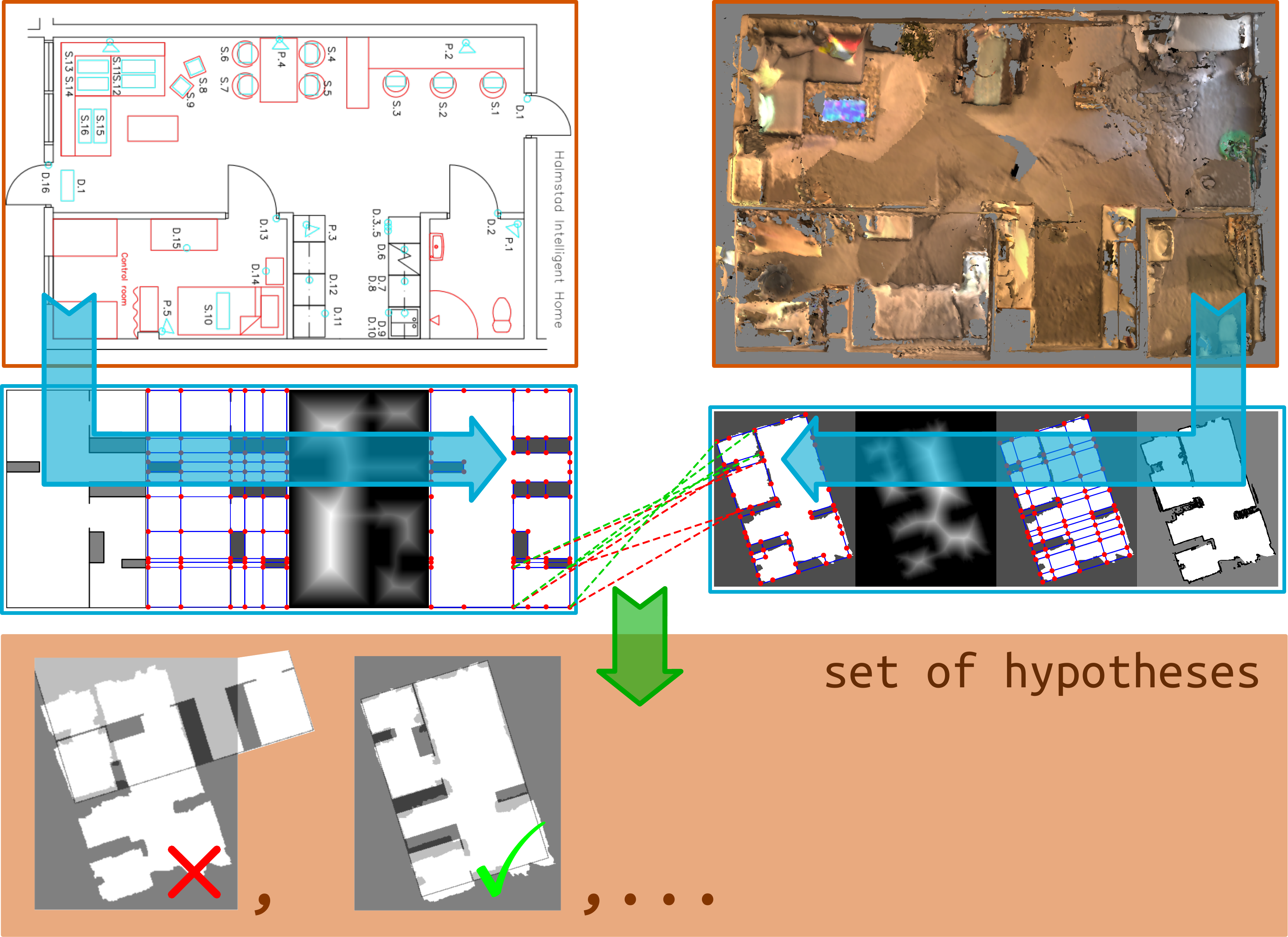}
    \caption{An example of the process on a pair real world maps.}
    \label{subfig:method_detailed}
  \end{subfigure}%
  \caption[xxx]{
    The outline of our method in Figure~\ref{subfig:method_core}, and a concrete example demonstrating the process on real maps in Figure~\ref{subfig:method_detailed}.
  } \label{fig:method}
\end{figure}

The essence of our method, as depicted in Figure~\ref{fig:method}, is to abstract the representation of input maps in order to facilitate the search for alignment.
This abstraction consists of modeling the physical entities of the environment with 2D geometric objects (such as lines and circles.)
These models are then used to partition the map into separate regions with a 2D arrangement.
We have shown in our earlier works~\cite{shahbandi2014sensor},~\cite{shahbandi2015semi}, how this representation could be used for semantic annotation and place categorization of occupancy grid maps.
Section~\ref{subsec:interpretation} describes the 2D arrangement representation.
Furthermore, we explain how this representation is adjusted to capture \emph{meaningful} regions, i.e. adjusting the structural decomposition to region segmentation.
Section~\ref{subsec:alignment} describes the alignment procedure, that is the matching of regions in the maps and estimation of alignment transformation for each match, resulting in a pool of plausible hypotheses.
While each hypothesis is estimated from matching only two regions, they are evaluated based on how well they align the two maps as a whole.
We introduce a ``match score'' in Section~\ref{subsec:match_score}, that is used for comparing the quality of alignments, which in turn is used to pick the best hypothesis.

%%%%%%%%%%%%%%%%%%%%%%%%%%%%%%%%%%%%%%%%%%%%%%%%%%%%%%%%%%%%%%%%%%%%%%%%%%%%%%%%
\subsection{Map interpretation} \label{subsec:interpretation}% parent={sec:method}
\begin{figure}%[!ht]
  \centering
  \begin{subfigure}{.32\linewidth}
    \centering
    \includegraphics[width=\linewidth]{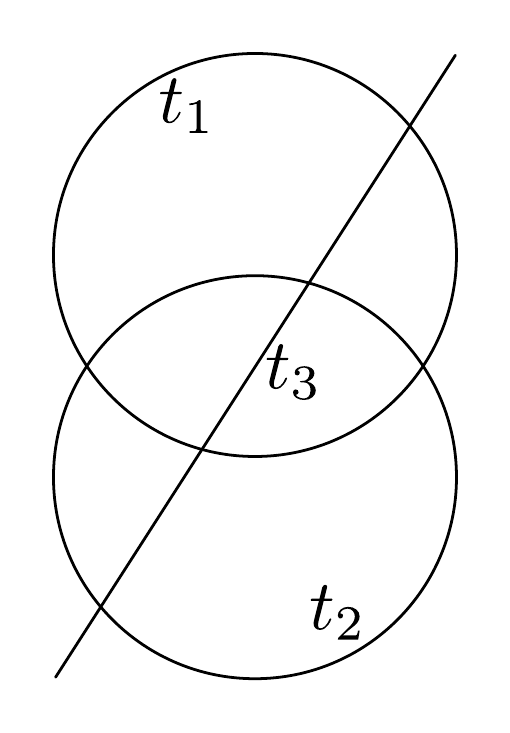}
    \caption{geometric traits}% $\mathcal{T}$}
    \label{subfig:arrangement_traits}
  \end{subfigure}%
  ~%
  \begin{subfigure}{.32\linewidth}
    \centering
    \includegraphics[width=\linewidth]{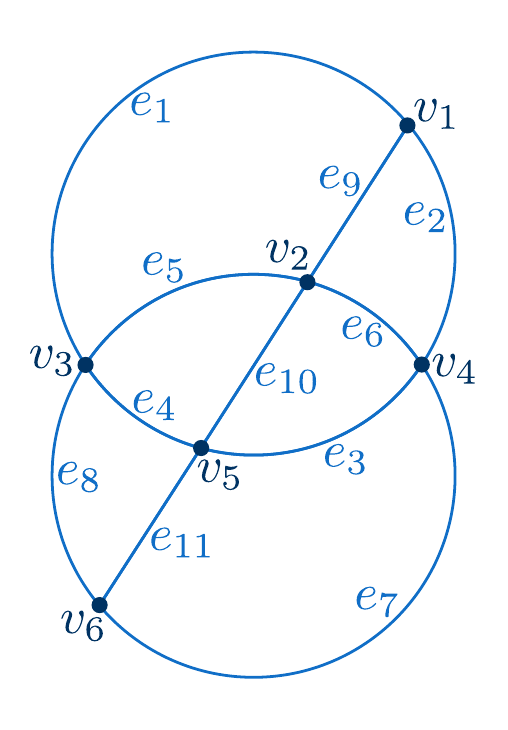}
    \caption{prime graph}% $\mathcal{P}$}
    \label{subfig:arrangement_graphs}
  \end{subfigure}%
  ~%  
  \begin{subfigure}{.32\linewidth}
    \centering
    \includegraphics[width=\linewidth]{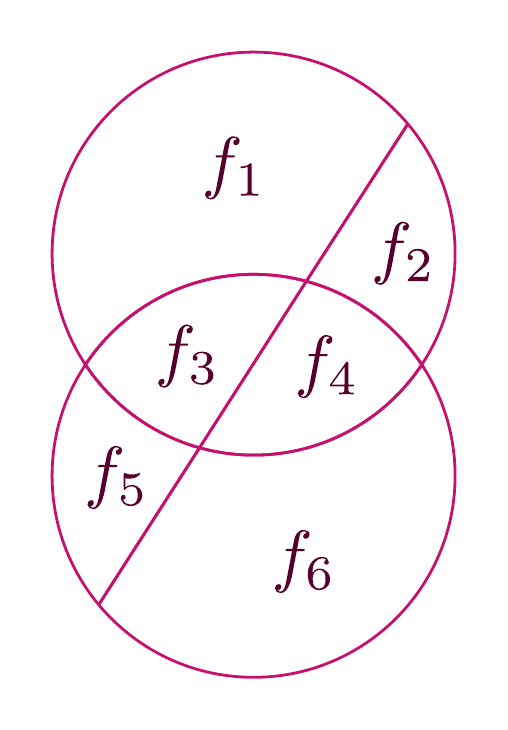}
    \caption{faces}%{partitioned faces}% $\mathcal{F}$}
    \label{subfig:arrangement_faces}
  \end{subfigure}%
  \caption[xxx]{
    An arrangement $\mathcal{A}:= (\mathcal{T}, \mathcal{P},\mathcal{F})$.
    This example involves straight lines and circles to demonstrates the ability to handle geometric traits beyond straight lines.
    However, due to the nature of buildings in our maps, our map interpretation with arrangement only relies on straight lines.
  } \label{fig:arrangment_demo}
\end{figure}

First step towards map alignment is the modeling of maps with an abstract representation, i.e. arrangement \cite{agarwal2000arrangements}.
Algorithm~\ref{alg:interpretation} outlines the process of this abstraction, composed of geometric trait detection, decomposition (arrangement), and pruning of the arrangement from a structural decomposition to a region segmentation.
An arrangement partitions a 2D plane according to a set of geometric objects (such as, but not limited to, lines and circles), referred to as geometric traits and traits for short.
A set of geometric traits $\mathcal{T}$ will result in a unique arrangement $A$ identified by a prime graph $\mathcal{P}$, and a set of faces $F$.
The prime graph $\mathcal{P}$ is the result of intersecting all traits $\mathcal{T}$, and faces are \emph{irreducible} closed-regions (``Jordan Curve'') bounded to edges from the prime graph $V(\mathcal{P})$.
Neighborhood $N(\mathcal{F})$ is an attribute associated with the set of faces $\mathcal{F}$, defined as a set of tuples of faces where each tuple identifies a pair of neighboring faces.
Figure~\ref{fig:arrangment_demo} demonstrates an arrangement and its components on a toy example.
For technical details of the arrangement algorithm, see~\cite{agarwal2000arrangements}.

\begin{algorithm}[t]
  \caption {Map Interpretation} \label{alg:interpretation}
  \begin{algorithmic}%[1]

    \Function {Interpret}{Map}
    \State $\mathcal{T}$ = \Call{DetectGeometricTraits}{Map}
    \State $A(\mathcal{T}, \mathcal{P},\mathcal{F})$ = \Call{Arrangement}{$\mathcal{T}$}
    \State /* \emph{Pruning} */
    \State // $M_d$: normalized Distance Transform of Map
    \State // $N_p(e)$: neighboring cells (pixels) to an edge $e$
    \State // $V(e) = \frac{1}{size(N_p(e))} \sum_{p \in N_p(e)} M_d(p) $
    \State // $thr_e$: wall/gateway detection threshold ($\sim 0.075$)
    \State $E(\mathcal{P}) = E(\mathcal{P}) - \{e \mid e \in E(\mathcal{P}) \land V(e) < thr_e\}$
    \State \Call{Update}{$A(\mathcal{T}, \mathcal{P},\mathcal{F})$}
    \State \Return $A$
    \EndFunction 

    \\\hrulefill
    \Function {Arrangement}{$\mathcal{T}=\{t_i\}$}
    \State /* \emph{construct prime graph} $\mathcal{P}$ \emph{from traits} $\mathcal{T}$ */
    \State $V(\mathcal{P}): \{ v_i(x,y) \mid (\exists t_j,t_k \in \mathcal{T}) [v_i \in t_j \land v_i \in t_k] \}$
    \State $E(\mathcal{P}): \{ e_i(t_k, v^i_s, v^i_e ) \mid (\nexists v_m \in e_i) [v_m \neq v^i_s \land v_m \neq v^i_e] \}$
    %% \State // $E(\mathcal{P}): \{ e_i: (t_i, v_s, v_e ) \mid t_i \in \mathcal{T}, \quad v_s, v_e \in V(\mathcal{P}) \}$
    \State /* \emph{identify irreducible faces} $\mathcal{F}$ \emph{in} $\mathcal{P}$ */
    \State $\mathcal{F}: \{ f_i:=\{e_j\} \mid (\forall e_j \in f_i) [\exists! e_k \in f_i \mid v^j_s = v^k_e]\}$
    %% \State $\mathcal{F}: \{ f_i:=\{e_j\} \mid e_j \in E(\mathcal{P})\}$
    \State $N(\mathcal{F}): \{ (f_i,f_j) \mid (\exists e_k) [e_k\in f_i \land e_k\in f_j] \}$
    \State \Return $A\left( \mathcal{T},\mathcal{P},\mathcal{F}\right)$
    \EndFunction 

  \end{algorithmic}
\end{algorithm}

%%%%%%%%%%%%%%%%%%%%%%%%%%%%%%%%%%%%%%%%
\subsubsection{Geometric traits}
We model the physical elements of the buildings (e.g occupied pixels in occupancy maps) with geometric traits, which represent the boundary between open spaces and occupied (or unexplored) areas.
Accordingly, an arrangement manifests a \emph{dual} characteristic:
\begin{inparaenum}[i)]
\item $\mathcal{F}$ is a geometric representation of the open-space and its boundaries, and 
\item $N(\mathcal{F})$ captures the topology of open-space.
\end{inparaenum}
The detection of the geometric traits from 2D maps could be achieved by common algorithms such as \emph{Generalized Hough Transform}~\cite{ballard1981generalizing} and \emph{Radon Transform}~\cite{radon1986determination}.
Given that all maps used in our experiments could be modeled with only straight lines, in this work we use \emph{radiography}, which is a variation of the aforementioned algorithms.
Radiography operates as a Radon Transform that is filtered by the oriented gradient of the image.
That is, the projection of each point is weighted by the magnitude of the image's gradient at that point, multiplied by the difference between the orientation of the image's gradient and the direction of the Radon projection.
We have shown previously~\cite{shahbandi2014sensor} that radiography is more robust in modeling physical elements of the environment (e.g walls) that suffer from a discrepancy in their continuity or too much noise.
Nonetheless, the arrangement representation is neither limited to straight lines, nor dependent on the trait detection technique.

%%%%%%%%%%%%%%%%%%%%%%%%%%%%%%%%%%%%%%%%
\subsubsection{Abstraction compatibility and arrangement pruning}
\begin{figure}%[!ht]
  \centering
  \begin{subfigure}{.49\linewidth}
    \includegraphics[width=\linewidth]{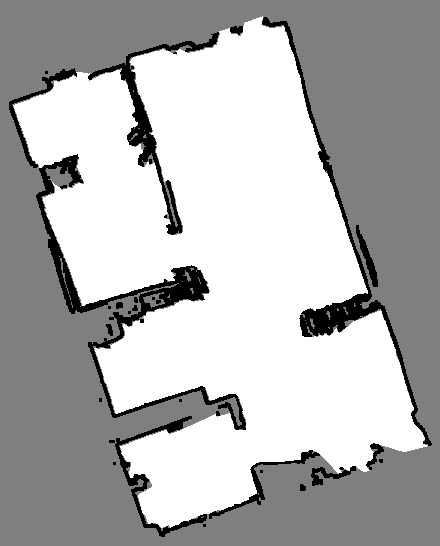}
    \caption{occupancy map}
    \label{subfig:HIH_ogm}
  \end{subfigure}%
  ~%
  \begin{subfigure}{.49\linewidth}
    \includegraphics[width=\linewidth]{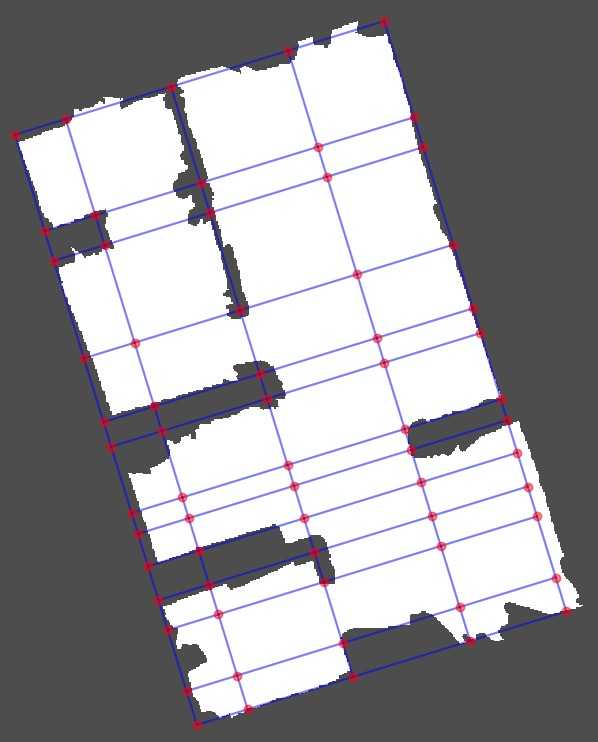}
    \caption{arrangement}
    \label{subfig:HIH_arrangement}
  \end{subfigure}%

  \begin{subfigure}{.49\linewidth}
    \includegraphics[width=\linewidth]{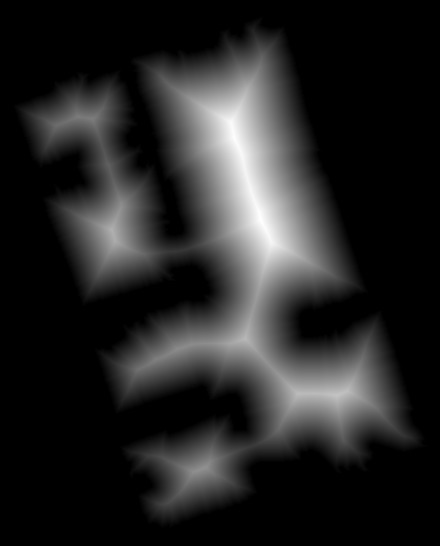}
    \caption{distance image}
    \label{subfig:HIH_dist_map}
  \end{subfigure}%
  ~%
  \begin{subfigure}{.49\linewidth}
    \includegraphics[width=\linewidth]{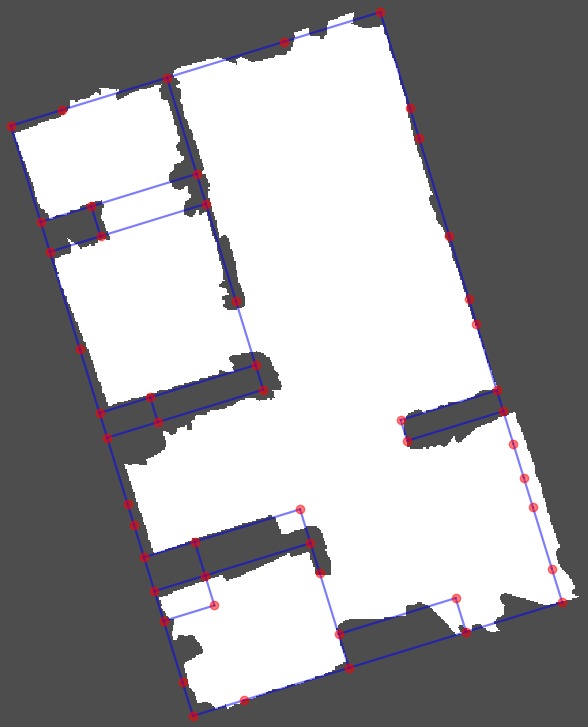}
    \caption{after pruning}
    \label{subfig:HIH_arrangement_prune}
  \end{subfigure}%
  \caption[xxx]{
    An Occupancy Map in Figure~\ref{subfig:HIH_ogm},
    its original decomposition in Figure~\ref{subfig:HIH_arrangement},
    and a cleaned-up version of the arrangement (pruned) in Figure~\ref{subfig:HIH_arrangement_prune},
    based on the distance map $M_d$ in Figure~\ref{subfig:HIH_dist_map}.}
  \label{fig:arrangement_prunning}
\end{figure}

Despite its merits in detecting discontinuous traits in noisy maps and capturing the global structure of environments~\cite{shahbandi2014sensor}, radiography detects unbounded traits (e.g. infinite lines instead of line segments).
Consequently the partitioning of the space is not equivalent to a plausible region segmentation, due to over-decomposition of areas that are conceptually a single region (e.g. a kitchen or an office).
Figure~\ref{fig:arrangement_prunning} demonstrates the over-decomposition of a real map.
This inconsistency of region segmentation is non-deterministic, depends on the noise, partiality and inconsistencies of the maps, and could vary from sensor maps to layout maps.
Since the essence of our alignment method is to match corresponding regions, it is crucial for the maps to have representations on the same level of abstraction (regions segmentation), what we call \emph{abstraction compatibility}.
That is to say, if a single face represents a room in one map, the same room must be represented by a similar face in the other map.
Based on empirical observations, the success rate of our method seems to be most sensitive to this compatibility assumption.
However, the sensitivity of the alignment method to abstraction compatibility is not critically obstructive, and not every corresponding region should have compatible abstraction in the maps.

We remedy this challenge by \emph{pruning} the arrangement to a more plausible region segmentation, presented in the function \textproc{Interpret} in Algorithm~\ref{alg:interpretation}.
This pruning is the process of removing all face boundaries (i.e. $e_i \in E(\mathcal{P})$) which do not correspond to a \emph{wall} or \emph{gateway}, followed by merging all adjacent faces whose boundaries are removed.
As described in Algorithm~\ref{alg:interpretation}, an edge is considered a gateway or wall if the average value of the pixels $V(e)$ from the distance map $M_d$ (Figure~\ref{subfig:HIH_dist_map}) within a neighborhood of the that edge $N_p(e)$ is below a certain threshold $thr_e$.
Since the distance map $M_d$ is normalized (scaled to $[0,1]$), the threshold $thr_e$ over the averaged pixel values $V(e)$ is independent of the map scale.
The result of a pruning process applied to the over-decomposed example of Figure~\ref{subfig:HIH_arrangement}, is presented in Figure~\ref{subfig:HIH_arrangement_prune}.

Our pruning of an over-decomposed arrangement to region segmentation is a variation of ``Morphological Region Segmentation'' as presented in the ``Room Segmentation'' survey by Bormann et al.~\cite{bormann2016room}.
In more elaborate scenarios, one could employ other region segmentation methods presented in \cite{bormann2016room}, or more recent works by Fermin-Leon et al.~\cite{leon2017incremental} and Mielle et al.~\cite{mielle2017method}.
Nevertheless, we observed that the arrangement pruning, along with an approximation of faces with \emph{Oriented Minimum Bounding Boxes} (details in Section~\ref{subsec:alignment}), satisfies the abstraction compatibility assumption.

%%%%%%%%%%%%%%%%%%%%%%%%%%%%%%%%%%%%%%%%%%%%%%%%%%%%%%%%%%%%%%%%%%%%%%%%%%%%%%%%
\subsection{Alignment procedure} \label{subsec:alignment}% parent={sec:method}

A hypothesis in the context of this work is a transformation function between the coordinate frames of the two maps.
Hypothesis generation is the process of proposing such plausible transformations.
According to the \emph{uniform scaling assumption} stated in Section~\ref{subsec:our_approach}, the transformations estimation is restricted to ``similarity'' (i.e. translation, rotation and uniform scaling.)

To propose hypotheses, faces of the open space regions with similar \emph{shapes} are associated and a transformation is estimated for each pair of faces with similar shapes.
The \emph{shape descriptor} is an ordered sequence of vertex-edge tuples
\[
\begin{array}{l}
  D(f):= ( (v_i,e_l), \cdots, (v_j,e_k))\\
\end{array}
\]
in which, all the entries (vertices and edges) are ordered counter clock-wise from an arbitrary reference point.
Since the choice of reference point in each face is arbitrary, the descriptor also contains all its ``cyclic shifts''.
For example, the descriptor for face $f_1$ of the arrangement from the Figure~\ref{fig:arrangment_demo} with all its $s$-step shifts are
\[
\begin{array}{l l}
  D(f^{s=1}_1) & = ( (v_2,e_9) , (v_1,e_1) , (v_3,e_5) )\\
  D(f^{s=2}_1) & = ( (v_3,e_5) , (v_2,e_9) , (v_1,e_1) )\\
  D(f^{s=3}_1) & = ( (v_1,e_1) , (v_3,e_5) , (v_2,e_9) )\\
\end{array}
\]

\begin{figure}%[!ht]
  \centering
  \begin{subfigure}{.33\linewidth}
    \centering
    \includegraphics[width=.8\linewidth]{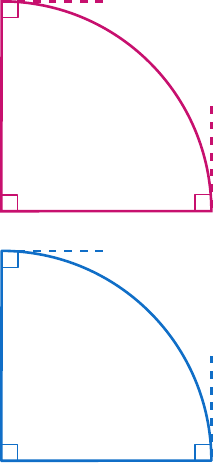}
    \caption{} \label{subfig:shape_mismatch_a}
  \end{subfigure}%
  \hfill%
  \begin{subfigure}{.33\linewidth}
    \centering
    \includegraphics[width=.8\linewidth]{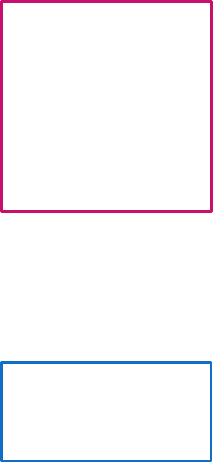}
    \caption{} \label{subfig:shape_mismatch_b}
  \end{subfigure}%
  \hfill%
  \begin{subfigure}{.33\linewidth}
    \centering
    \includegraphics[width=.8\linewidth]{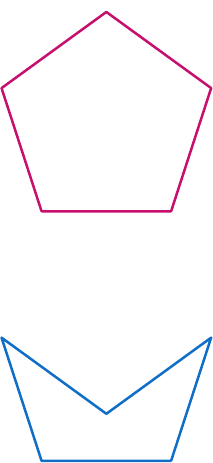}
    \caption{} \label{subfig:shape_mismatch_c}
  \end{subfigure}%
  \caption[xxx]{
    Examples where a missing feature results in wrong matches.
    In the absence of \emph{edge length ratio} those faces in Figure~\ref{subfig:shape_mismatch_a} could have three matches instead of one, and the faces in Figure~\ref{subfig:shape_mismatch_b} could have four matches instead of zero.
    In the absence of \emph{corner angle} those faces in Fig.~\ref{subfig:shape_mismatch_c} could have five matches instead of none.
  } \label{fig:shape_mismatch}
\end{figure}

In the context of the shape descriptor, vertices denote corners, where corners are defined as vertices with any internal angles other than $\pi$.
The features of the descriptor are the internal angles of vertices (i.e. \emph{corner angle}), and the length of edges normalized to the perimeter of the face (i.e. \emph{edge length ratio}).
Figure~\ref{fig:shape_mismatch} demonstrates the necessity of these features, through examples where the absence of these two features would result in false matches.
Descriptor size equivalency is the first necessary condition for a potential match.
A match is then identified as a $s$-step ``circular shift'' of one descriptor, so that all corresponding entries in the descriptors of the faces are equivalent.
After a correspondence between the two point sets (face corners) is proposed via face matching, a transformation between the two point sets is estimated based on the ``Least-squares estimation'' method proposed by Umeyama~\cite{umeyama1991least}, which uses the singular value decomposition of a covariance matrix of the data points.

Algorithm~\ref{alg:method} presents the map alignment procedure in pseudo-code, and Figure~\ref{subfig:method_detailed} depicts two examples of correct (in green) and wrong (in red) association and their consequent transformation.

\begin{algorithm}[t]
  \caption {Map Alignment Procedure} \label{alg:method}
  \begin{algorithmic}%[1]

    \State \textbf{Input:} Map$_1$, Map$_2$
    \State \textbf{Output:} Alignment (similarity transformation)

    \State
    \State /* \emph{Map Interpretation} */
    \State $A_1(\mathcal{T}_1, \mathcal{P}_1,\mathcal{F}_1)$ = \Call{Interpret}{Map$_1$}
    \State $A_2(\mathcal{T}_2, \mathcal{P}_2,\mathcal{F}_2)$ = \Call{Interpret}{Map$_2$}

    \State        
    \State /* \emph{Face Matching} \& \emph{Hypotheses Generation} */
    \State // $f.c :$ corners of the face $f$
    \State // $D(f) :$ shape descriptor of face $f$
    \State // $f^{s} :=$ $s$-step circular shift of corners and $D(f)$
    \State $H = \emptyset$
    \For {$f_{i} \in \mathcal{F}_1, f_{j} \in \mathcal{F}_2 \mid \Call{Size}{f_i.c}=\Call{Size}{f_j.c}$}
    %% \For {$step = (1,$ \Call{Size}{$D(f_1)$} $,++)$}
    \For {$s$ := 1 to \Call{Size}{$f_i.c$}, step=1}
    \If {$D(f^{s}_i) = D(f_j)$}
    \State $H = H \cup \{ \Call{EstimateTransform}{f^{s}_i.c, f_j.c} \}$ %, \text{\lq similarity\rq}
    \EndIf
    \EndFor
    \EndFor

    \State
    \State /* \emph{Selecting The Best Hypotheses} */
    \State Alignment = $\operatorname*{arg\,max}_{h \in H} $ \Call{MatchScore}{$h, A_1, A_2$}

  \end{algorithmic}
\end{algorithm}

%%%%%%%%%%%%%%%%%%%%%%%%%%%%%%%%%%%%%%%%
\paragraph{Simplified Alignment Procedure}
In the presence of too much noise in a map, the pruning of the arrangement might not return clean-cut shapes desired for face matching.
One wrong corner missed in the pruning process will render the shape of that region useless for matching, if the same error does not occur in the other map for the corresponding region.
One example of missed corners is visible at the bottom of Figure~\ref{subfig:HIH_arrangement_prune}.
Alternatively, due to such cases, we simplify shapes with their \emph{Oriented Minimum Bounding Boxes} (OMBB).
This counts as an interpretation of the ``well-structured environments'' assumption stated in Section~\ref{subsec:our_approach}.
This substitution of the shapes with OMBB renders the descriptor size and corner angles redundant, and consequently, the only relevant shape feature would be the aspect-ratio of the OMBBs (i.e. edge length ratio).
We have noticed that it is computationally cheaper to replace this feature (i.e. edge length ratio), with an equivalent process of \emph{False Positive rejection} which is based on the uniform scaling assumption.
Accordingly, the uniform scale assumption is relaxed, where all the transformations are estimated with an affine model, and then any transformation that does not qualify as a similarity transformation is rejected.
Algorithm~\ref{alg:simp_method} is the modified version of the alignment procedure, which reflects the simplification of the shape descriptor and the rejection of non-uniformly scaled transformations.
The result of the False Positive rejection can be seen in Table~\ref{tab:res_layout_sensor}, where $\sim 90\%$ of initial hypotheses are rejected.
We have carefully monitored the hypotheses pools of cases where the method failed to find the correct alignment.
We can report that a correct alignment was never generated.
In other words, we can safely assure that a rejection of potentially correct hypotheses has never been a cause of failure.

\begin{algorithm}[t]
  \caption {Map Alignment Procedure\\
    with simplified hypotheses generation} \label{alg:simp_method}
  \begin{algorithmic}%[1]

    \State /* \emph{\textbf{Input, Output} (see Algorithm~\ref{alg:method})} 

    \State
    \State /* \emph{Map Interpretation (see Algorithm~\ref{alg:method})} */
    
    \State        
    \State /* \emph{Generate Hypotheses (without face matching)} */
    \State // $\hat{f} = \Call{OrientedMinimumBoundingBox}{f}$
    \State $H = \emptyset$
    \For {$f_{i} \in \mathcal{F}_1, f_{j} \in \mathcal{F}_2$}
    \For {$s$ := 1 to 4, step=1}
    \State $H = H \cup \{ \Call{EstimateTransform}{\hat{f}^s_i.c, \hat{f}_j.c} \}$ % , \text{\lq affine\rq}
    \EndFor
    \EndFor

    \State
    \State /* \emph{Reject False Positives Hypotheses} */
    \State // $h.s_x, h.s_y$: scales of transformation in $x/y$ directions
    \State // $thr_{s}$: acceptable ratio between scales ($\sim 1.2$)
    \For {$h \in H \mid \neg (1/thr_{s} < (h.s_x/h.s_y) < thr_{s}) $}
    \State reject $h$
    \EndFor

    \State
    \State /* \emph{Selecting The Best Hypotheses (see Algorithm~\ref{alg:method})} */

  \end{algorithmic}
\end{algorithm}

%%%%%%%%%%%%%%%%%%%%%%%%%%%%%%%%%%%%%%%%%%%%%%%%%%%%%%%%%%%%%%%%%%%%%%%%%%%%%%%%
%%%%%%%%%%%%%%%%%%%%%%%%%%%%%%%%%%%%%%%%%%%%%%%%%%%%%%%%%%%%%%%%%%%%%%%%%%%%%%%%
%%%%%%%%%%%%%%%%%%%%%%%%%%%%%%%%%%%%%%%%%%%%%%%%%%%%%%%%%%%%%%%%%%%%%%%%%%%%%%%%
\subsection{Alignment match score} \label{subsec:match_score}
To select the winning hypothesis, each hypothesis is evaluated based on how well the arrangements of the two maps ($A_1$ and $A_2$) are aligned under that transformation.
To this end, an \emph{arrangement match score} ($S_A$) is defined to measure the alignment quality of each hypothesis.
The arrangement match score between two arrangements $A_1$ and $A_2$, under the transformation $^1T_2$, is defined as
\[
S_A(A_1,A_2,^1\!T_2) = \displaystyle\sum_{\substack{
    f_i \in \mathcal{F}_1\\
    f_j \in \mathcal{F}_2\\
}} min(w(f_i), w(f_j)) \times s_f(f_i,f_j)
\]
where $w(f)$ is a \emph{weight} assigned to individual faces,
and $s_f$ is the \emph{face match score}.
The weight is defined as the relative surface area of faces to the surface area of the whole arrangement they belong to:
\[
w(f_i) = \frac{\textit{area}(f_i)}{\textit{area}(A)}, \quad \textit{area}(A) = \displaystyle\sum_{\substack{f_k \in \mathcal{F}}} \textit{area}(f_k)
\]

The larger a face is, the higher impact it will have in the arrangement match score.
The face match score $s_f$ is defined as:
\[
s_f(f_i,f_j) =
\begin{cases}
  \frac{e^{\left(\frac{f_i \cap f_j}{f_i \cup f_j}\right)} - 1}{e-1} & if \quad (f_i,f_j) \in \textit{association}\\
  0 & otherwise\\
\end{cases}
\]
where $f_i \cap f_j$ is the surface area of the faces' intersection and $f_i \cup f_j$ is the surface area of the faces' union.
The match score of a face with itself (perfect match) equals one, and the match score of two non-intersecting faces equals zero.
The exponential expression rewards slight improvements close to perfect match more than the slight improvements close to a bad match.

The $\textit{association}$ represents pairs of faces from two arrangements that are associated (not just overlapping) under the transformation.
We define $\textit{association}$ based on three conditions:
\[ \begin{array}{l}
  \textit{association}: \{ (f_i,f_j) \mid \forall f_i \in \mathcal{F}_1, f_j \in \mathcal{F}_2, c_1 \land c_2 \land c_3 \} \\
  c_1: \quad \textit{center}(f_i) \in f_j \land \textit{center}(f_j) \in f_i\\
  c_2: \quad \nexists f_k \in \mathcal{F}_2 \mid \textit{center}(f_k)\in f_i, d(f_i,f_j) > d(f_i,f_k)\\
  c_3: \quad \nexists f_k \in \mathcal{F}_1 \mid \textit{center}(f_k)\in f_j, d(f_i,f_j) > d(f_j,f_k)\\
\end{array} \]
where $\textit{center}(f_i)$ is the center of $f_i$ and $d(f_i,f_j)$ is the difference in surface area of $f_i$ and $f_j$.
First, for the two faces $f_i$ and $f_j$ to be associated, they must enclose each other's center.
We define the center of a face as the ``centroid'' (geometric center) of the vertices of the face.
Condition number two assures a one to one association where a face overlaps with multiple faces from the other arrangement.
In such cases, among all faces of $\mathcal{F}_2$ with their centers enclosed by $f_i \in \mathcal{F}_1$, the face ($f_j \in \mathcal{F}_2$) with most similar size (surface area) is associated with $f_i$.
And the third condition is symmetric to the second condition, i.e. vice versa for $f_i$ to $f_j$.

This match score is devised only for the comparison of different hypotheses for a single pair of maps.
That is to say, the alignment of different sensory maps over a layout map could not be compared with this score,
nor is it suitable to detect the layout to which a sensor map belongs (i.e. layout recognition),
and neither is it suitable as a quantified match accuracy measure.
This matter is better observed in Figures~\ref{fig:match_score_matrix} and \ref{fig:match_score_boxplot} from Section~\ref{sec:results}, where it is discussed with experimental observations.

\begin{figure}
  \centering
  \begin{subfigure}{.5\linewidth}
    \centering
    \includegraphics[width=.8\linewidth]{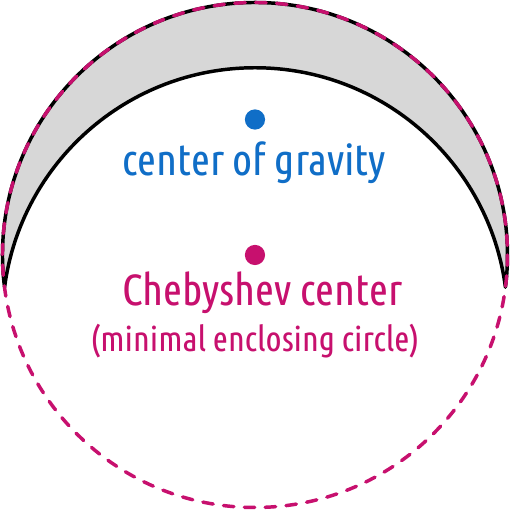}
    \caption{} \label{subfig:centre_challenge_concave}
  \end{subfigure}%
  \hfill%
  \begin{subfigure}{.5\linewidth}
    \centering
    \includegraphics[width=.8\linewidth]{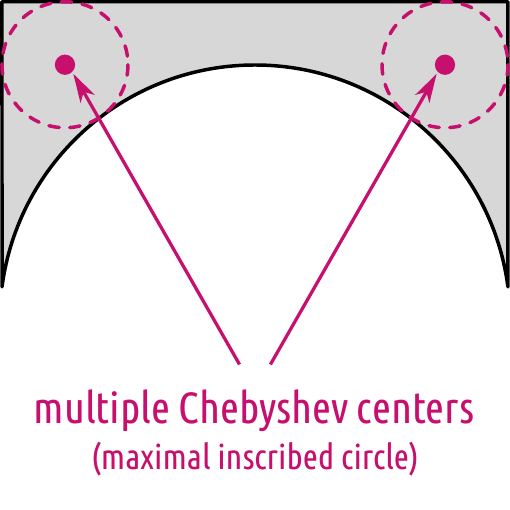}
    \caption{} \label{subfig:centre_challenge_multiple}
  \end{subfigure}%

  \begin{subfigure}{.5\linewidth}
    \centering
    \includegraphics[width=.8\linewidth]{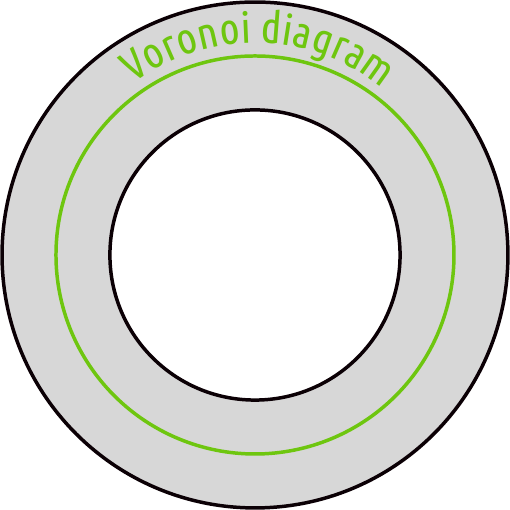}
    \caption{} \label{subfig:centre_challenge_concentric}
  \end{subfigure}%
  \hfill%
  \begin{subfigure}{.5\linewidth}
    \centering
    \includegraphics[width=.8\linewidth]{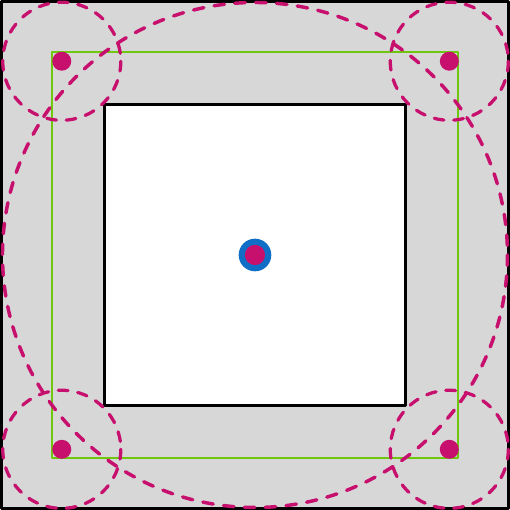}
    \caption{} \label{subfig:centre_challenge_squares}
  \end{subfigure}%
  \caption[xxx]{
    Examples where ``center of gravity'', ``Chebyshev centers'' and ``Voronoi-based'' definitions fail to identify a center point of a region that is both enclosed by the region \emph{and} unique.
    Figures~\ref{subfig:centre_challenge_concave}, \ref{subfig:centre_challenge_multiple} and \ref{subfig:centre_challenge_concentric} highlight the failures of each definition under different circumstances.
    While those three may not seem probable in a real world scenario, Figure~\ref{subfig:centre_challenge_squares} presents the failure of such definitions in a more realistic example.
  } \label{fig:centre_challenge}
\end{figure}

%%%%%%%%%%%%%%%%%%%%%%%%%%%%%%%%%%%%%%%%
\paragraph{The challenge of face center}
The center points of the faces can be defined differently according to the context of the application, such as ``center of gravity'', ``Chebyshev centers'' and ``polygon centroid''.
However, it proves to be very hard to lay down a definition that guarantees to be enclosed by the region \emph{and} unique.
If a face is non-convex as in Figure~\ref{subfig:centre_challenge_concave}, there is no guarantee that the center of gravity would be enclosed by the face.
Chebyshev centers are defined as the center of either 
\begin{inparaenum}[i)]
\item the minimal-radius circle enclosing a region, or
\item the maximal-radius inscribed circle inside the region.
\end{inparaenum}
The example of Figure~\ref{subfig:centre_challenge_concave} shows that the minimal-radius enclosing circle is susceptible to the same problem as the center of gravity, and Figure~\ref{subfig:centre_challenge_multiple} presents an example where the maximal-radius inscribed circle is not necessarily unique.
Another example of tackling this challenge that we have explored, is to extract the Generalized Voronoi Diagram, and picking a point on the skeleton of the influence zone (SKIZ) that has the minimum sum of distance to all other points.
This definition is also prone to degenerate cases, such the example in Figure~\ref{subfig:centre_challenge_concentric} shows.
A variety of definitions could be considered for the center of a region that guarantee to be enclosed by the region.
Alas, they would ultimately depend on the interpretation of ``center point'' with respect to the domain of application, and most are prone to degenerate cases where such a point is not unique.
Ultimately, we have observed that in the setting of our problem such degenerate cases are not so frequent to disturb the performance of the method.
Either of these definition would satisfy the requirements of our method as long as it guarantees uniqueness, and we chose the centeroid of the vertices.

%%%%%%%%%%%%%%%%%%%%%%%%%%%%%%%%%%%%%%%%%%%%%%%%%%%%%%%%%%%%%%%%%%%%%%%%%%%%%%%%
%%%%%%%%%%%%%%%%%%%%%%%%%%%%%%%%%%%%%%%%%%%%%%%%%%%%%%%%%%%%%%%%%%%%%%%%%%%%%%%%
%%%%%%%%%%%%%%%%%%%%%%%%%%%%%%%%%%%%%%%%%%%%%%%%%%%%%%%%%%%%%%%%%%%%%%%%%%%%%%%%
\section{Experimental Results and Verification} \label{sec:results}
In this section we present the results from a series of experiments, on a data-set of forty maps collected specifically for this task.
The experiments are designed to show the method's performance, under different circumstances and in comparison with other methods.
All the experiments are based on an implementation of the method in Python, using many libraries \cite{scipy, NumPy, Matplotlib, SymPy, scikit-image, opencv_library, GPC-python, hagberg2008exploring}.
The source code to our implementation is also available online\footnote{{\scriptsize \url{https://github.com/saeedghsh/Map-Alignment-2D/}}}.

%%%%%%%%%%%%%%%%%%%%%%%%%%%%%%%%%%%%%%%%%%%%%%%%%%%%%%%%%%%%%%%%%%%%%%%%%%%%%%%%
\subsection{Data collection} \label{subsec:data}
To evaluate our method, we collected maps of four different environments in two modalities, of CAD drawings and sensor maps.
All the maps are available online\footnote{{\scriptsize \url{https://github.com/saeedghsh/Halmstad-Robot-Maps/}}}, and presented in Appendix~\ref{app:datasets}.

%%%%%%%%%%%%%%%%%%%%%%%%%%%%%%%%%%%%%%%%
\paragraph{Modalities}
A series of sensor maps were collected by a \emph{Google Tango tablet}, and the \emph{Tango Constructor application} from Google.
The 3D meshes were sliced horizontally and converted to an occupancy-like bitmap, where all the space is open except for the vertices of the mesh.
From there, we generated a pseudo-occupancy map through an interactive ray-casting process.
Detection of the geometric traits from foregoing maps were done via a variation of the Radon transform, namely radiography~\cite{shahbandi2014sensor}.

As for the other modality, the layout maps were obtained from CAD drawings in Portable Document Format (PDF).
These CAD drawings had to be manually simplified before further processing, due to the presence of furniture and other common appliances.
The process involved removing all elements of the drawings, except for the building's elements (i.e mainly walls).
This simplification can be observed in Figure~\ref{subfig:method_detailed} from Section~\ref{sec:method}.
It should be mentioned that the simplified version of the layout map is not tailored to accurately reflect the real layout and what is captured by the sensor map, with no intention to benefit the alignment method.
For instance, walls are represented with single lines in layout maps (width=$\sim$1-2 pixel), while they are much wider in the sensor maps ($\sim$5-10 pixels).
The drawings were converted to Scalable Vector Graphics (SVG) and the geometric traits were obtained directly by parsing the SVG files~\cite{port2017svgpathtools}.
In order to acquire segmented regions and for the sake of convenience, the SVG files were converted to bitmap format (PNG) and the same process of decomposition and arrangement pruning based on distance transform has been employed.
However, if CAD drawings of the layouts are accessible in a richer format (e.g. DXF or DWG), the process of simplification and parsing could also be automated.
Furthermore, if the regions are accessible in such formats, there would not be a need for conversion to bitmap and distance transform for region segmentation.

While all the sensor maps have the same scale, that is, they could be correctly aligned with each other under a rigid transformation, layout maps have different scales compared to the sensor maps, and a rigid transformation could not correctly align sensor maps to layout maps.

%%%%%%%%%%%%%%%%%%%%%%%%%%%%%%%%%%%%%%%%
\paragraph{Environment types}
We collected data from four different environments, two of which are homes and the other two are office buildings.
Table~\ref{tab:data_sets} lists the number of available maps for each environment, and all the maps can be found in Appendix~\ref{app:datasets}.
In total there are forty maps, four of which are layout maps and the rest are sensor maps.
Most sensor maps are partial and vary in their coverage of the environment.

\begin{table}
  \centering
  \begin{tabular}{c | c | c | c}
    environment & type & \# sensor maps & \# layout maps\\
    \hline
    HH\_E5 & office & 14 & 1 \\
    HH\_F5 & office & 14 & 1 \\
    HIH & home  & 4 & 1 \\
    KPT & home    & 4 & 1 \\
    %% \hline
  \end{tabular}
  \caption[xxx]{A list of all maps of four different environments}
  \label{tab:data_sets}
\end{table}

%%%%%%%%%%%%%%%%%%%%%%%%%%%%%%%%%%%%%%%%
\paragraph{Maps that violate our assumptions}
The map collection contains maps that violate some of the initial assumptions.
For instance, maps HH\_E5\_2, HH\_E5\_3, HH\_E5\_4 and HH\_F5\_2 only cover corridors and halls and do not contain any room, and therefore there are not enough segment-able regions for hypotheses generation.
Other examples include HH\_E5\_12 and HH\_F5\_1 which are bent (deformed) and violate the global consistency assumption.
There exist further minor defects in some other maps.
Consequently, the performance results presented here are not the representative of the method's performance under all the assumptions.
Nevertheless, we include these maps to better observe the dependency of the method on the aforementioned assumptions, and provide a more inclusive performance result under different conditions.

\begin{figure}%[!ht]
  \centering
  \begin{subfigure}{.33\linewidth}
    \includegraphics[width=\linewidth]{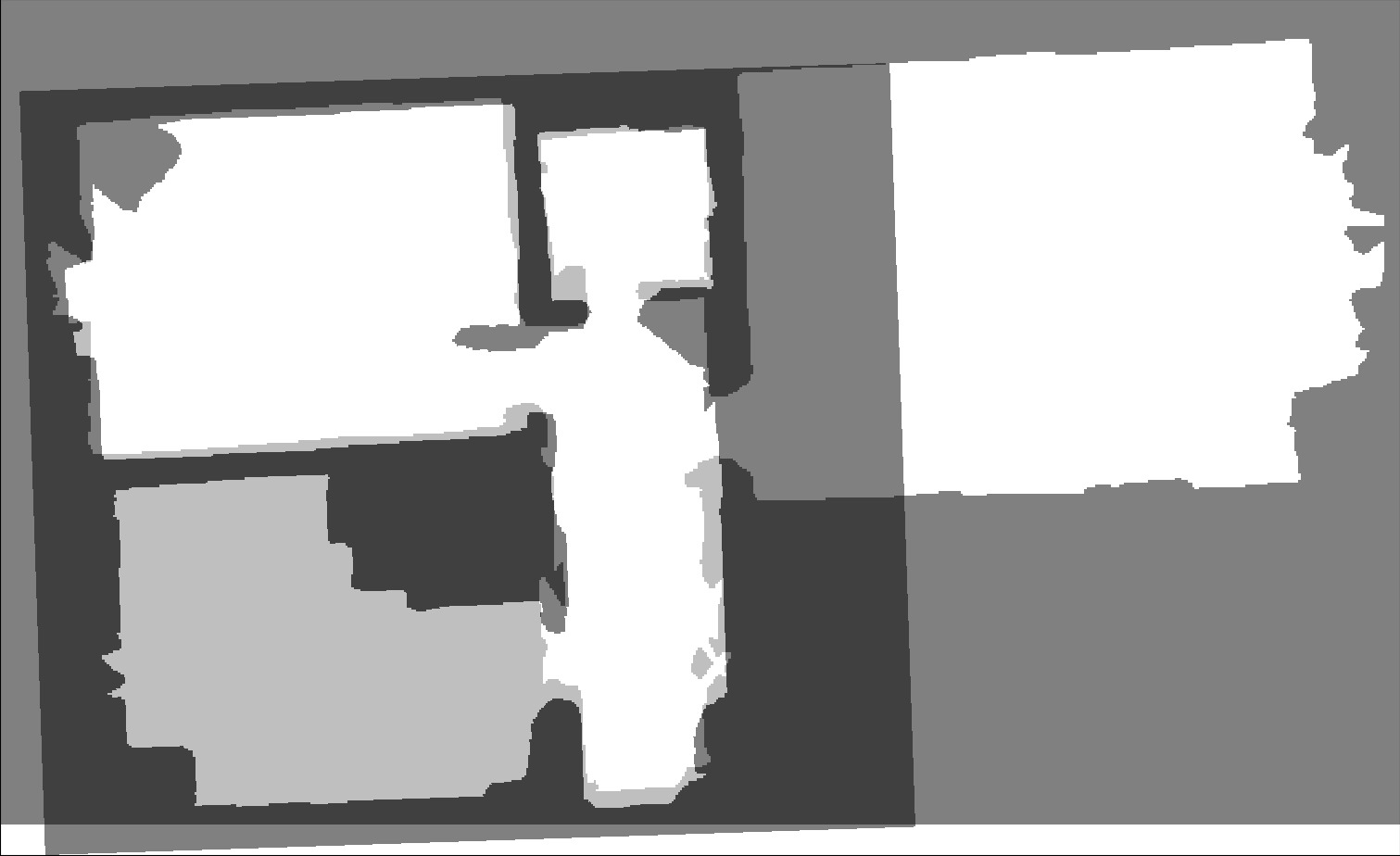}
    \caption{correct}
    \label{subfig:correct_alignment}
  \end{subfigure}%
  ~%
  \begin{subfigure}{.33\linewidth}
    \includegraphics[width=\linewidth]{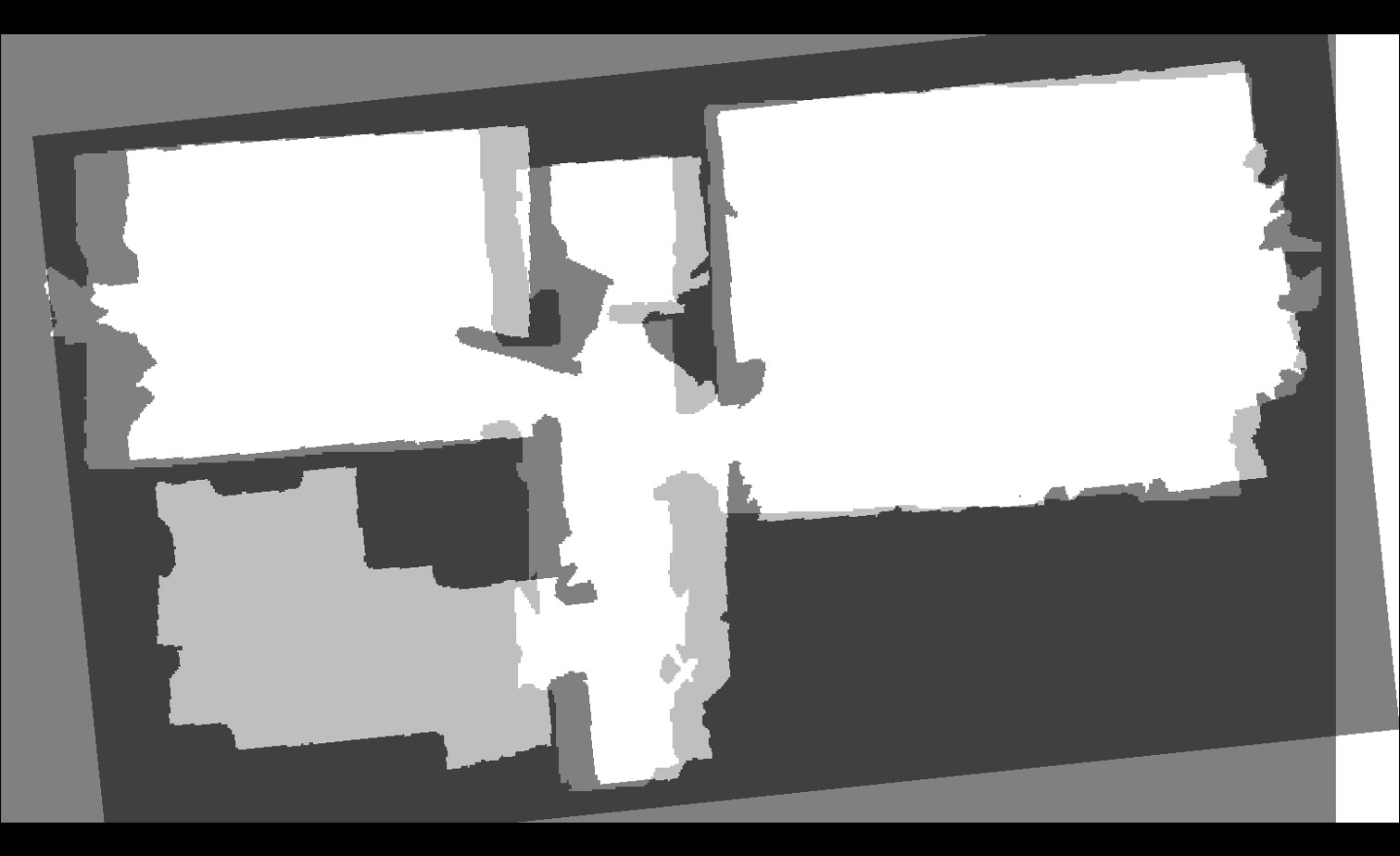}
    \caption{defected}
    \label{subfig:defect_alignment}
  \end{subfigure}%
  ~%    
  \begin{subfigure}{.33\linewidth}
    \includegraphics[width=\linewidth]{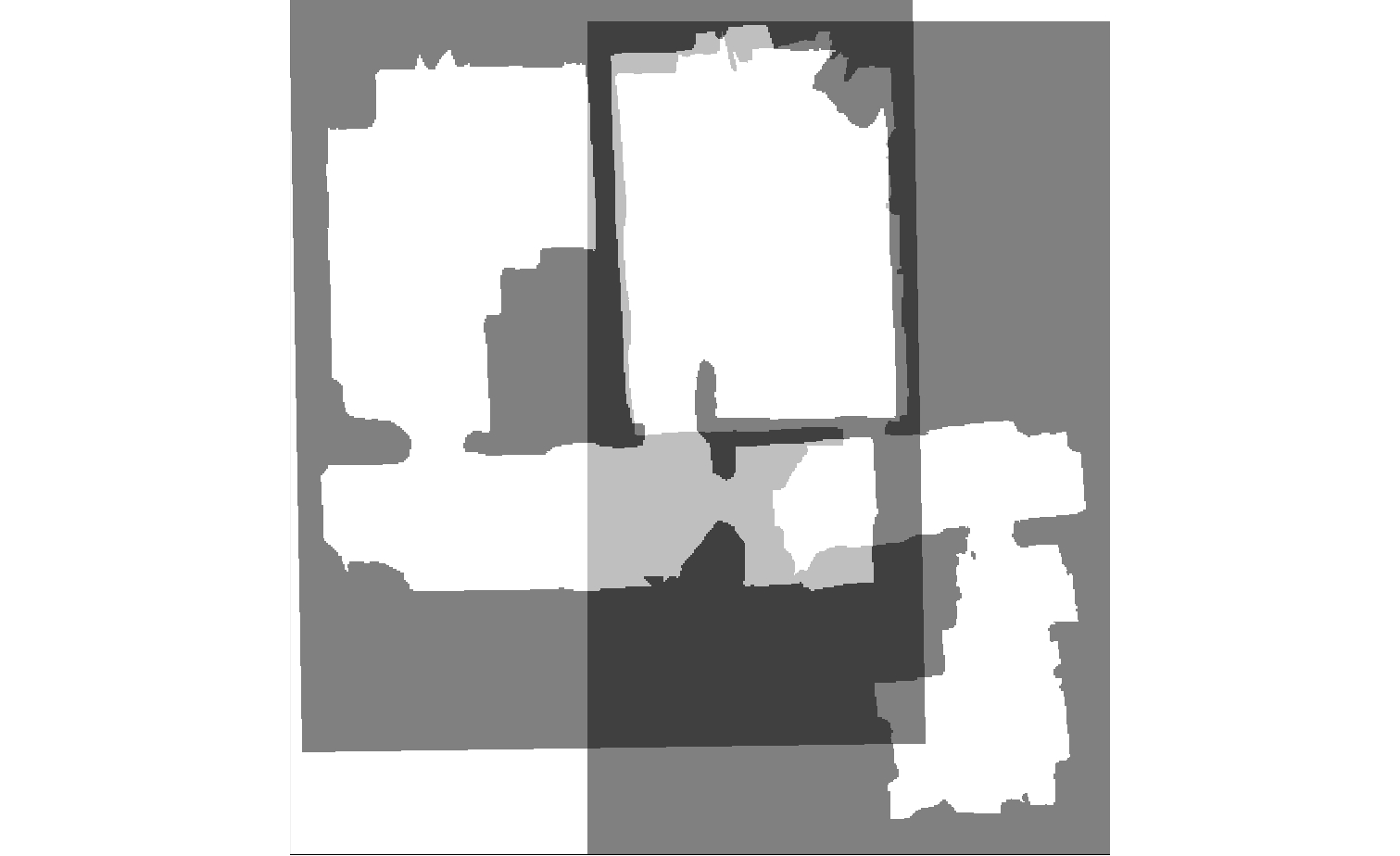}
    \caption{wrong}
    \label{subfig:wrong_alignment}
  \end{subfigure}%
  \caption[xxx]{
    Examples of correct, defected, and wrong alignments.
    Both correct and defected are considered as successful alignments.
  } \label{fig:alignment_correctness}
\end{figure}

%%%%%%%%%%%%%%%%%%%%%%%%%%%%%%%%%%%%%%%%
\paragraph{Evaluations are based on success rate}
The performance of each method is provided as success rate, which is a percentage of successful alignments.
We skip a \emph{quantified accuracy measure} for the alignment.
It proved very hard (impossible for our data) to provide a per map \emph{alignment accuracy}, due to:
\begin{inparaenum}[i)]
\item the lack of ground truth for the sensor maps,
\item the inaccuracy of layout maps, and
\item the presence of noise and global inconsistency of the sensor maps.
\end{inparaenum}
Figure~\ref{fig:alignment_correctness} illustrates our quality assessment of the alignments.

%%%%%%%%%%%%%%%%%%%%%%%%%%%%%%%%%%%%%%%%%%%%%%%%%%%%%%%%%%%%%%%%%%%%%%%%%%%%%%%%
\subsection{Experiments and results} \label{subsec:experiments}
The performance of the method is evaluated under three different experimental setups:
\begin{itemize}
\item \emph{sensor map to layout map alignment}, which is the main objective of the proposed method.
\item \emph{sensor map to sensor map alignment}, where we observe how partial coverage, noise and inconsistency of sensor maps affect the performance.  
\item \emph{evaluation of alignment match score}, where the match score is studied for the alignments of intra and inter environment maps.
  Accordingly every sensor map is aligned to all other layout maps, whether from the same environment or not.
\end{itemize}

%%%%%%%%%%%%%%%%%%%%%%%%%%%%%%%%%%%%%%%%
\subsubsection{Sensor map to layout map alignment}

Table~\ref{tab:res_layout_sensor} presents the performance of the method in aligning sensor maps to layout maps (within the same environment).
The column \emph{initial} represents the number of initial estimated transformations,
\emph{after rejection} represents the number of remaining hypotheses after rejecting non-uniformly scaled transformations ($\sim 90\%$ are rejected),
and the last column marks the success of each alignment.
In total, the method has successfully aligned all maps of the home environments, and yielded $~83\%$ in success rate for the office buildings.
According to our investigations, failures are mainly due to the violation of the prior assumptions, such as global inconsistency and not enough segment-able regions in sensor maps.

\begin{table}
  \centering
  \begin{tabular}{c | c | c | c }
  %% \begin{tabular}{c | m{2cm} | m{1cm} | c }
    & \multicolumn{2}{c|}{number of hypotheses} & \\
    map & initial & after rejection & result\\
    \hline
    HIH\_01 & 24 & 6 & \checkmark \\
    HIH\_02 & 48 & 8 & \checkmark \\
    HIH\_03 & 24 & 6 & \checkmark \\
    HIH\_04 & 36 & 4 & \checkmark \\
    \hline
    KPT\_01 & 256 & 14 & \checkmark\\
    KPT\_02 & 128 & 10 & \checkmark\\
    KPT\_03 & 144 & 10 & \checkmark\\
    KPT\_01 & 160 & 10 & \checkmark\\
    \hline
    HH\_E5\_01 & 9152 & 726 & \checkmark\\
    HH\_E5\_02 & 5280 & 458 & $\times$\\
    HH\_E5\_03 & 6688 & 704 & $\times$\\
    HH\_E5\_04 & 5984  & 458  & $\times$\\
    HH\_E5\_05 & 5808  & 368  & \checkmark\\
    HH\_E5\_06 & 2992 & 178 & \checkmark\\
    HH\_E5\_07 & 3696  & 214 & \checkmark\\
    HH\_E5\_08 & 4400 & 352 & $\times$\\
    HH\_E5\_09 & 9152  & 616  & \checkmark\\
    HH\_E5\_10 & 9152 & 794 & \checkmark\\
    HH\_E5\_11 & 5808 & 470 & \checkmark\\
    HH\_E5\_12 & 8624 & 644 & \checkmark\\
    HH\_E5\_13 & 4928 & 336 & $\times$\\
    HH\_E5\_14 & 3344 & 328 & \checkmark\\
    \hline
    HH\_F5\_01 & 1292 & 128 & \checkmark\\
    HH\_F5\_02 & 1088 & 82 &  \checkmark\\
    HH\_F5\_03 & 816 & 86 & \checkmark\\
    HH\_F5\_04 & 680 & 70 & \checkmark\\
    HH\_F5\_05 & 544 & 56 & \checkmark\\
    HH\_F5\_06 & 408 & 28 & \checkmark\\
    HH\_F5\_07 & 476 & 26 & \checkmark\\
    HH\_F5\_08 & 3604 & 158 & \checkmark\\
    HH\_F5\_09 & 952 & 78 & \checkmark\\
    HH\_F5\_10 & 680 & 56 & $\times$\\
    HH\_F5\_11 & 680 & 264 & \checkmark\\
    HH\_F5\_12 & 680 & 46 & \checkmark\\
    HH\_F5\_13 & 1020 & 92 & \checkmark\\
    HH\_F5\_14 & 1088 & 158 & \checkmark\\
  \end{tabular}
  \caption[xxx]{
    Performance of the method in aligning sensor maps to layout maps.
    The column \emph{initial} represents the number of initial estimated transformations,
    \emph{after rejection} represents the number of remaining hypotheses after rejecting non-uniformly scaled transformations ($\sim 90\%$ are rejected),
    and the last column marks the success of each alignment.
  }
  \label{tab:res_layout_sensor}
\end{table}

%%%%%%%%%%%%%%%%%%%%%%%%%%%%%%%%%%%%%%%%
\subsubsection{Sensor map to sensor map alignment}
Table~\ref{tab:success_rate} compares the success rate of the method in aligning sensor maps to sensor maps, versus aligning sensor maps to layout maps.
It can be observed that the success rate of the method drops in aligning sensor maps to sensor maps.
There are two main reasons for this drop;
\begin{inparaenum}[i)]
\item many sensor maps are partial and consequently they overlap with each other marginally,
\item the violation of initial assumptions.
\end{inparaenum}
In the presence of layout map there is one source of noise and global inconsistency, but in case of aligning two sensor maps the noise and inconsistencies are amplified.

\begin{table}
  \centering
  \begin{tabular}{c | c | c}
    environment & sensor vs layout & sensor vs sensor\\% & sensor map vs sensor map (discarding non overlapping)
    \hline
    HH\_E5 & $64.28\%$ $(\sfrac{9}{14})$ & $50.54\%$ $(\sfrac{46}{91})$\\
    HH\_F5 & $92.85\%$ $(\sfrac{13}{14})$ & $68.13\%$ $(\sfrac{62}{91})$\\
    HIH & $100\%$ $(\sfrac{4}{4})$ & $100\%$ $(\sfrac{6}{6})$\\
    KPT & $100\%$ $(\sfrac{4}{4})$ & $83.33\%$ $(\sfrac{5}{6})$\\
  \end{tabular}
  \caption[xxx]{The success rate of the method in aligning sensor maps to sensor maps, versus aligning sensor maps to layout maps.}
  \label{tab:success_rate}
\end{table}

%%%%%%%%%%%%%%%%%%%%%%%%%%%%%%%%%%%%%%%%
\subsubsection{Evaluation of the alignment match score}
\begin{figure}%[!ht]
  \centering
  \includegraphics[width=\linewidth]{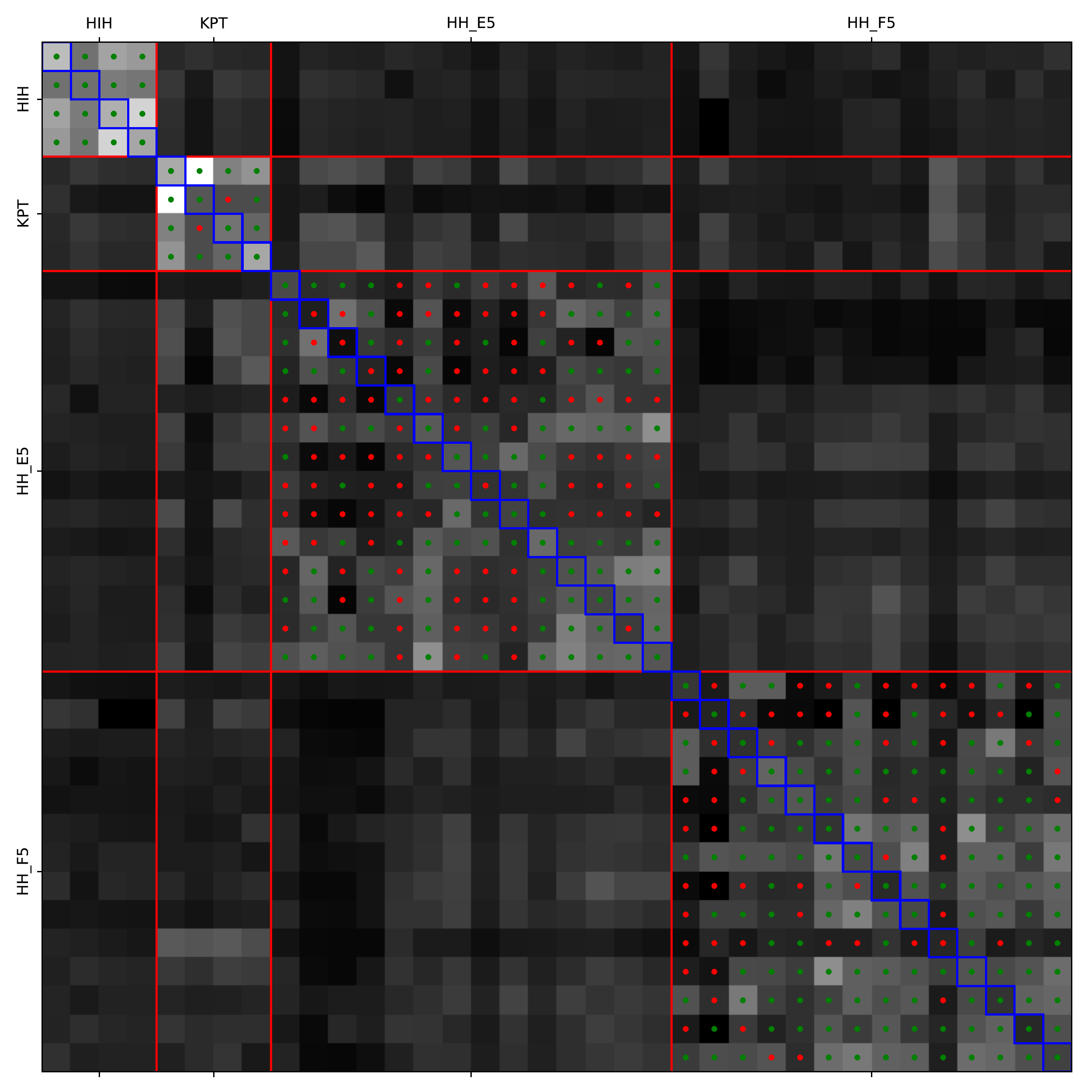}
  \caption[xxx]{
    The match score of the \emph{winning hypotheses} for all pairs of sensor maps (includes pairing sensor maps of different environment).
    Gray-scale encodes the value of the alignment match score ($0\leq S_A \leq 1$).
    The cells on diagonal (marked with blue borders) represent the alignment of sensor maps versus the layout maps, and the red lines separate different environments.
    Green and red dots mark the success and failure of the alignments respectively.
  } \label{fig:match_score_matrix}
\end{figure}

Figure~\ref{fig:match_score_matrix} represents the match score of the \emph{winning hypotheses} for all pairs of sensor maps (includes pairing sensor maps of different environment).
Gray-scale encodes the value of the match score ($0\leq S_A \leq 1$).
The cells on diagonal (marked with blue borders) represent the alignment of sensor maps versus the layout maps, and the red lines separate different environments.
Green and red dots mark the success and failure of the alignments respectively.
The squares on diagonal, corresponding to intra environment alignments, are slightly brighter compared to the rest of the matrix which corresponds to inter environment alignments.
However, this is not conclusive enough to employ this measure across different environments and to identify a layout map to which a sensor map belongs.
Under scrutiny it can be seen that maps of a smaller environment (KPT) align with a maps of a bigger environment (HH\_E5) with a high score.
Also, some maps of the same environment have low match score due to the small overlap, even though some are successfully aligned.

\begin{figure*}%[!ht]
  \centering
  \includegraphics[width=\linewidth]{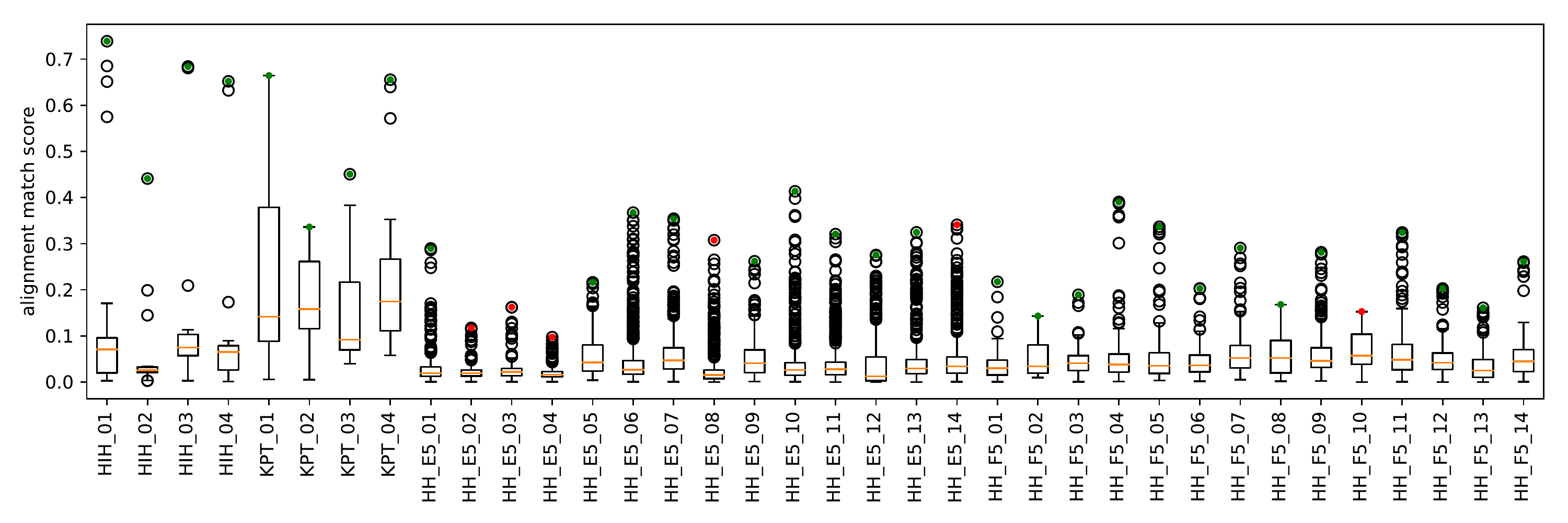}
  \caption[xxx]{
    The alignment match score for \emph{all hypotheses} in sensor maps to layout maps alignments.
    The winning hypotheses are marked red and green, representing the failure and success of each alignment.
  } \label{fig:match_score_boxplot}
\end{figure*}

Figure~\ref{fig:match_score_boxplot} presents a box plot of the alignment match score for \emph{all hypotheses} in aligning sensor maps to layout maps.
The winning hypotheses are marked red and green, representing the failure and success of each alignment.
There seems to be a cut-off point on the match score value across all maps ($\sim$0.15), which separates successful alignments from failures.
However there is no reliable margin to this cut-off to be used as a threshold between success and failure.
The take away message here is that the value of match score is not a reliable indicator of the alignment success.

In conclusion we can say, even though the \emph{match score} has proven useful in selecting the winning alignment among all hypotheses, yet it is not conclusively reliable to detect to which layout map a sensor map belongs, nor to autonomously detect a successful alignment.

%%%%%%%%%%%%%%%%%%%%%%%%%%%%%%%%%%%%%%%%%%%%%%%%%%%%%%%%%%%%%%%%%%%%%%%%%%%%%%%%
\subsection{Comparison with other methods} \label{sec:comparison}
Our initial investigation and experiments towards map alignment lead to methods which we categorize into two groups.
First the generic approaches in data association, such as image alignment, image registration and point set registration.
And second the map alignment methods, such as Hough transform-based algorithms.
%% Here we present the performances of these methods, and leave an in-depth review of these methods to Section~\ref{sec:related}.
The performance of all methods in terms of success rate is available in Table~\ref{tab:success_rate_comparison}.
And finally, a brief account of computational costs, presented in Table.~\ref{tab:computation_time_comparison}, will follow the performance evaluation.

%%%%%%%%%%%%%%%%%%%%%%%%%%%%%%%%%%%%%%%%
\subsubsection{Generic data association methods} \label{subsubsec:generic}

Here the performances of three generic data association methods in map alignment are presented:
\begin{inparaenum}[i)]
\item image alignment with Enhanced Correlation Coefficient Maximization (ECC) \cite{evangelidis2008parametric},
\item image registration with Scale-Invariant Feature Transform (SIFT) \cite{lowe1999sift}, and
\item point set registration with Coherent Point Drift (CPD) \cite{NIPS2006_2962}, \cite{5432191}.
\end{inparaenum}
All the performance results are available in Table~\ref{tab:success_rate_comparison}.
We observed that the methods based on ECC and SIFT perform slightly better when they are applied to the distance transform of the maps, instead of the occupancy maps.
Accordingly the presented results are based on distance images of the maps.

%%%%%%%%%%
ECC maximization performs worse on aligning sensor maps to layouts, due to its higher sensitivity to data-level similarity.
A detailed review of the failed cases in aligning sensor maps to sensor maps reveals that the main causes of failure are the global inconsistencies of the sensor maps and small overlaps between the maps.

%%%%%%%%%%
Image registration with SIFT~\cite{lowe1999sift} was tested in combination with Fast Approximate Nearest Neighbors \cite{muja2009fast} for feature matching.
This method works best on maps with unique patterns represented by a unique constellation of ``key points'', and consequently has a slightly better performance on bigger maps with more key points.
Although the data-level similarity between sensor maps is in favor of resulting more similar features, however, this method yields better results in aligning sensor maps to layouts of bigger environments thanks to higher overlaps.

%%%%%%%%%%
For the experiments with CPD, point sets have been generated from the occupied cells of the maps.
CPD is superior to Iterative Closest Point (ICP) as it supports affine transformations.
However, it is computationally expensive and memory demanding, so that the original point sets had to be sub-sampled.
This in turn conceals the structural patterns of the maps, and becomes more sensitive to local minima.

%%%%%%%%%%%%%%%%%%%%%%%%%%%%%%%%%%%%%%%%
\subsubsection{Map alignment methods} 
We have chose the works by Carpin~\cite{carpin2008fast} and Saeedi et al. (PGVD) \cite{saeedi2012efficient} as the representatives of this category.
Implementations of both methods are made publicly available by the authors.
The performances of these methods are presented in Table~\ref{tab:success_rate_comparison}.

%%%%%%%%%%
One interesting aspect of these methods is their independence from the assumption of maps' ``segment-able regions''.
Therefore they could be considered to have a broader target applications.
For instance, we have observed that such methods perform better on maps that mostly contain corridors, which is a challenge for the region segmentation phase of our method.
Also, thanks to the underlying decoupling of rotation and translation estimation, they could be relatively faster than other methods, specifically the method proposed by Carpin~\cite{carpin2008fast}.
However, these advantages come with a price in performance, while these methods perform better on particular cases, they do have a lower overall success rate over our collection of maps.
By inspecting individual results, we observed that many of the failures were due to a wrong orientation alignment.
And many of those cases which survived the orientation estimation, they still failed at the translation estimation.
Fundamentally, these methods exploit the structural similarities in maps, by finding similarity is Hough spectra and cross correlating the maps after orientation alignment.
We believe the noise, the global inconsistencies, and the repetitive patterns of our maps are the top challenges for such methods.

%%%%%%%%%% scale mismatch
These methods are limited to rigid transformation, and as a result they could solve only the alignment of sensor to sensor map.
Therefore, we manually adjusted scales of the layout maps, so that these methods could be evaluated over the alignments of sensor to layout maps.
These results (manually adjusted scales) are marked with asterisks in Table~\ref{tab:success_rate_comparison}.
They score very low, which we believe is due to the significant disparity in the representations (i.e. different modalities).
When contrasted with our method, this is an interesting result.
Compared to the alignment of sensor to layout maps, our method scores lower when both maps are sensor maps.
This is mainly due to the amplification of noise, inconsistency and partial coverage when both maps are sensor maps.
On the other hand, these methods (\cite{carpin2008fast}, \cite{saeedi2012efficient}) perform worse when aligning sensor maps to layouts, due to their sensitivity to representation disparity.

On a final note, it is important to note that due to a lack of proper insight to the implementations of these methods, we could not fine-tune them, to maximize their performances in the setting of our experiments.
Therefore we would like to point out, that the success rates of the methods presented in Table~\ref{tab:success_rate_comparison} might not represent their best performances, but rather they provide an insight into advantages and drawbacks of each method.

\begin{table*}
  \begin{minipage}{\textwidth}
    \centering
    \begin{tabular}{l|ccccc|ccccc}
      ~ &
      \multicolumn{5}{c}{sensor to layout} &
      \multicolumn{5}{c}{sensor to sensor}\\
      method &
      HH\_E5 & HH\_F5 & HIH & KPT & total &
      HH\_E5 & HH\_F5 & HIH & KPT & total\\
      \hline
      ECC maximization~\cite{evangelidis2008parametric} &
      $0.0$ & $7.14$ & $0.0$ & $0.0$ & $2.77$ & % (1/36)
      $37.36$ & $29.67$ & $0.0$ & $16.67$ & $31.9$ \\  % (62/194)
      
      SIFT~\cite{lowe1999sift} &
      $21.43$ & $50$ & $0.0$ & $0.0$ & $27.7$ & % (10/36)
      $17.58$ & $28.57$ & $16.67$ & $0.0$ & $22.1$\\  % (43/194)
      
      CPD~\cite{NIPS2006_2962}~\cite{5432191} & % (0/36)
      $0.0$ & $0.0$ & $0.0$ & $0.0$ & $0.0$ &  % (11/194)
      $8.79$ & $3.3$ & $0.0$ & $0.0$ & $5.6$\\
      
      Saeedi et al. (PGVD)~\cite{saeedi2012efficient} &
      $0.0^*$ & $0.0^*$ & $25^*$ & $0.0^*$ & $2.77^*$ & % (1/36)
      $10.98$ & $12.08$ & $33.33$ & $16.66$ & $12.4$\\  % (24/194)
      
      Carpin~\cite{carpin2008fast} &
      $0.0^*$ & $7.14^*$ & $0.0^*$ & $0.0^*$ & $2.77^*$ & % (1/36)
      $16.48$ & $29.67$ & $100$ & $83.33$ & $27.31$\\  % (53/194)
      
      our method & 
      $64.28$ & $92.85$ & $100$ & $100$ & $83.3$ & % (30/36)
      $50.54$ & $68.13$ & $100$ & $83.33$ & $66.5$\\  % (129/194)
      
    \end{tabular}
    \caption[xxx]{
      Success rates (in $\%$) of different methods on map alignment.
      Numbers marked with an asterisk (*) refer to the methods that are not able to handle scaling.
      In those cases, the experiments were performed on manually scaled maps.}
    \label{tab:success_rate_comparison}
  \end{minipage}
\end{table*}

%%%%%%%%%%%%%%%%%%%%%%%%%%%%%%%%%%%%%%%%
\subsubsection{Computation time}
The timings of all methods are provided in Table~\ref{tab:computation_time_comparison}.
All the experiments were carried out on a computer with an Intel$^\circledR$ Core\texttrademark~i5-3340M CPU @ 2.70GHz $\times$4, and 8GiB SODIMM DDR3 Synchronous 1600 MHz of memory, running Ubuntu 14.04.
The timings of experiments are separated into home and office building, which provides a sense of methods' scalability with respect to the size of maps.
The average map size for home environments is $2.2 \cdot 10^5$ pixels, and it is $1.0 \cdot 10^6$ pixels for office buildings (roughly 5 times bigger).

Since CPD is expensive and not scalable, the original point sets were reduced from $1.2 \cdot 10^4$ points on average in small maps and $3.3 \cdot 10^4$ points in bigger maps, to $500$ (close to memory limit of the algorithm on our hardware.)
Therefore a meaningful computation time could not be provided here.

In comparison, our method falls behind some other approaches in terms of computational cost.
Specifically, those methods designed for real-time applications such as Carpin's method~\cite{carpin2008fast} for multi-robot mapping, are extremely fast and hard to beat.
Our method is based on the decomposition of the space and requires an interpretation through abstract models, which is in general computationally more expensive than signal based interpretations such as Hough-spectra.
However, if one intends to exploit the fast speed of the Hough transform-based methods in combination with our method, there is a trade-off between thoroughness of the hypotheses generation and computational time.
In conclusion we speculated that, under certain assumptions (such as orthogonally structured environments), one can create a set of constraints imposed on the hypotheses generation to narrow down the search space.
Although, a better understanding of such potential combination requires further development and more experiments.

At the end, we would like to emphasize that the timings of each method provided here can portray a rough scale, and should not be taken as an accurate computational cost comparison.
This is mainly due to the heterogeneity of the implementations (C++, Python, Matlab).
Furthermore, some of the algorithms are borrowed from other context (e.g. CPD, ECC) and applied to map alignment problem.
Some are intended for offline applications with not much concern for computational time, while others were specifically designed to be fast for real-time applications.
As a result, these computation times are not sufficient to generalize on the performance of each approach.

\begin{table*}
  \begin{minipage}{\textwidth}
    \centering
    \begin{tabular}{l|c|c c}
      & & \multicolumn{2}{c}{time in seconds}\\
      method & implementation & home & office \\
      \hline
      ECC maximization~\cite{evangelidis2008parametric} & Python \& C++ & $32.79 (28.24)$ & $73.46 (85.46)$ \\
      SIFT~\cite{lowe1999sift} & Python \& C++ & $0.20 (0.05)$ & $0.67 (0.14)$ \\
      Saeedi et al. (PGVD)~\cite{saeedi2012efficient} & Matlab & $4.91 (1.42)$ & $50.20 (19.84)$ \\
      Carpin~\cite{carpin2008fast} & C++ & $3.07 \cdot 10^{-4}(9.28 \cdot 10^{-5})$ & $2.65 \cdot 10^{-4}(6.72 \cdot 10^{-5})$ \\
      our method & Python & $8.86 (2.13)$ & $41.86 (41.92)$
    \end{tabular}
    \caption[xxx]{Average (and standard deviation) of the computation times (in seconds) of different methods, separated to home and office environments.
    }
    \label{tab:computation_time_comparison}
    \end{minipage}
\end{table*}

%%%%%%%%%%%%%%%%%%%%%%%%%%%%%%%%%%%%%%%%%%%%%%%%%%%%%%%%%%%%%%%%%%%%%%%%%%%%%%%%
%%%%%%%%%%%%%%%%%%%%%%%%%%%%%%%%%%%%%%%%%%%%%%%%%%%%%%%%%%%%%%%%%%%%%%%%%%%%%%%%
%%%%%%%%%%%%%%%%%%%%%%%%%%%%%%%%%%%%%%%%%%%%%%%%%%%%%%%%%%%%%%%%%%%%%%%%%%%%%%%%
\section{Conclusion} \label{sec:conclusion}
In this paper, we present our work and findings on solving the map alignment problem, for 2D spatial maps.
Many interesting approaches have been proposed to address this problem.
However, existing algorithms hinge on assumptions that are not valid in (a number of) interesting use cases, such as aligning partial maps of different modalities.
Most often they are designed to perform map merging where maps are from similar modalities, hence they rely on sensor level similarity of the input maps, and consequently are sensitive to noise and inconsistencies of sensor maps.
In addition, maps of the same modality have similar scale, and as a result, such methods are limited to \emph{rigid transformations}.
Such assumptions do not hold where maps of different modalities, such as sensor maps and layout maps, are to be aligned.
Also, the scaling from one map to the other adds a new dimension to the search space and the desired solution becomes a \emph{similarity transformation} rather than a rigid transformation.

We have shown, with experimental results, the insufficiency of generic data association methods (e.g. SIFT, ECC), and some map alignment methods (designed for aligning maps of same modalities) in solving the problem in our experimental setup.
We have compared the performance of our method with that of other methods both for sensor to sensor map alignment and sensor to layout map alignment.
Except for few examples of similar performance, our method outperforms other methods.
In aligning sensor to sensor maps, we observed that the presence of noise and global inconsistency has been the main challenge for most other approaches.
The representation disparity between maps of different modalities has been even more challenging for those methods in aligning sensor to layout maps.
For the latter experiment, the layout maps were manually scaled to match the sensor maps in size, since other map alignment methods are limited to rigid transformation.
Our method relies on the notion that most human built environments are composed of regions.
Accordingly, our method finds the correct alignment by associating regions and selecting the best hypothesis among all candidates.
By exploiting the notion of regions and founding our method on spatial decomposition, our alignment method operates on a higher level of abstraction.
As a consequence, the method is more robust to dissimilarity and heterogeneity of the sensor-level data.
Furthermore, the approach of aligning regions rather that associating sensor-level data enables our method to handle the scaling factor like any other transformation parameter.

%%%%%%%%%%%%%%%%%%%%%%%%%%%%%%%%%%%%%%%%%%%%%%%%%%%%%%%%%%%%%%%%%%%%%%%%%%%%%%%%
\subsection{Discussion} \label{subsec:disc}
In the result section we tried to provide a thorough performance comparison between our proposed method and other approaches to solve the map alignment problem.
We do not claim, or believe, that our method is superior to other approaches in a generic problem formulation of data association and map alignment.
Rather, we tend to emphasize the particular characteristics and advantages that this method offers over alternatives in specific challenges, namely aligning maps from different modality, severe data level noise, and maps of different scales.
However, there might be some other objectives close to the core of the map alignment problem that our method falls short of.
Examples of such applications are, aligning maps of unstructured environment and real-time applications.

%%%%%%%%%%%%%%%%%%%%%%%%%%%%%%%%%%%%%%%%
\paragraph{Advantages}
Apart from the higher success rate of our proposed method, we would like to point out some other interesting features of it.
One important aspect, and one of the main motivations behind this work, is the ability to align maps of different modality, and specifically sensor maps to layout maps.
As stated earlier, such a task demands a method that is indifferent to heterogeneity and different scales of input maps.
Our proposed method shows a considerable performance for such cases (success rate $83.3\%$ compared to the best alternative $27.7\%$).
We have developed a region segmentation method based on the arrangement representation and distance transform, but the general framework of our alignment method is not dependent on any specific region segmentation technique.
Our decomposition based algorithm would be able to find the alignment as long as the input maps are effectively interpreted by the arrangement of the 2D plane.
That is to say, as long as the input maps are spatial and could be segmented into meaningful regions, the proposed method in this work could be employed to find the alignment.
We speculate that an improved region segmentation will have a positive effect on the performance of this alignment approach.
It is worth mentioning that the implementation of our proposed method, and the accompanied experiments presented in this paper, convert both maps to occupancy-like bitmaps in advance.
However it is not a requirement of the proposed alignment algorithm, but rather it was a convenient choice.
And finally, the intermediate representation that is constructed for alignment, by itself is a useful representation for different objectives~\cite{shahbandi2015semi},~\cite{shahbandi2014sensor}, and it is not alignment-specific.

%%%%%%%%%%%%%%%%%%%%%%%%%%%%%%%%%%%%%%%%
\paragraph{Drawbacks and limitations}
The main disadvantage of the proposed method is the computation time.
This means that this method is not suitable for real-time applications.
While exploiting the notion of \emph{meaningful} regions improves the map alignment under difficult circumstances, it also limits the applicability of the method.
Dependency on the region segmentation means it most likely will fail in maps of environments cluttered with furniture, or in a maze-like environment, unless an appropriate region segmentation algorithm is employed.
Partial maps which don't cover multiple regions (e.g. a map of only one room), in applications such as scan matching and incremental mapping, violate one of the initial assumptions and would cause our method to fail.
We speculate that this is a domain where other methods such as the ones proposed by Carpin \cite{carpin2008fast} and Saeedi et al. \cite{saeedi2012efficient} would outperform our method, given the maps are from the same modality.
As stated before, in Section~\ref{sec:results}, not all the maps satisfy our initial assumptions such as global consistency.
We included these maps to better explain the effects of aforementioned assumptions on the method and portray a fair picture of the method's performance under different conditions, even if they violate the assumptions of our method.
Other conditions that make our method unsuitable occurs when the prior assumptions are violated.
Examples are non-uniformly scaled maps like sketch maps, and maze-like environments such as underground tunnels and alike where the notion of meaningful regions might not apply.

%%%%%%%%%%%%%%%%%%%%%%%%%%%%%%%%%%%%%%%%
\paragraph{Model regression}
Random sample consensus (RANSAC) is a powerful regression technique in estimating a model from noisy data.
However, our empirical observation suggests that RANSAC is not a suitable replacement for the components of our method.
The first possibility is to employ RANSAC for hypothesis generation, i.e. estimating a transformation between faces with known correspondences.
We have found Umeyama's method to be a better fit as a non-iterative deterministic method for this objective.
Alternatively RANSAC could be considered for solving the alignment on the map level with unknown correspondences.
However, the point sets from our representation (vertices of the prime graph) are sparse and do not reflect the skeletal structure.
This challenge is exaggerated with relatively high level of noise and partiality of the maps.
We have experimented with RANSAC in this manner, with a simple setup and a denser sampling of the occupied points, the result of which has not been satisfactory.
This lead to our experimentation with Iterative Closest Points (ICP) and Coherent Point Drift (CPD), a continuation of the attempt in relying on the shape of the distribution of occupied points.
The results of CPD have been included in this manuscript as a representative of this category of approaches.
Despite the inadequacy of RANSAC in estimating the alignment, we speculate that such regression techniques could be beneficial in estimating other models as a part of a more elaborate method.
For instance, RANSAC can be used for the regression of the \emph{geometric coherency of hypotheses}.
That is to say, assuming a correct alignment is represented with multiple hypotheses, the pool of hypotheses is expected to contain clusters of similar transformations.
RANSAC can be used for estimating the geometric coherency of hypotheses and rejecting outliers.
We ran experiments with this idea, although with a clustering algorithm (DBSCAN~\cite{ester1996density}) and not RANSAC.
The challenge is that not always the correct alignment has multiple representatives, specially for small, deformed, and partial maps.
This idea needs further investigation, since treating the pool of hypotheses has to be done carefully with additional considerations.

%%%%%%%%%%%%%%%%%%%%%%%%%%%%%%%%%%%%%%%%%%%%%%%%%%%%%%%%%%%%%%%%%%%%%%%%%%%%%%%%
\subsection{Future work} \label{subsec:future}
In the continuation of this work we intend to address some interesting questions which were raised during the development of this work.
One of those questions is the challenge of autonomous detection of successful alignments.
This problem can be translated to a classification task, where an \emph{alignment match score} could be a multidimensional vector based on other sources of information in addition to arrangement based match score, such as graph matching metrics (e.g. GED), and data level distance between maps.
Towards that objective, we intended to enrich our collection of maps with a wider variety of environments.
Furthermore, we intend to carry out more challenging experiments and with other modalities to inspect the performance of the proposed alignment approach under different circumstances.

The direction of our future work is towards merging maps after alignment.
Specific examples of features to contain in a merging process would be the transferring of semantic labels from layout map to sensor map for high level task planning, and detecting and compensating global inconsistencies in sensor map by relying on the structure of the layout map.

%%%%%%%%%%%%%%%%%%%%%%%%%%%%%%%%%%%%%%%%%%%%%%%%%%%%%%%%%%%%%%%%%%%%%%%%%%%%%%%%
%%%%%%%%%%%%%%%%%%%%%%%%%%%%%%%%%%%%%%%%%%%%%%%%%%%%%%%%%%%%%%%%%%%%%%%%%%%%%%%%
%%%%%%%%%%%%%%%%%%%%%%%%%%%%%%%%%%%%%%%%%%%%%%%%%%%%%%%%%%%%%%%%%%%%%%%%%%%%%%%%
\begin{acknowledgements}
  This work was funded by the Swedish Knowledge Foundation (KK-Stiftelsen), and the European Union's Horizon 2020 research and innovation programme under grant agreement No 732737 (ILIAD).
  The authors would like to express their gratitude to Dr. Karl Iagnemma and the anonymous reviewers who helped us improve the quality of the manuscript with their valuable and constructive inputs.
\end{acknowledgements}

%%%%%%%%%%%%%%%%%%%%%%%%%%%%%%%%%%%%%%%%%%%%%%%%%%%%%%%%%%%%%%%%%%%%%%%%%%%%%%%%
%%%%%%%%%%%%%%%%%%%%%%%%%%%%%%%%%%%%%%%%%%%%%%%%%%%%%%%%%%%%%%%%%%%%%%%%%%%%%%%%
%%%%%%%%%%%%%%%%%%%%%%%%%%%%%%%%%%%%%%%%%%%%%%%%%%%%%%%%%%%%%%%%%%%%%%%%%%%%%%%%
\bibliographystyle{plain}
\bibliography{myref}

%%%%%%%%%%%%%%%%%%%%%%%%%%%%%%%%%%%%%%%%%%%%%%%%%%%%%%%%%%%%%%%%%%%%%%%%%%%%%%%%
%%%%%%%%%%%%%%%%%%%%%%%%%%%%%%%%%%%%%%%%%%%%%%%%%%%%%%%%%%%%%%%%%%%%%%%%%%%%%%%%
%%%%%%%%%%%%%%%%%%%%%%%%%%%%%%%%%%%%%%%%%%%%%%%%%%%%%%%%%%%%%%%%%%%%%%%%%%%%%%%%
\appendix
\section{Map Data-Set} \label{app:datasets}
\clearpage

%% \newpage
%%%%%%%%%%%%%%%%%%%%%%%%%%%%%% E5
\begin{figure}%[!ht]
  \centering
  \begin{subfigure}{.33\linewidth}
    \includegraphics[width=\linewidth]{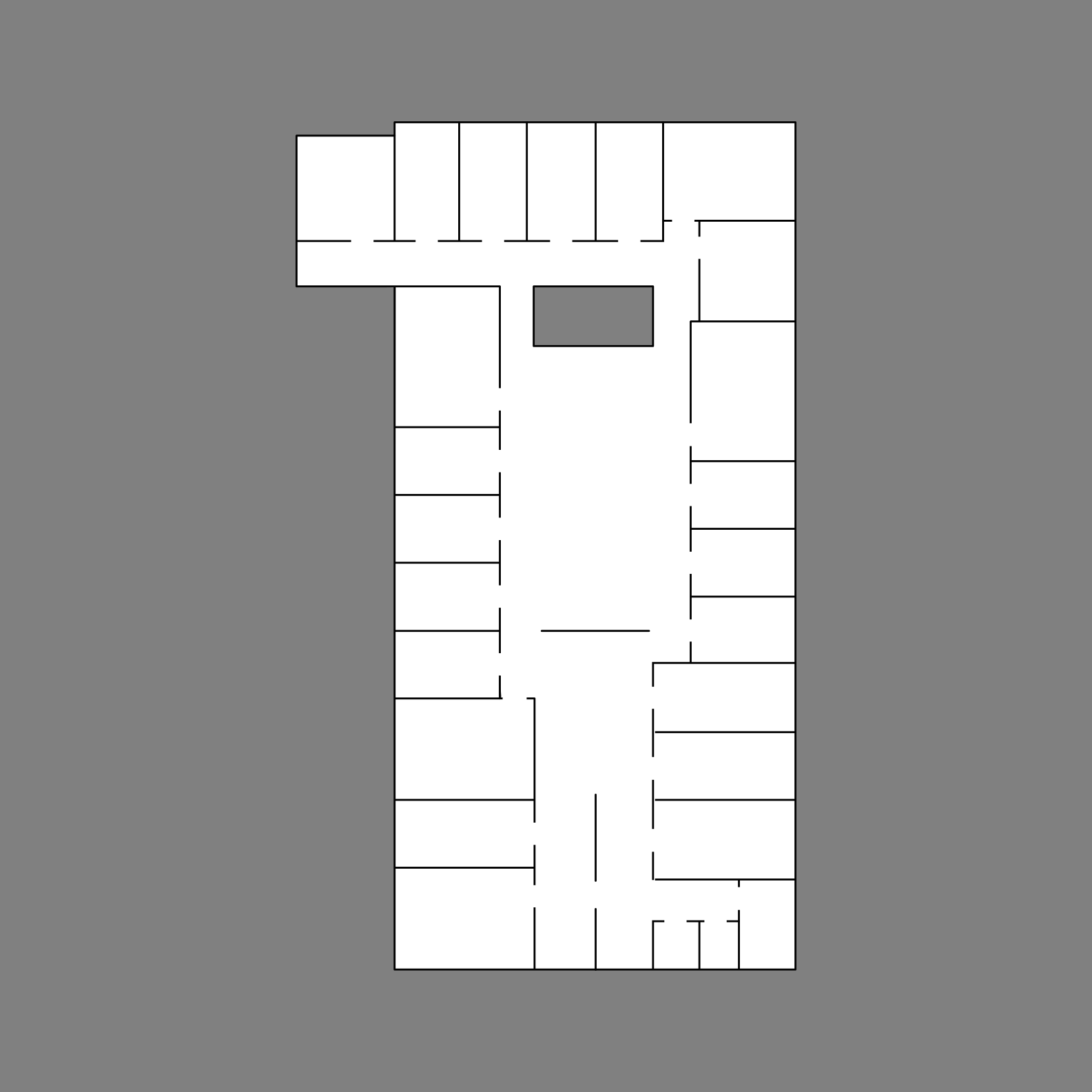}
    % \caption[]{layout}%% \label{}
  \end{subfigure}%
  ~%
  \begin{subfigure}{.33\linewidth}
    \includegraphics[width=\linewidth]{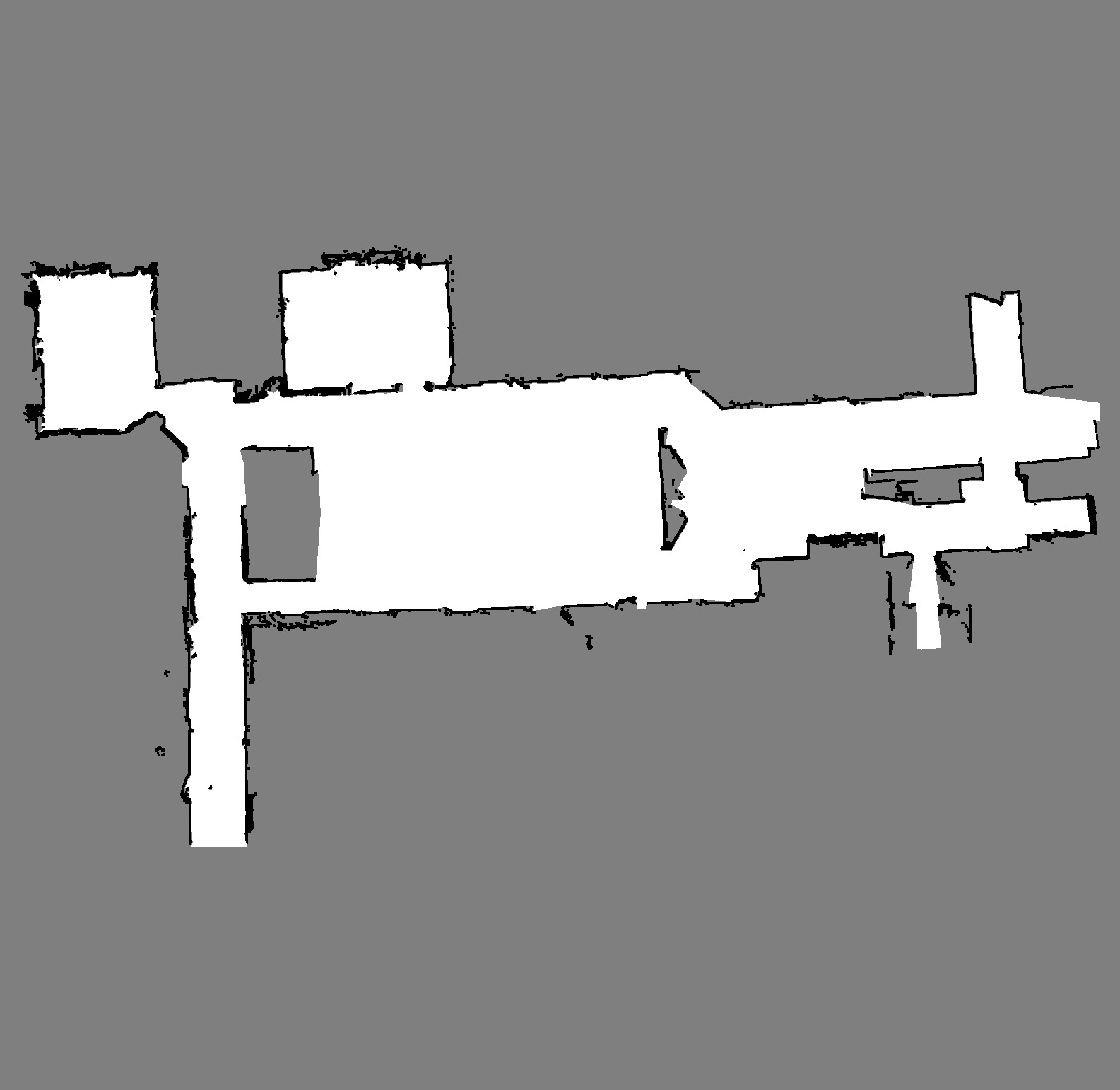}%_20170131150415.jpg}
    % \caption[]{sensor map 1}%% \label{xxx}
  \end{subfigure}%
  ~%
  \begin{subfigure}{.33\linewidth}
    \includegraphics[width=\linewidth]{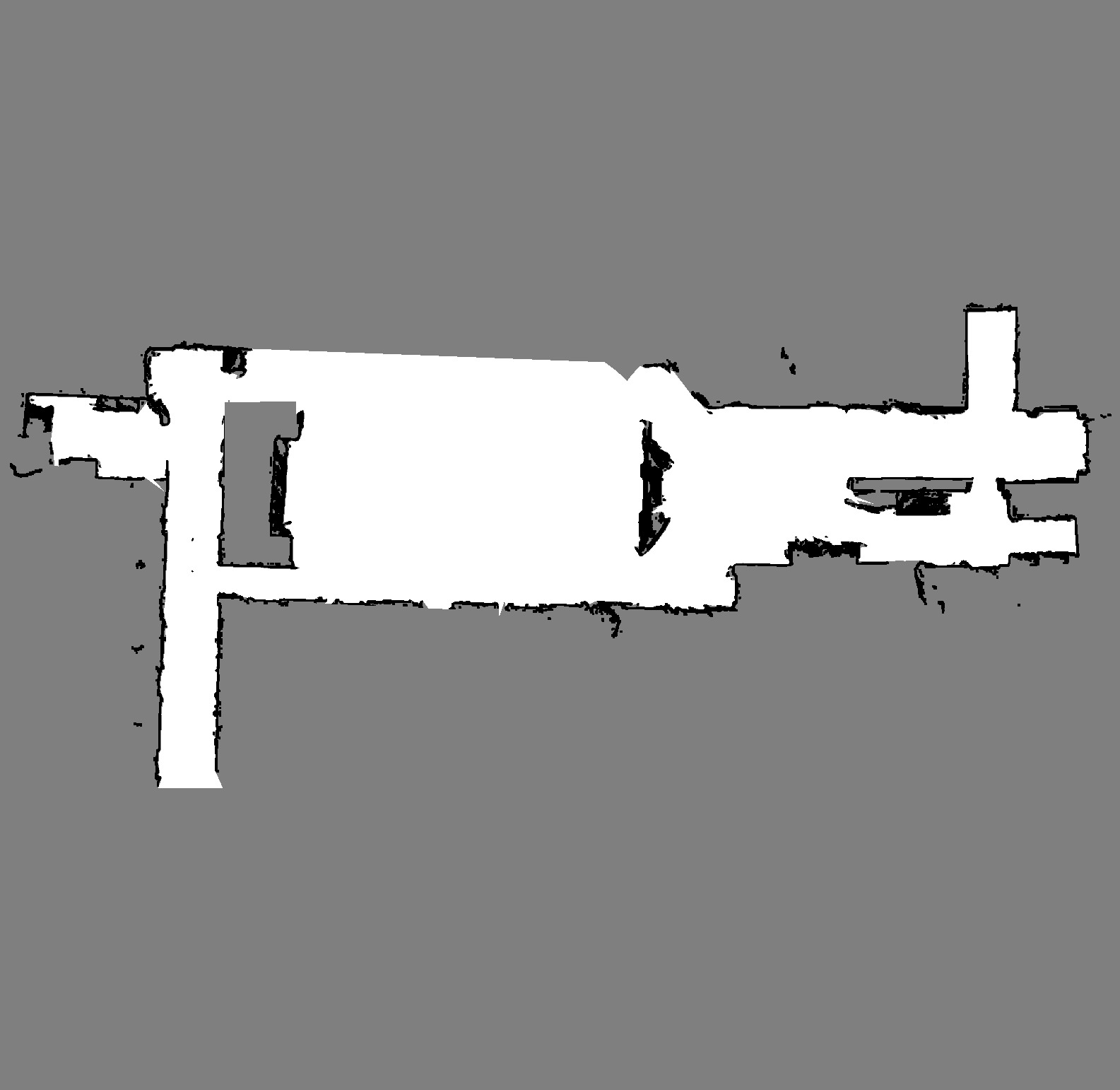}%_20170131131405.jpg}
    % \caption[]{sensor map 2}%% \label{xxx}
  \end{subfigure}%

  \begin{subfigure}{.33\linewidth}
    \includegraphics[width=\linewidth]{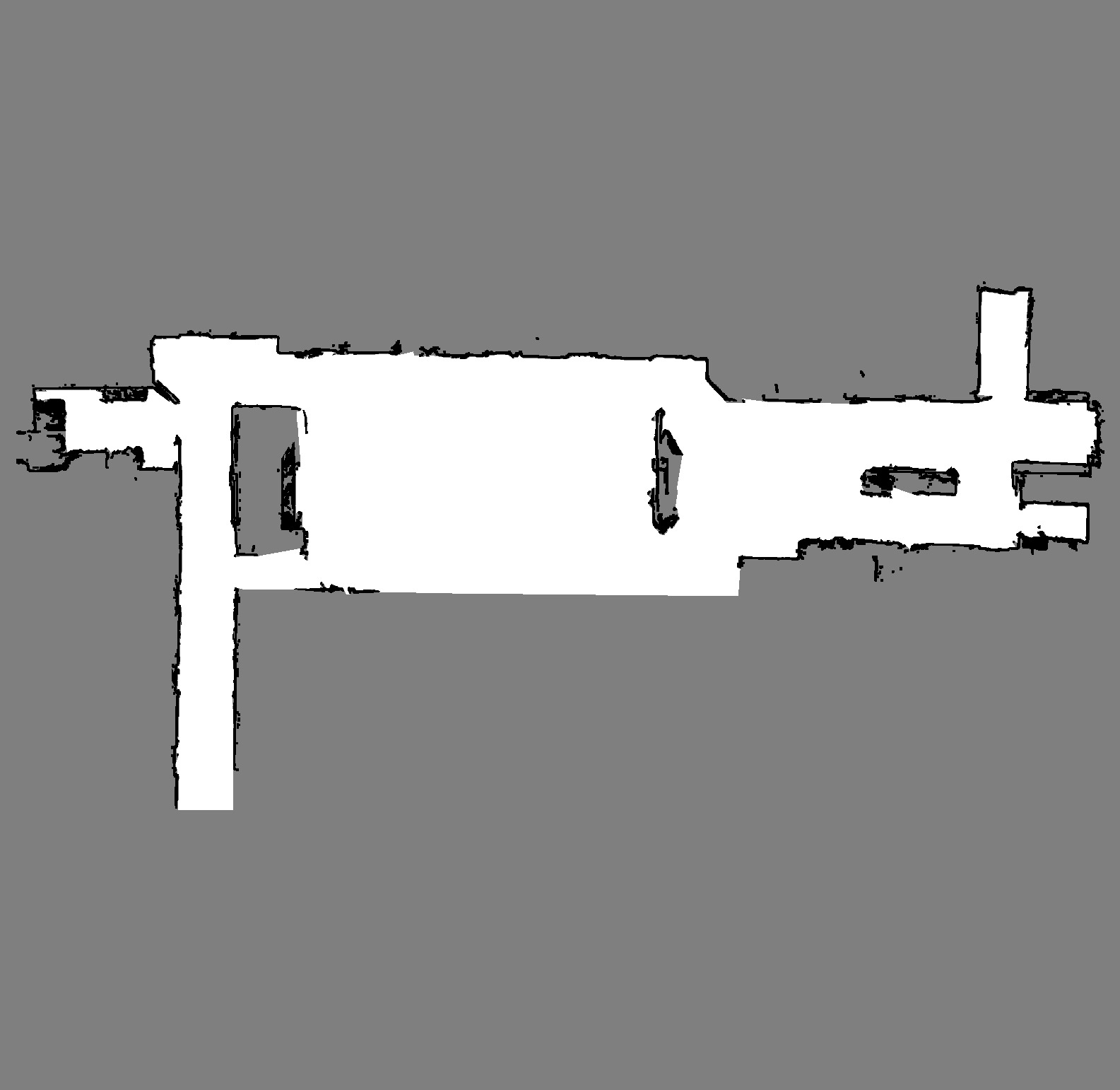}%_20170131130616.jpg}
    % \caption[]{sensor map 3}%% \label{xxx}
  \end{subfigure}%
  ~%
  \begin{subfigure}{.33\linewidth}
    \includegraphics[width=\linewidth]{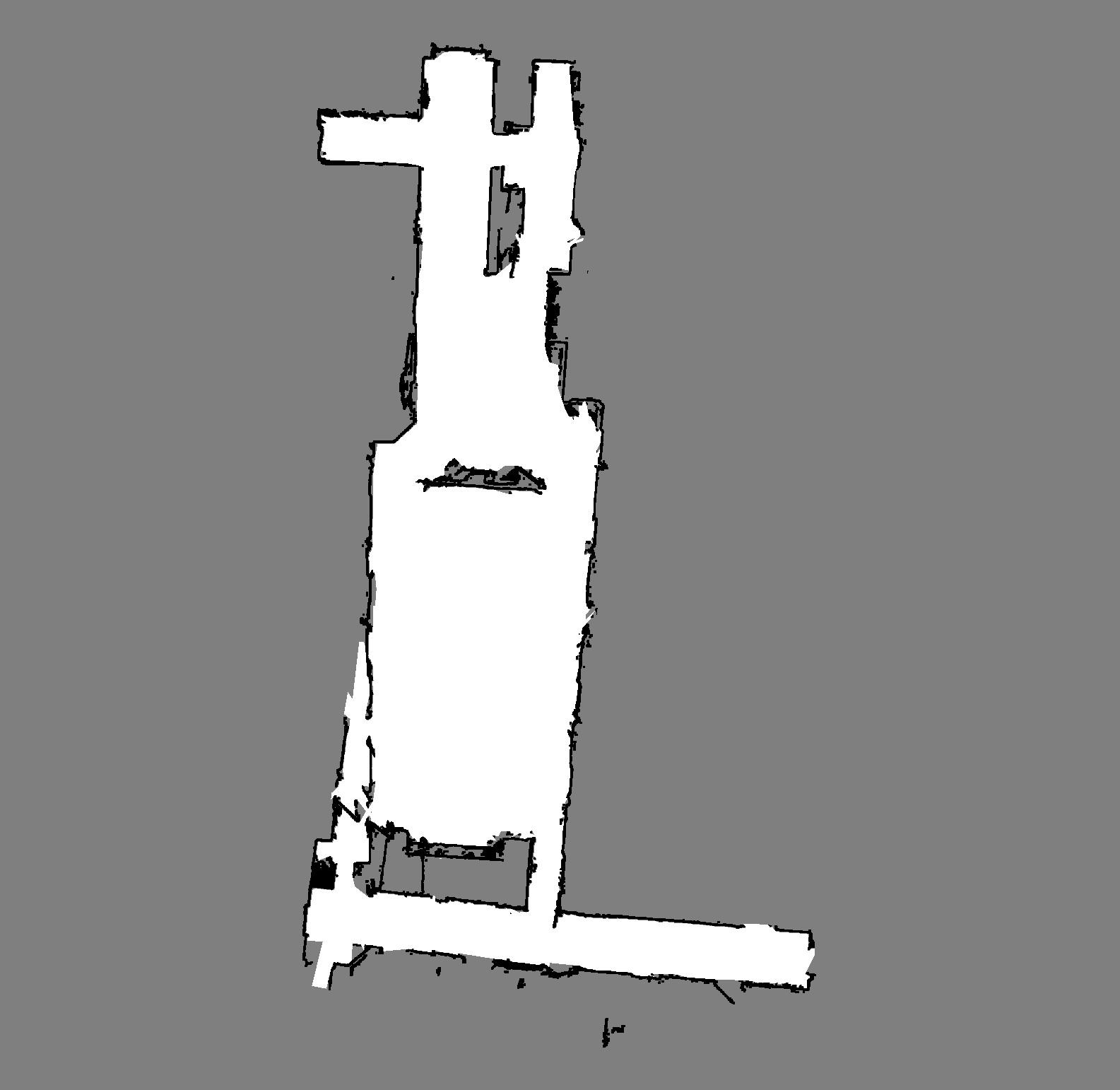}%_20170131122040.jpg}
    % \caption[]{sensor map 4}%% \label{xxx}
  \end{subfigure}%
  ~%
  \begin{subfigure}{.33\linewidth}
    \includegraphics[width=\linewidth]{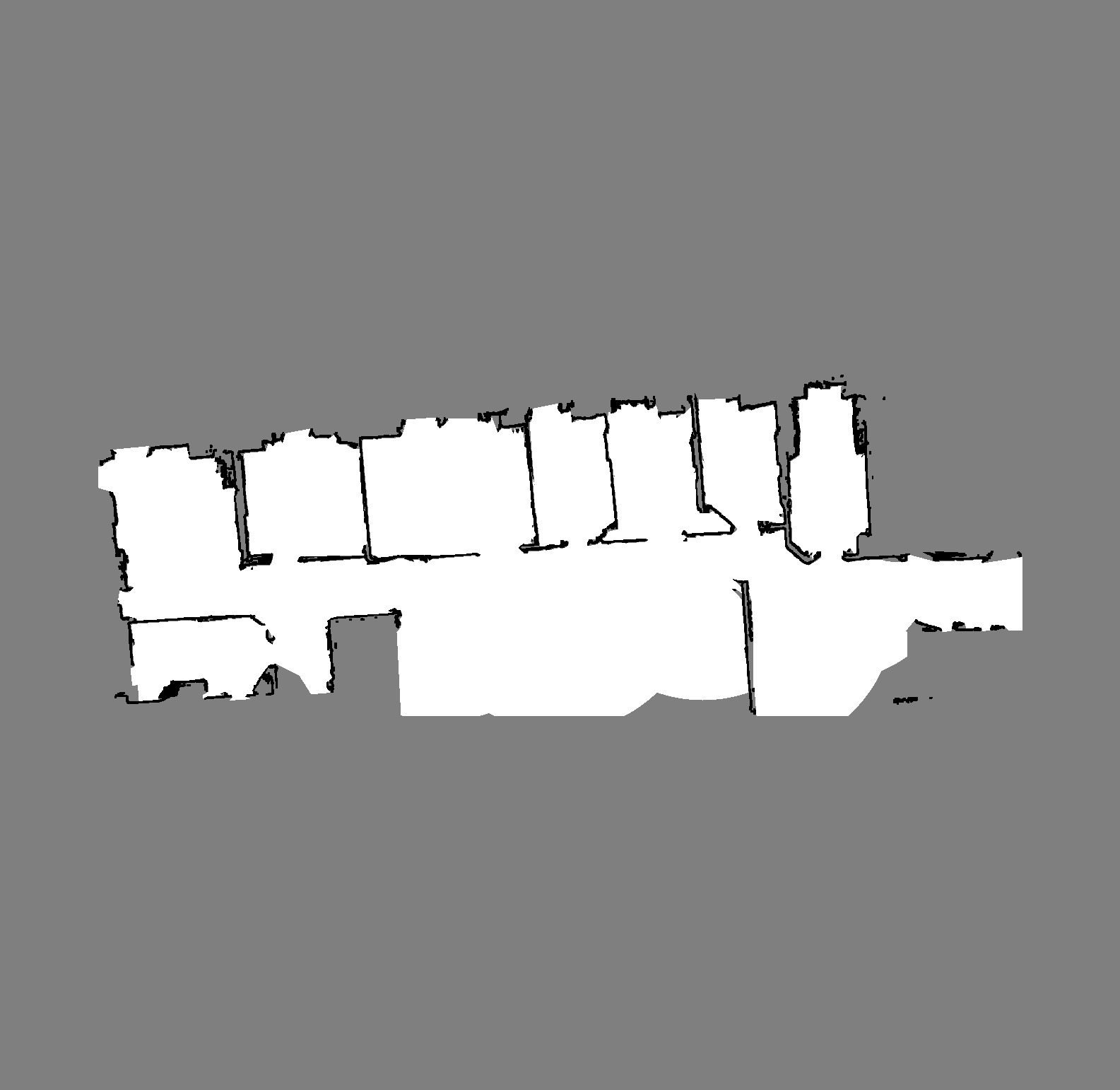}%_20170205104625.jpg}
    % \caption[]{sensor map 5}%% \label{xxx}
  \end{subfigure}%

  \begin{subfigure}{.33\linewidth}
    \includegraphics[width=\linewidth]{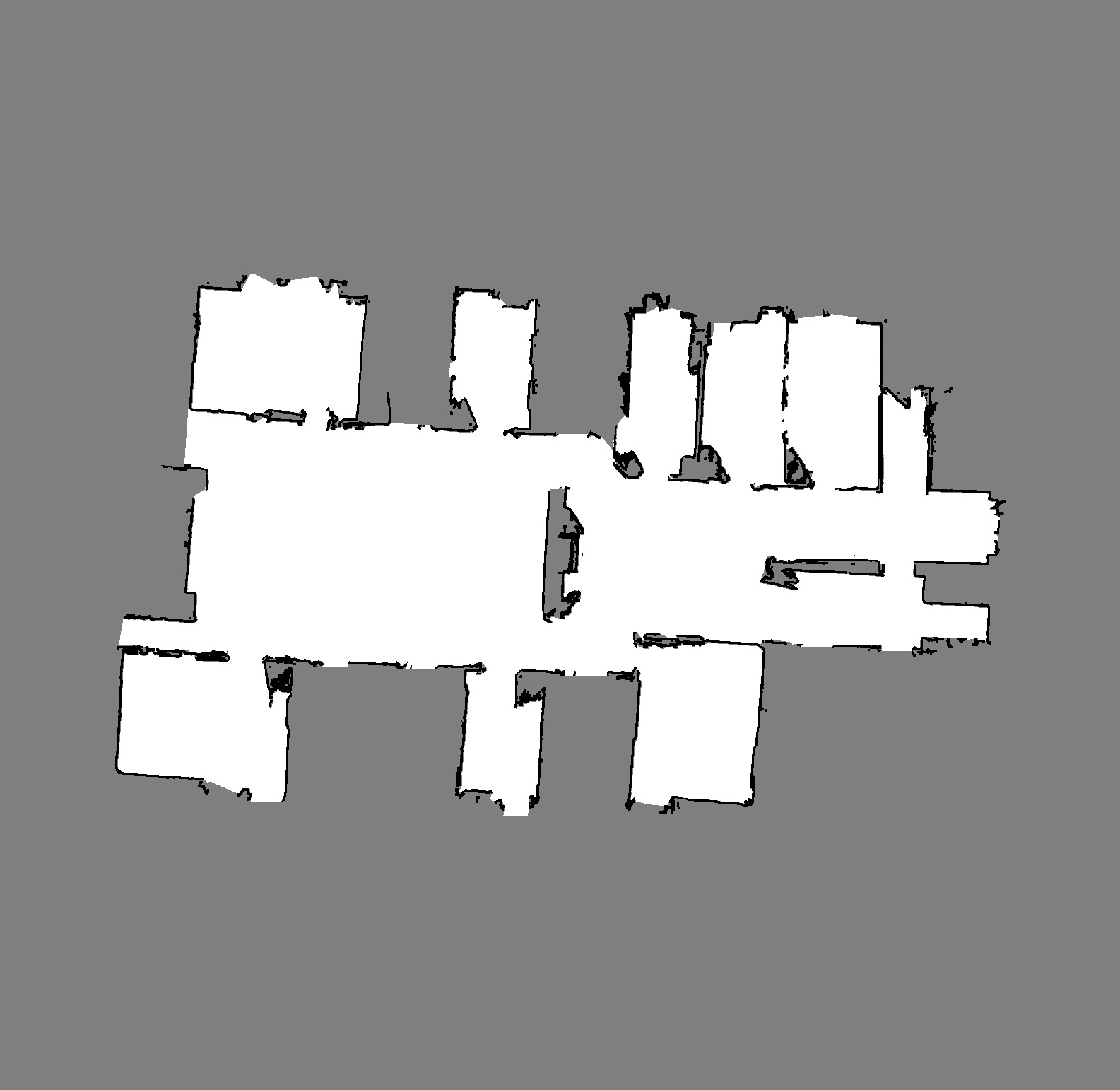}%_20170205105917.jpg}
    % \caption[]{sensor map 6}%% \label{xxx}
  \end{subfigure}%
  ~%
  \begin{subfigure}{.33\linewidth}
    \includegraphics[width=\linewidth]{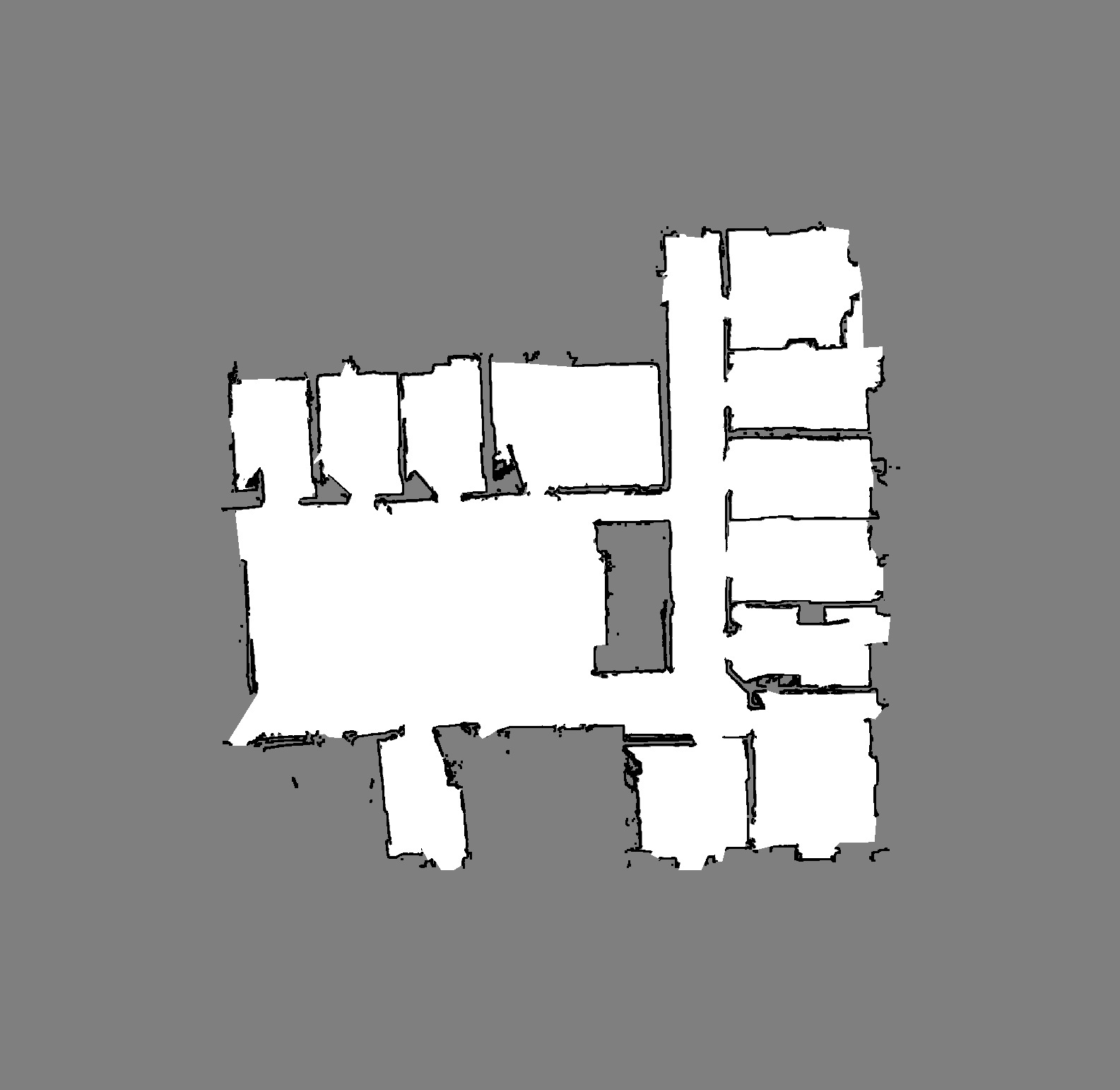}%_20170205111301.jpg}
    % \caption[]{sensor map 7}%% \label{xxx}
  \end{subfigure}%
  ~%
  \begin{subfigure}{.33\linewidth}
    \includegraphics[width=\linewidth]{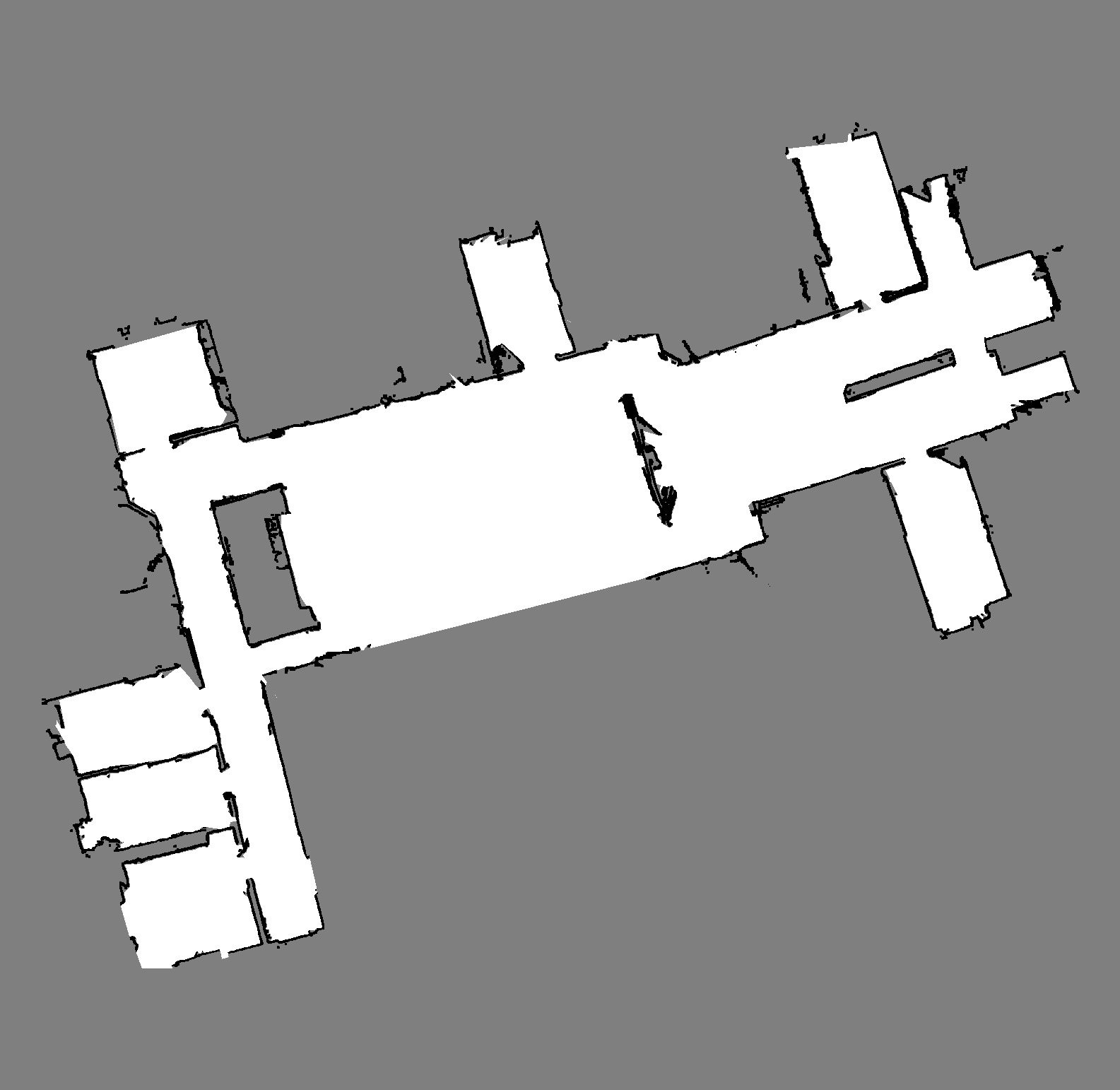}%_20170205112339.jpg}
    % \caption[]{sensor map 8}%% \label{xxx}
  \end{subfigure}%

  \begin{subfigure}{.33\linewidth}
    \includegraphics[width=\linewidth]{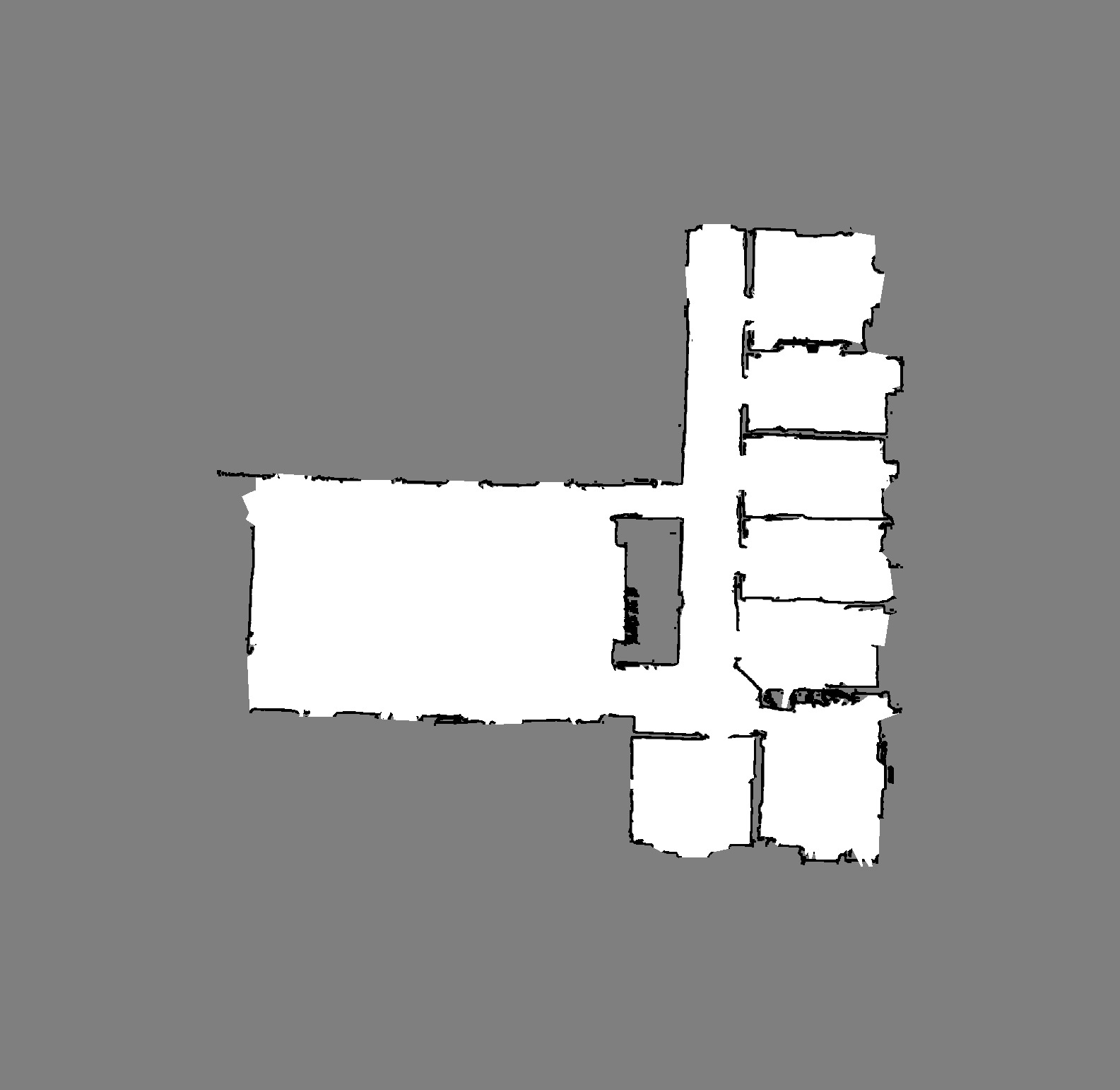}%_20170205110552.jpg}
    % \caption[]{sensor map 9}%% \label{xxx}
  \end{subfigure}%
  ~%
  \begin{subfigure}{.33\linewidth}
    \includegraphics[width=\linewidth]{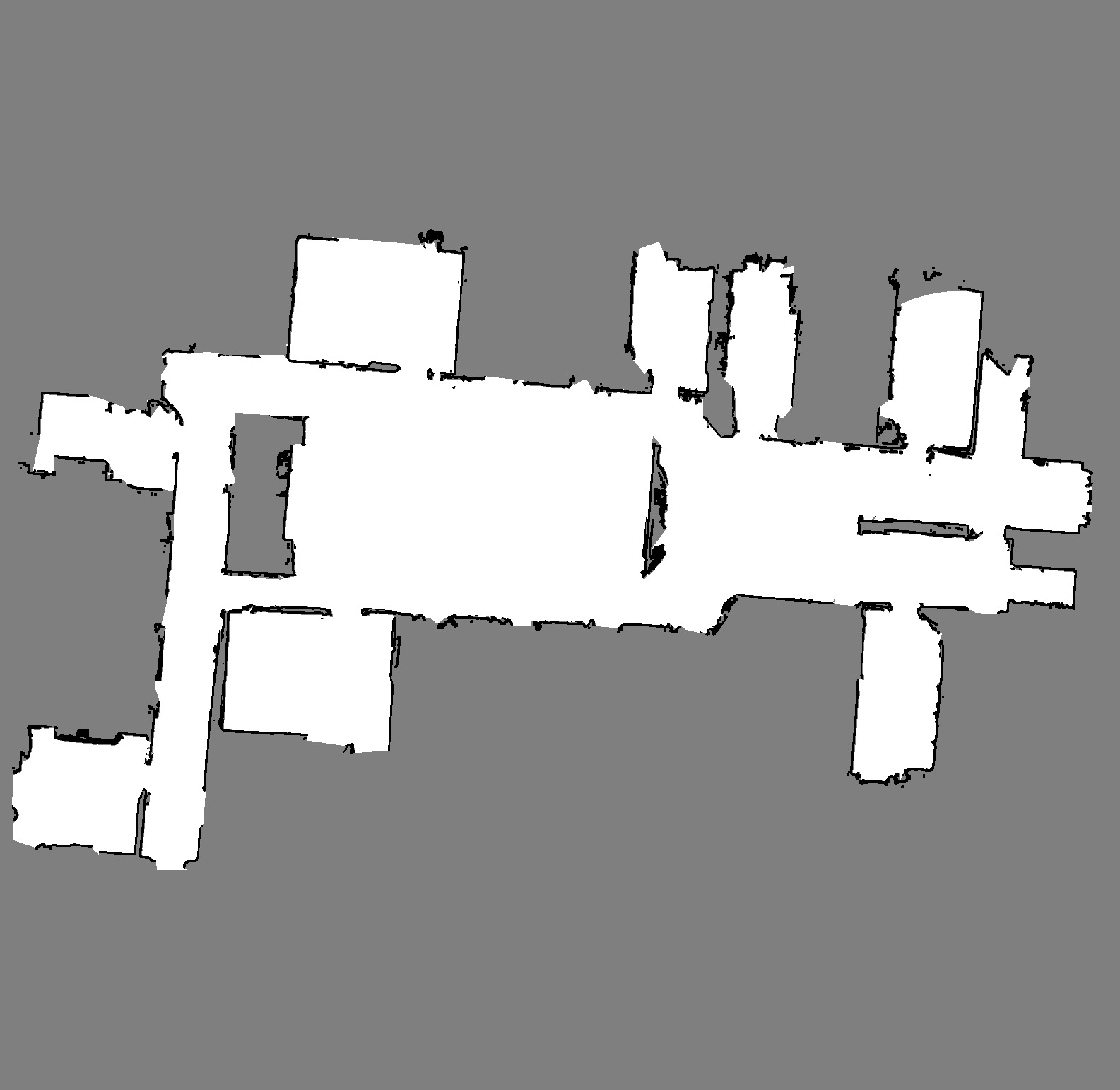}%_20170205111807.jpg}
    % \caption[]{sensor map 10}%% \label{xxx}
  \end{subfigure}%
  ~%
  \begin{subfigure}{.33\linewidth}
    \includegraphics[width=\linewidth]{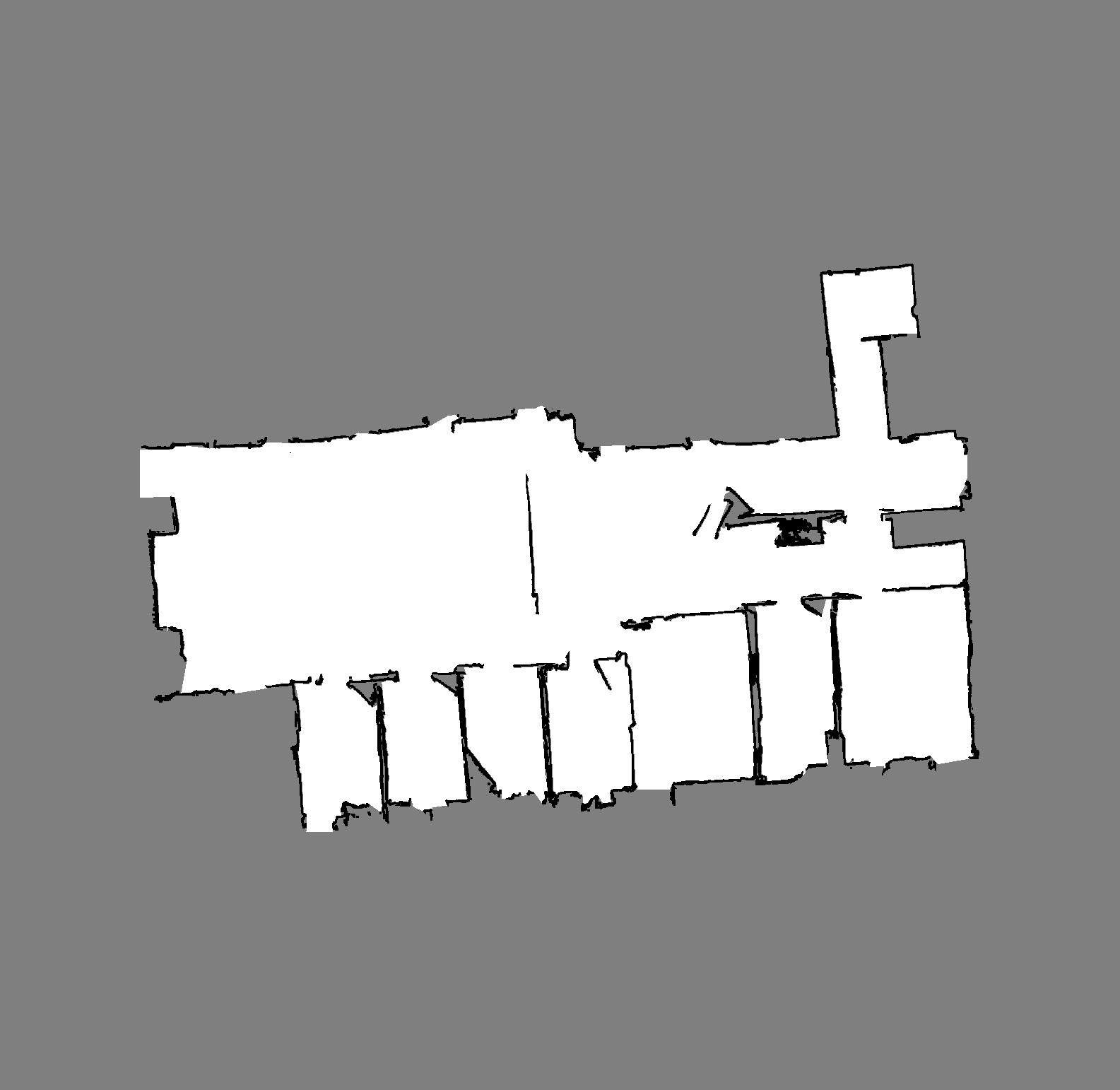}%_20170409125554.jpg}
    % \caption[]{sensor map 11}%% \label{xxx}
  \end{subfigure}%

  \begin{subfigure}{.33\linewidth}
    \includegraphics[width=\linewidth]{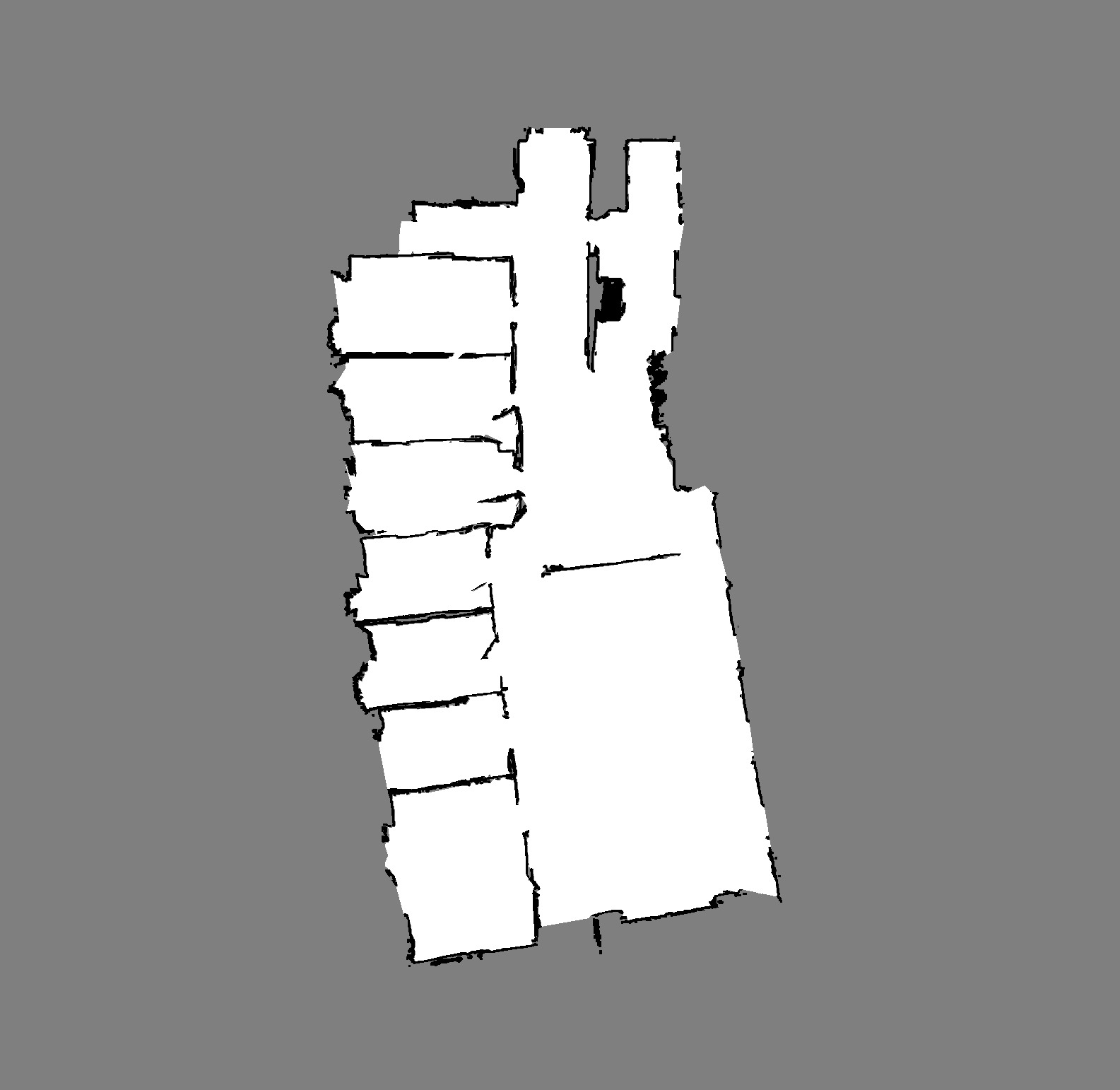}%_20170409130127.jpg}
    % \caption[]{sensor map 12}%% \label{xxx}
  \end{subfigure}%
  ~%
  \begin{subfigure}{.33\linewidth}
    \includegraphics[width=\linewidth]{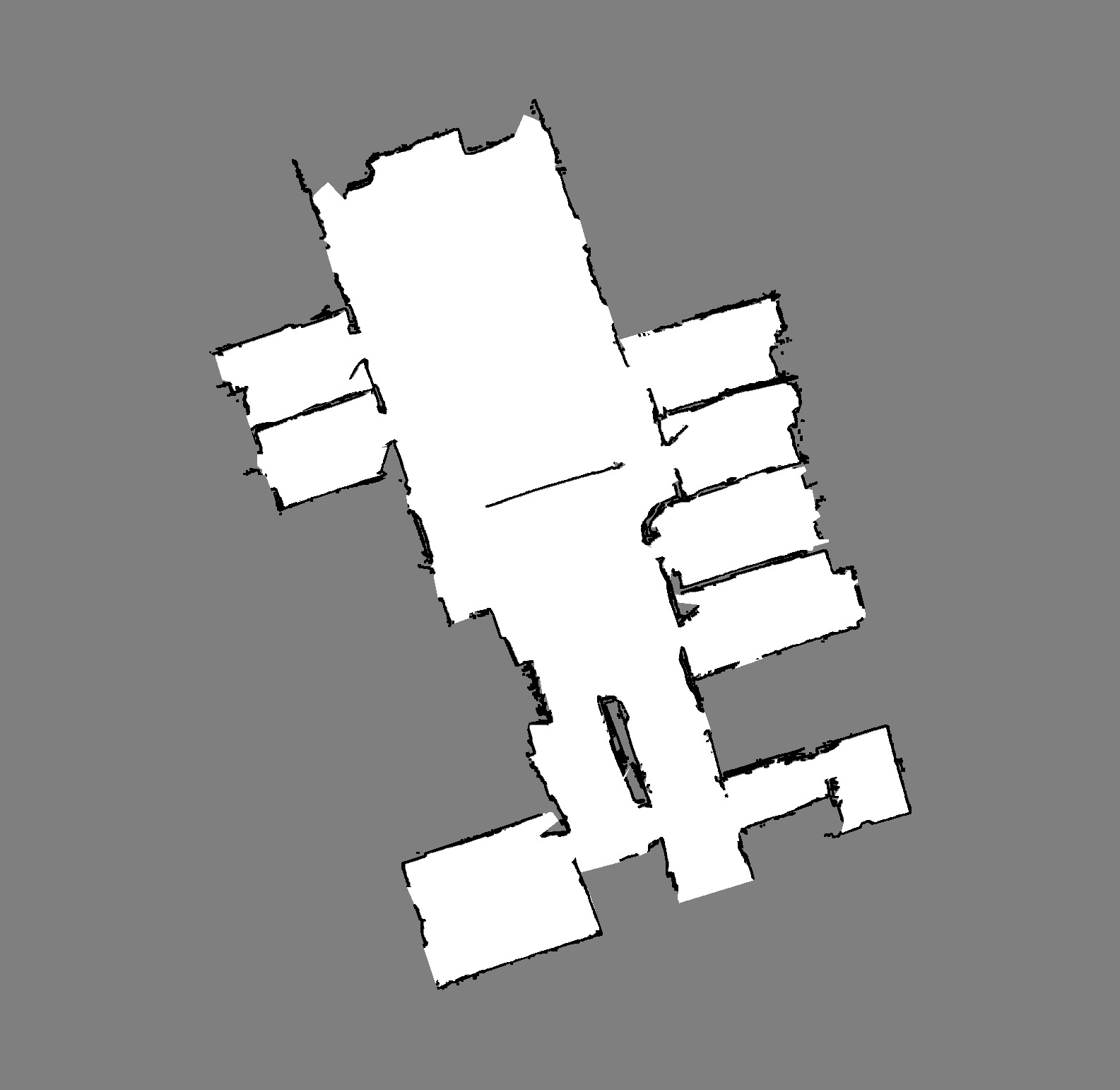}%_20170409130542.jpg}
    % \caption[]{sensor map 13}%% \label{xxx}
  \end{subfigure}%
  ~%
  \begin{subfigure}{.33\linewidth}
    \includegraphics[width=\linewidth]{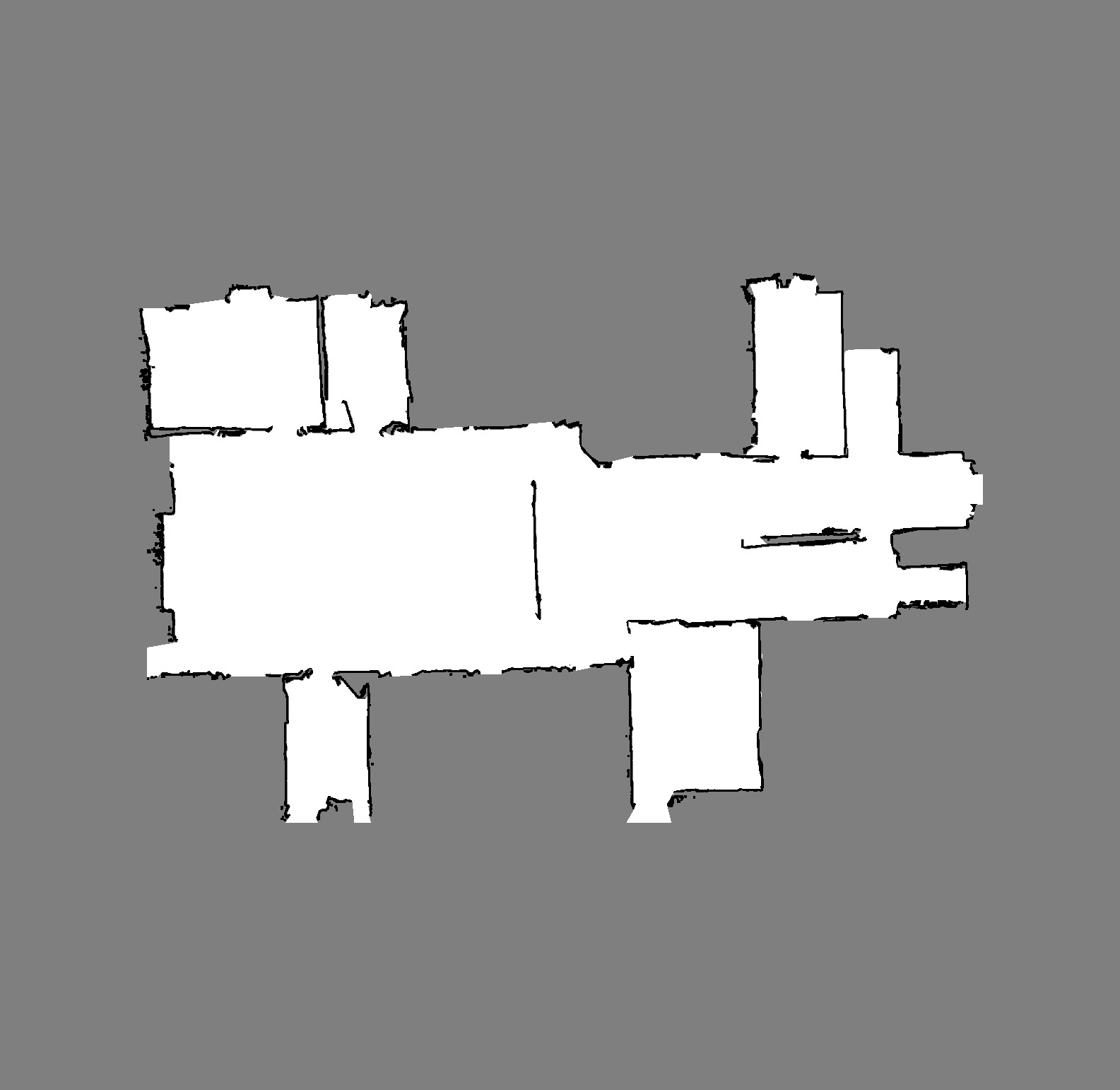}%_20170409131152.jpg}
    % \caption[]{sensor map 14}%% \label{xxx}
  \end{subfigure}%
  \caption[xxx]{HH\_E5 (office building in Halmstad, Sweden)}
  \label{fig:E5_HH}
\end{figure}

%%%%%%%%%%%%%%%%%%%%%%%%%%%%%% HIH
\begin{figure}%[!ht]
  \centering
  \begin{subfigure}{.33\linewidth}
    \includegraphics[width=\linewidth]{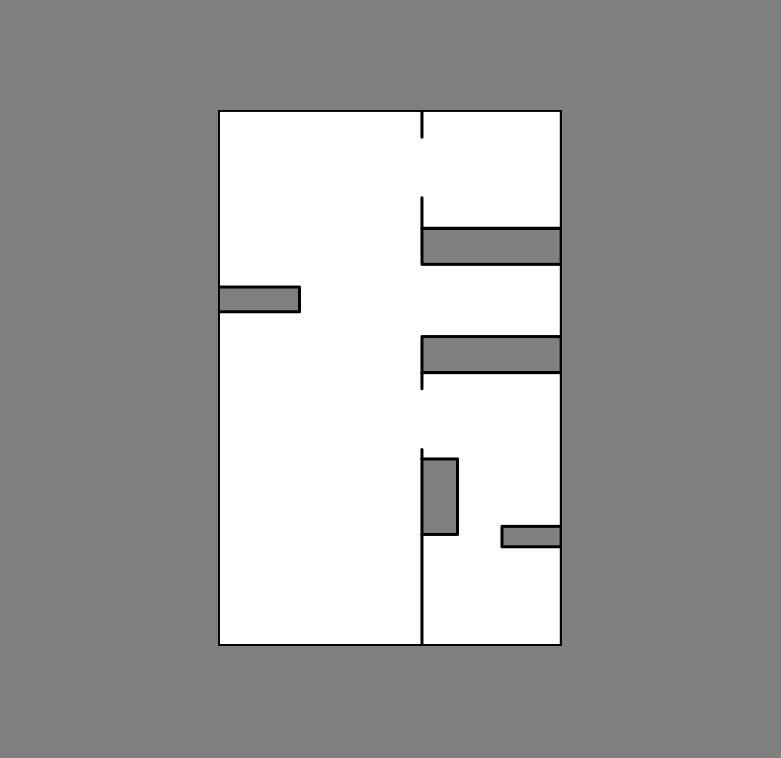}
    % \caption[]{layout}%% \label{xxx}
  \end{subfigure}%
  ~%
  \begin{subfigure}{.33\linewidth}
    \includegraphics[width=\linewidth]{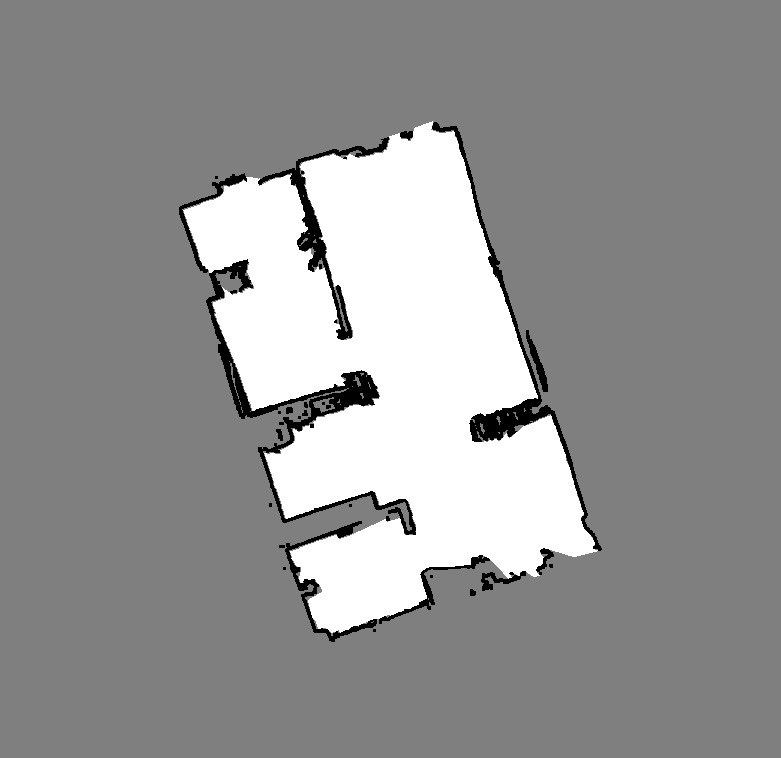}%_20170131135829.jpg}
    % \caption[]{sensor map 1}%% \label{xxx}
  \end{subfigure}%
  ~%
  \begin{subfigure}{.33\linewidth}
    \includegraphics[width=\linewidth]{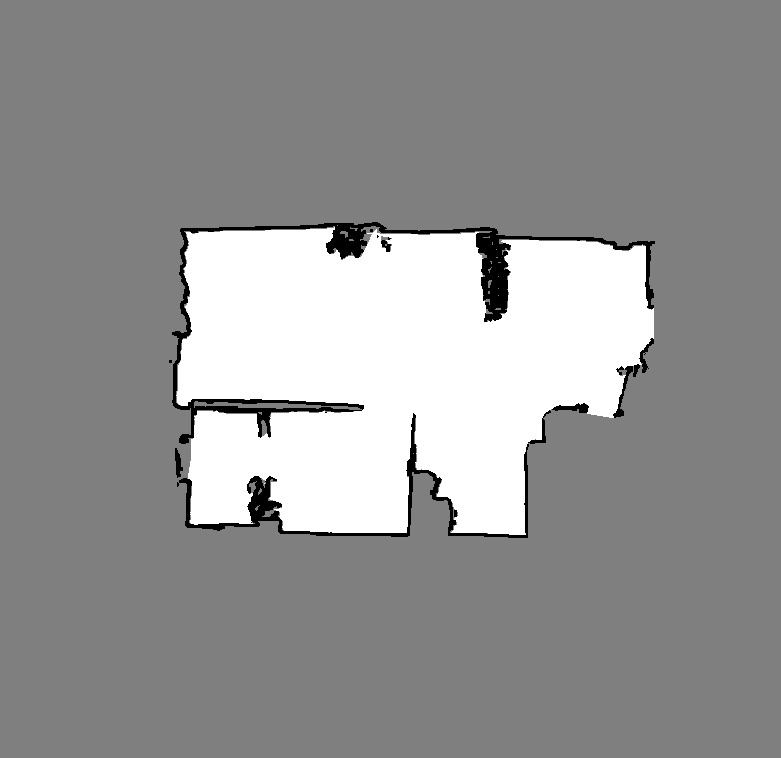}%_20170409123351.jpg}
    % \caption[]{sensor map 2}%% \label{xxx}
  \end{subfigure}%

  \begin{subfigure}{.33\linewidth}
    \includegraphics[width=\linewidth]{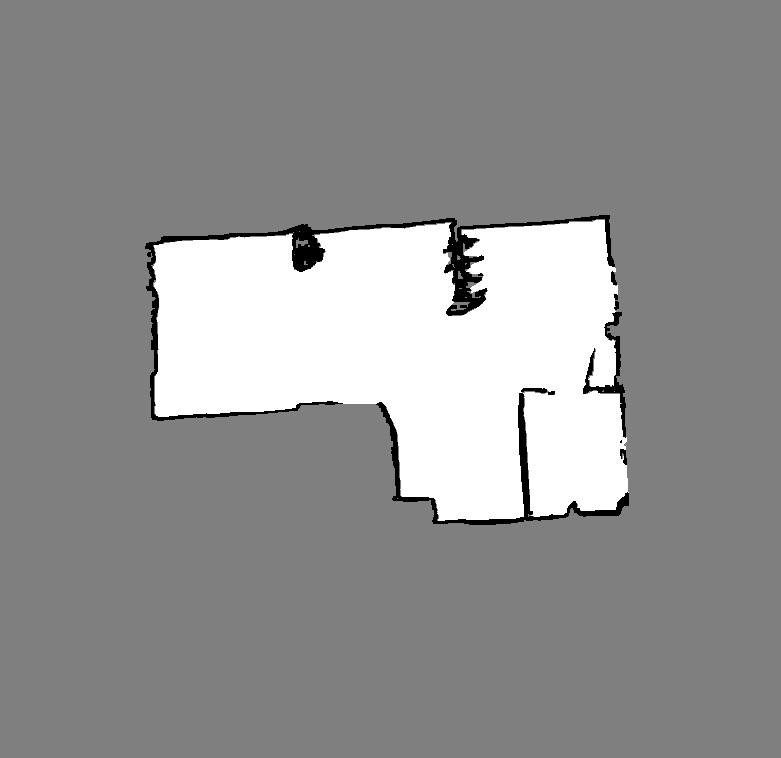}%_20170409123544.jpg}
    % \caption[]{sensor map 3}%% \label{xxx}
  \end{subfigure}%
  ~%
  \begin{subfigure}{.33\linewidth}
    \includegraphics[width=\linewidth]{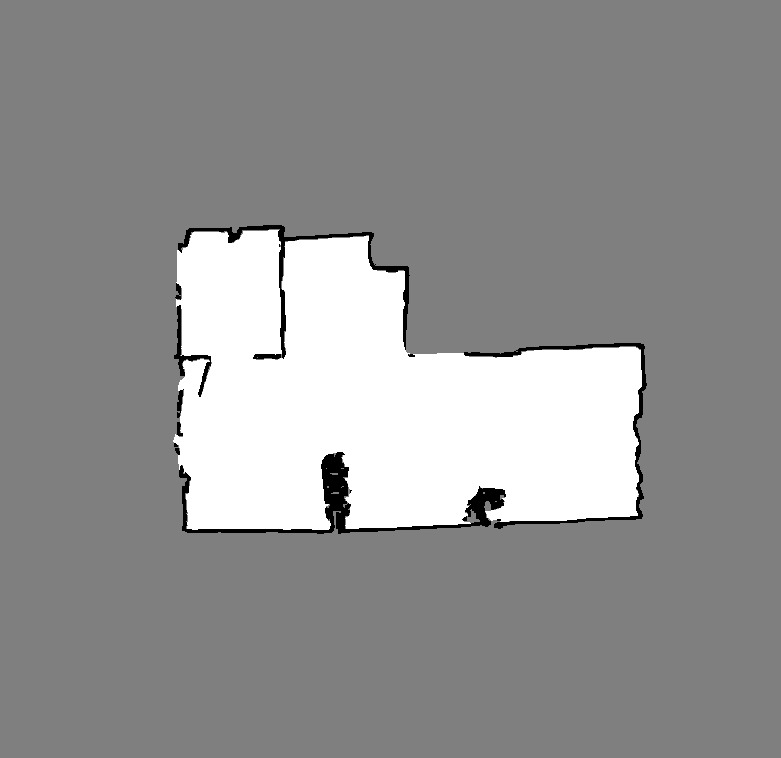}%_20170409123754.jpg}
    % \caption[]{sensor map 4}%% \label{xxx}
  \end{subfigure}
  \caption[xxx]{HIH (Halmstad Intelligent Home~\cite{lundstrom2016halmstad})}
  \label{fig:HIH_HH}
\end{figure}

%%%%%%%%%%%%%%%%%%%%%%%%%%%%%% F5
\begin{figure}%[!ht]
  \centering
  \begin{subfigure}{.33\linewidth}
    \includegraphics[width=\linewidth]{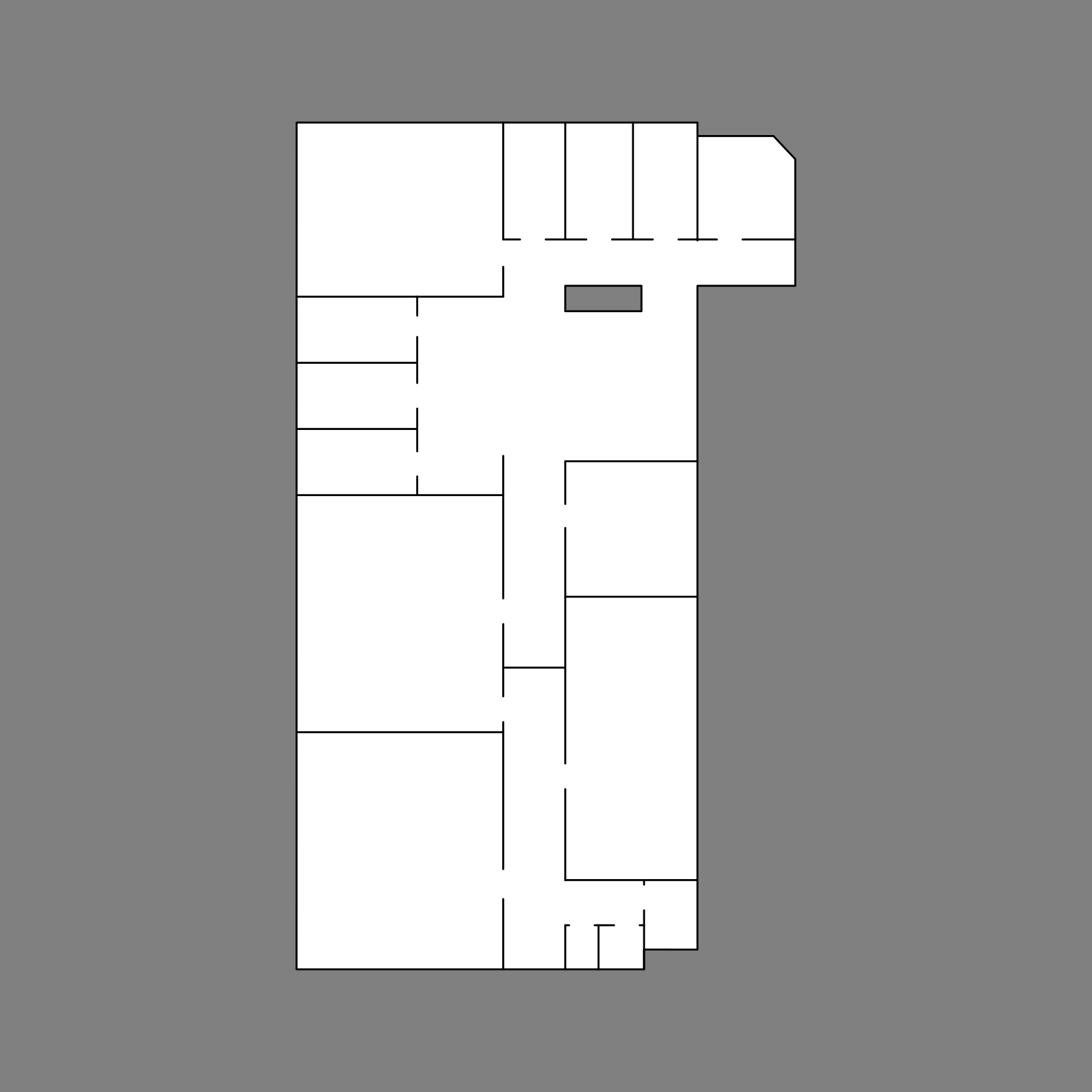}
    % \caption[]{layout}%% \label{xxx}
  \end{subfigure}%
  ~%
  \begin{subfigure}{.33\linewidth}
    \includegraphics[width=\linewidth]{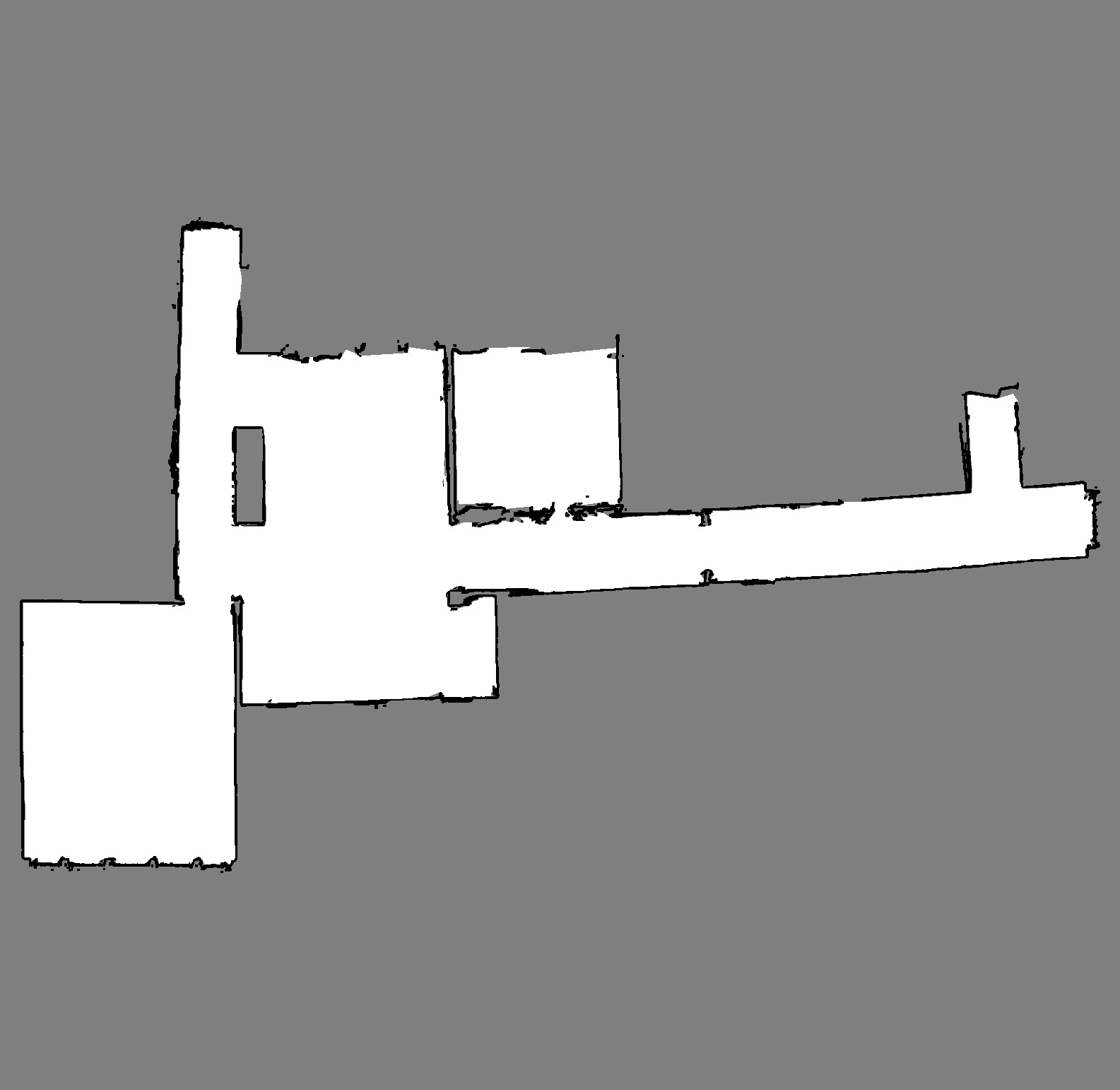}%_20170131132256.jpg}
    % \caption[]{sensor map 1}%% \label{xxx}
  \end{subfigure}%
  ~%
  \begin{subfigure}{.33\linewidth}
    \includegraphics[width=\linewidth]{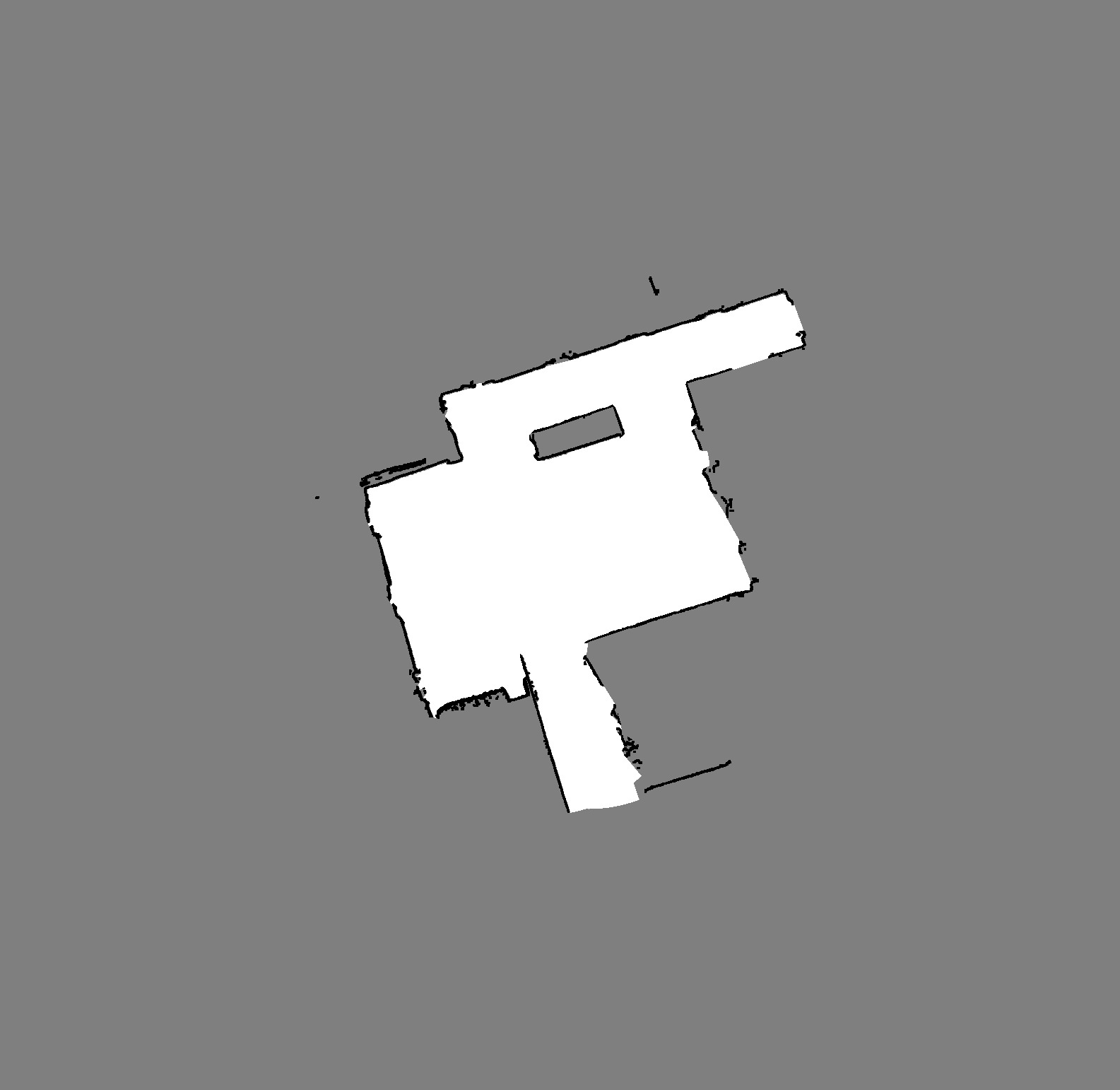}%_20170131125250.jpg}
    % \caption[]{sensor map 2}%% \label{xxx}
  \end{subfigure}%

  \begin{subfigure}{.33\linewidth}
    \includegraphics[width=\linewidth]{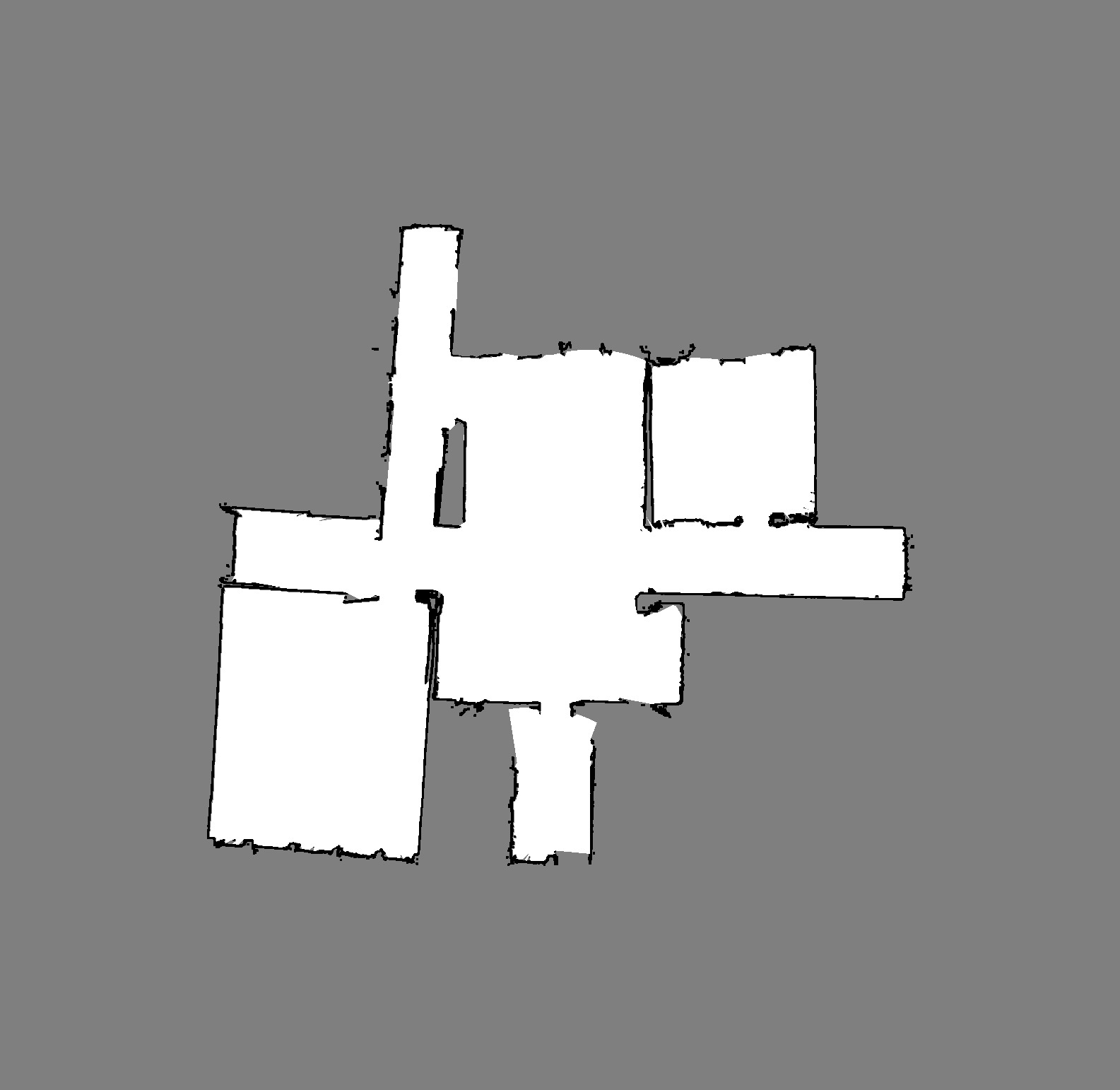}%_20170205114543.jpg}
    % \caption[]{sensor map 3}%% \label{xxx}
  \end{subfigure}%
  ~%
  \begin{subfigure}{.33\linewidth}
    \includegraphics[width=\linewidth]{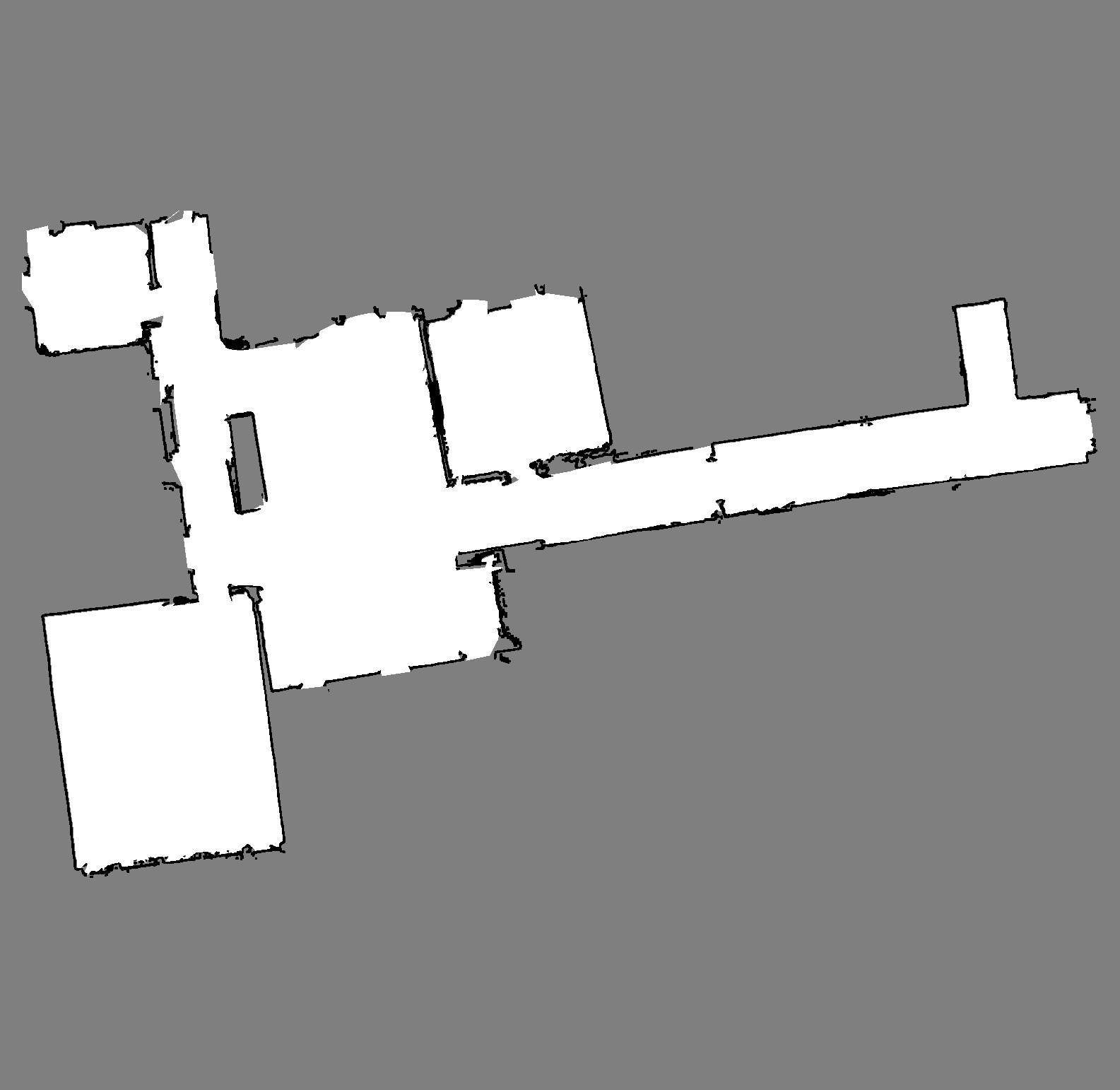}%_20170205115252.jpg}
    % \caption[]{sensor map 4}%% \label{xxx}
  \end{subfigure}%
  ~%
  \begin{subfigure}{.33\linewidth}
    \includegraphics[width=\linewidth]{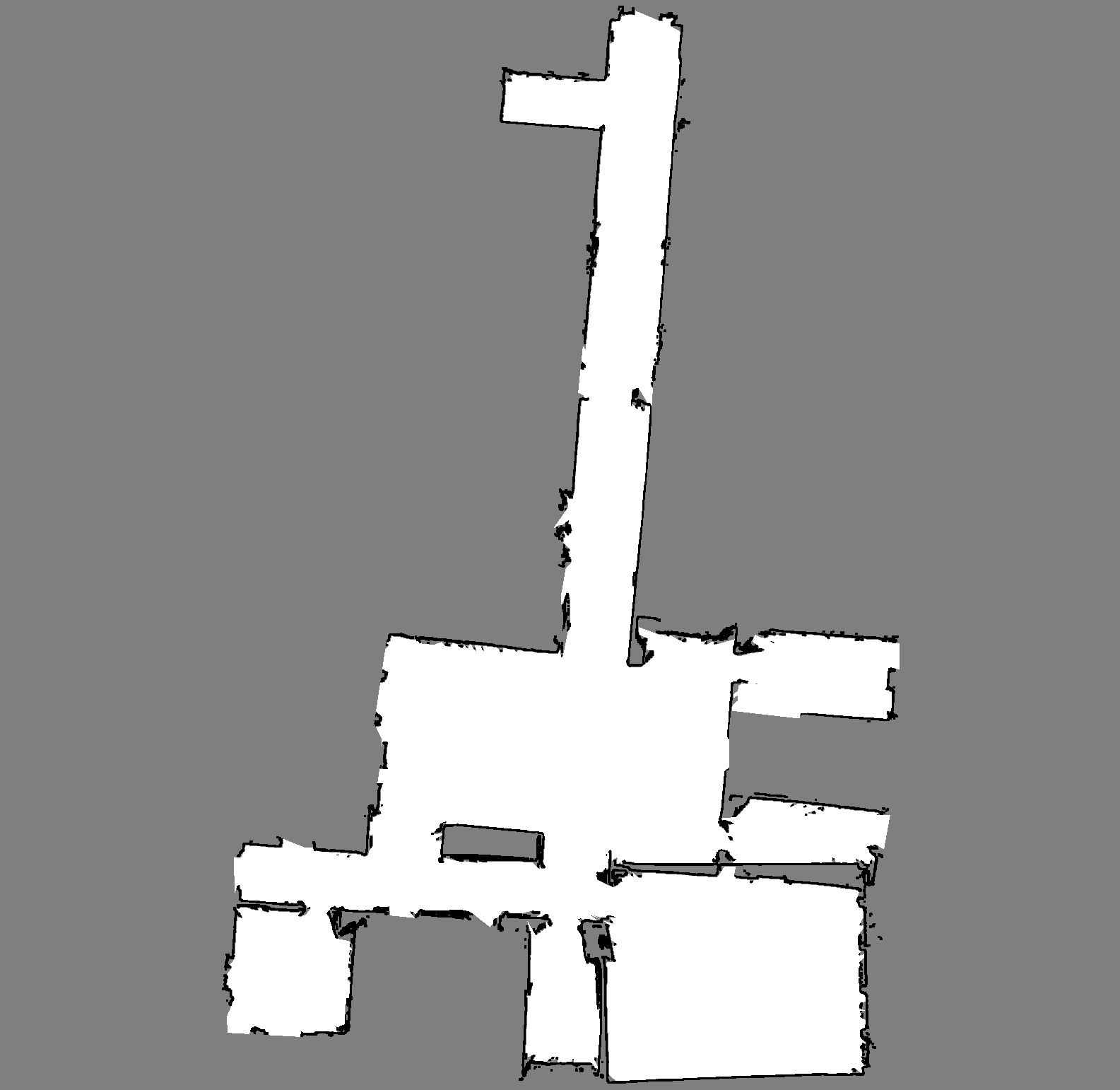}%_20170205115820.jpg}
    % \caption[]{sensor map 5}%% \label{xxx}
  \end{subfigure}%

  \begin{subfigure}{.33\linewidth}
    \includegraphics[width=\linewidth]{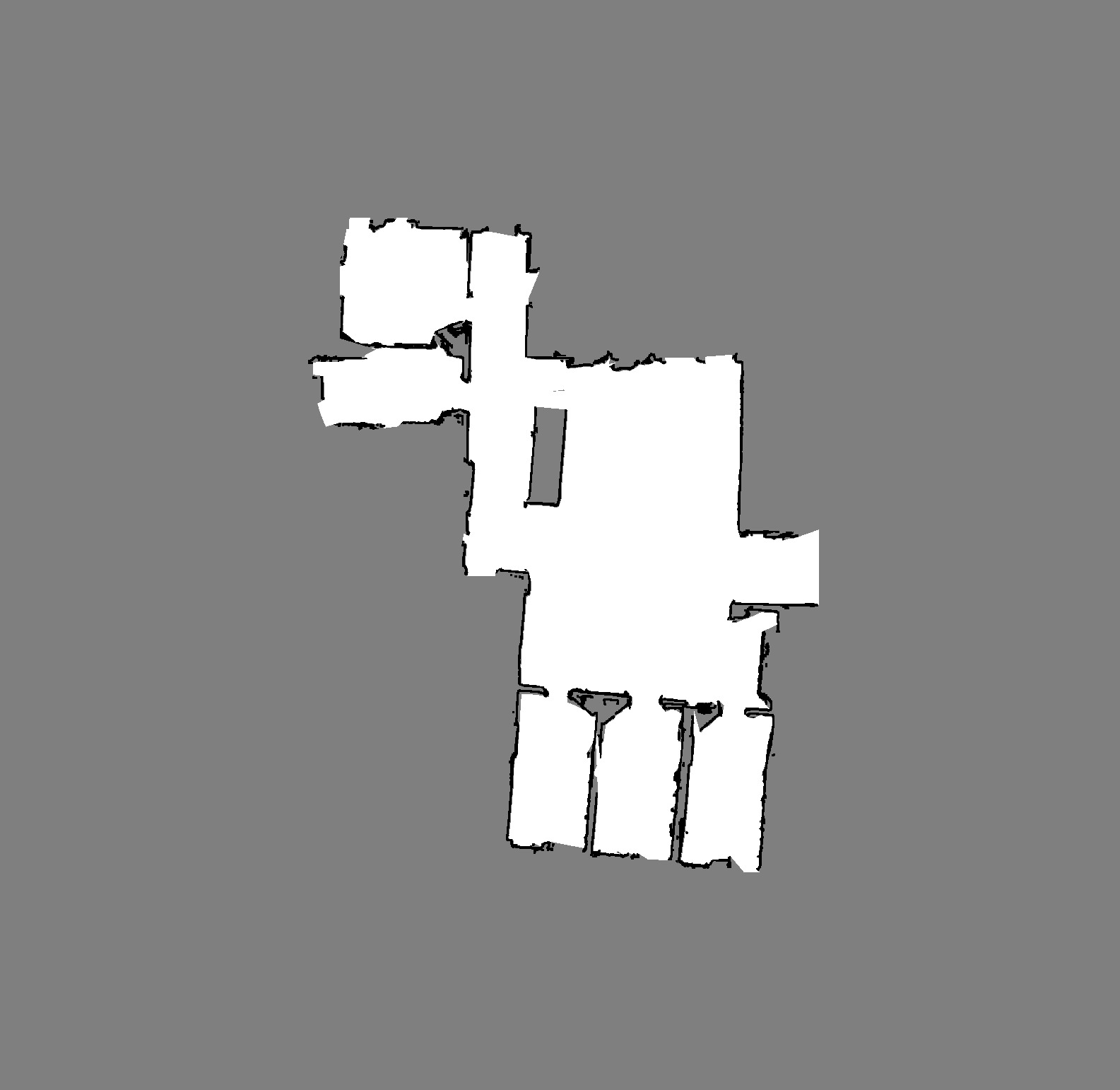}%_20170205114156.jpg}
    % \caption[]{sensor map 6}%% \label{xxx}
  \end{subfigure}%
  ~%
  \begin{subfigure}{.33\linewidth}
    \includegraphics[width=\linewidth]{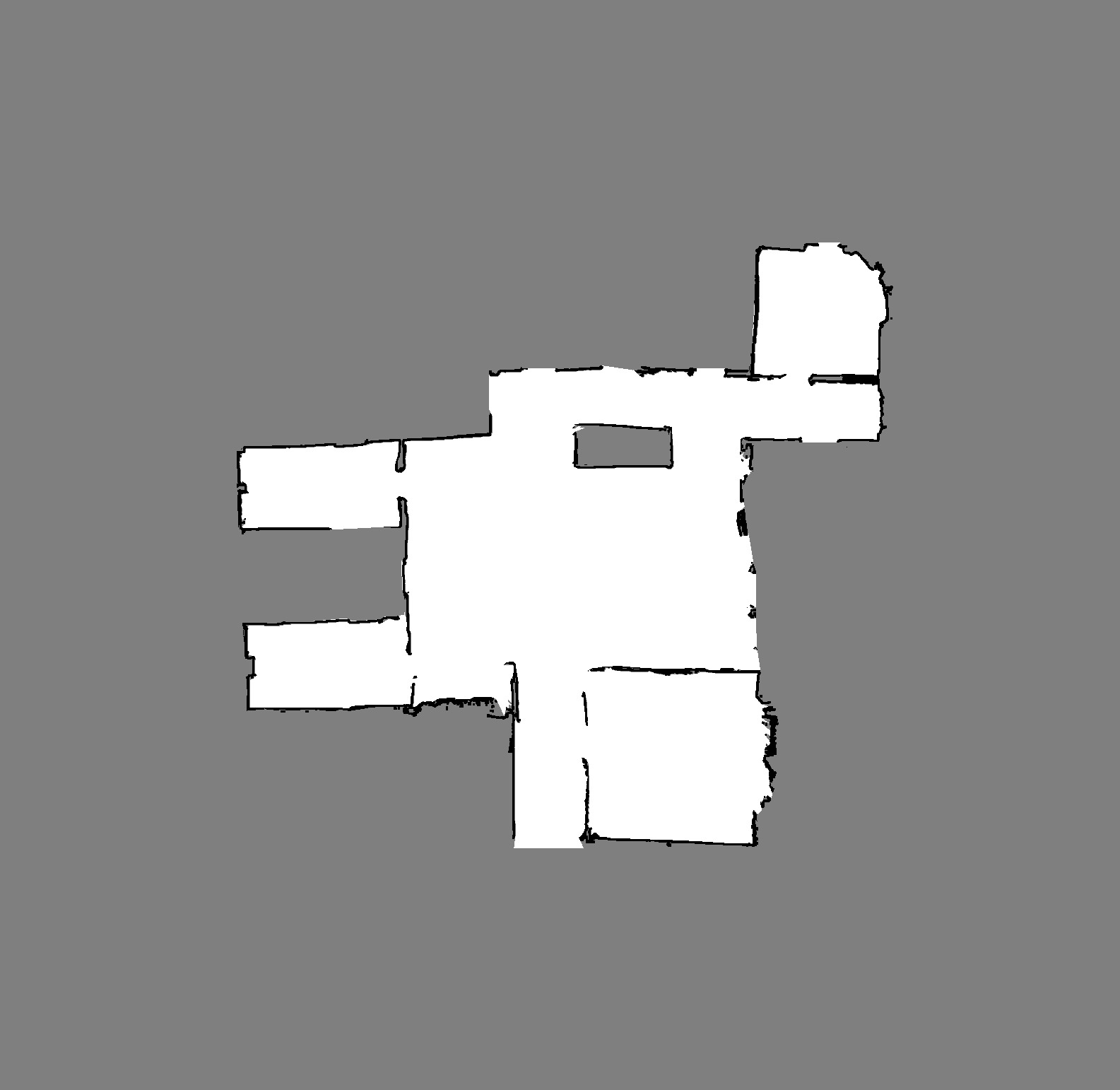}%_20170409113201.jpg}
    % \caption[]{sensor map 7}%% \label{xxx}
  \end{subfigure}%
  ~%
  \begin{subfigure}{.33\linewidth}
    \includegraphics[width=\linewidth]{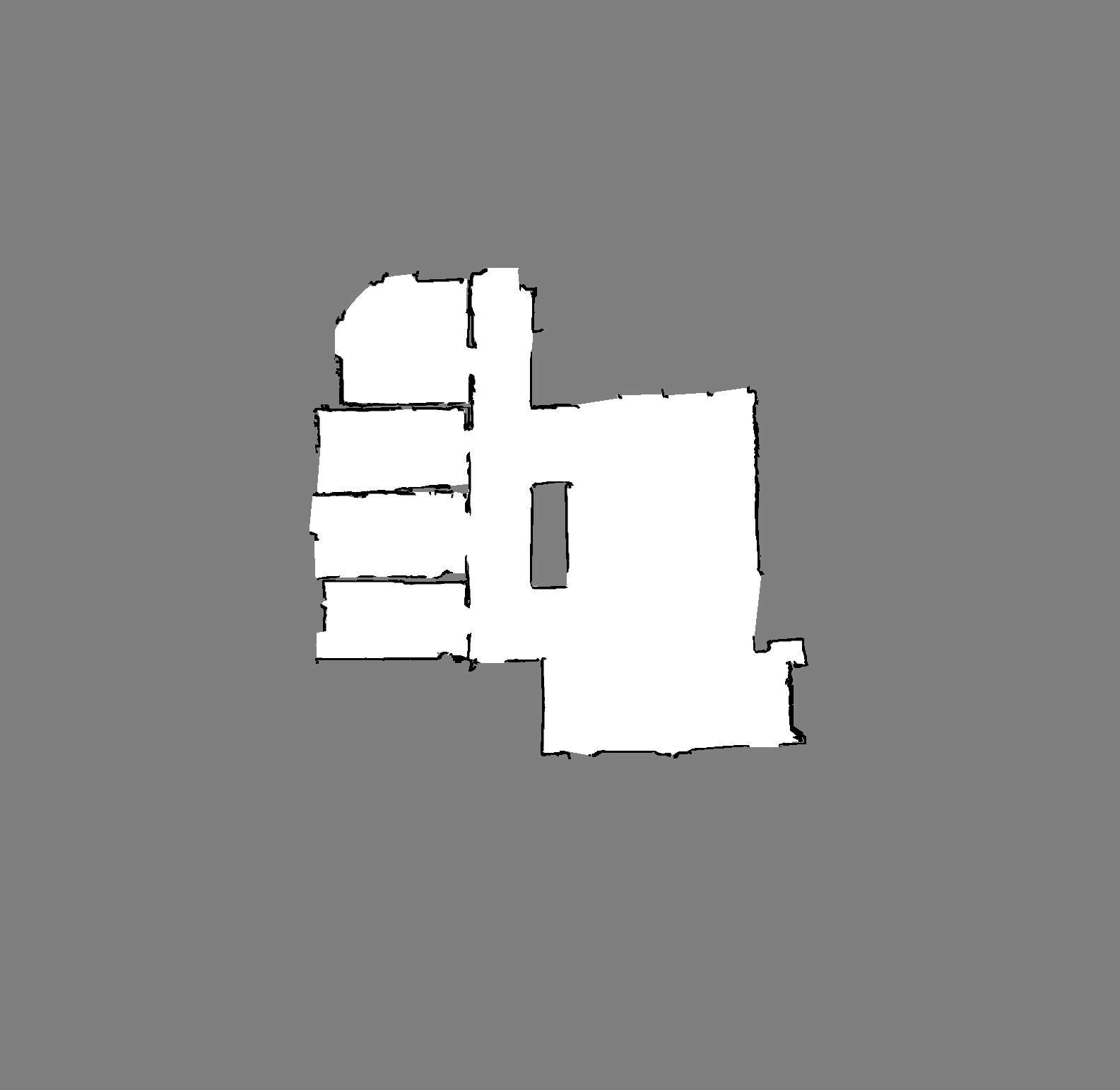}%_20170409113636.jpg}
    % \caption[]{sensor map 8}%% \label{xxx}
  \end{subfigure}%

  \begin{subfigure}{.33\linewidth}
    \includegraphics[width=\linewidth]{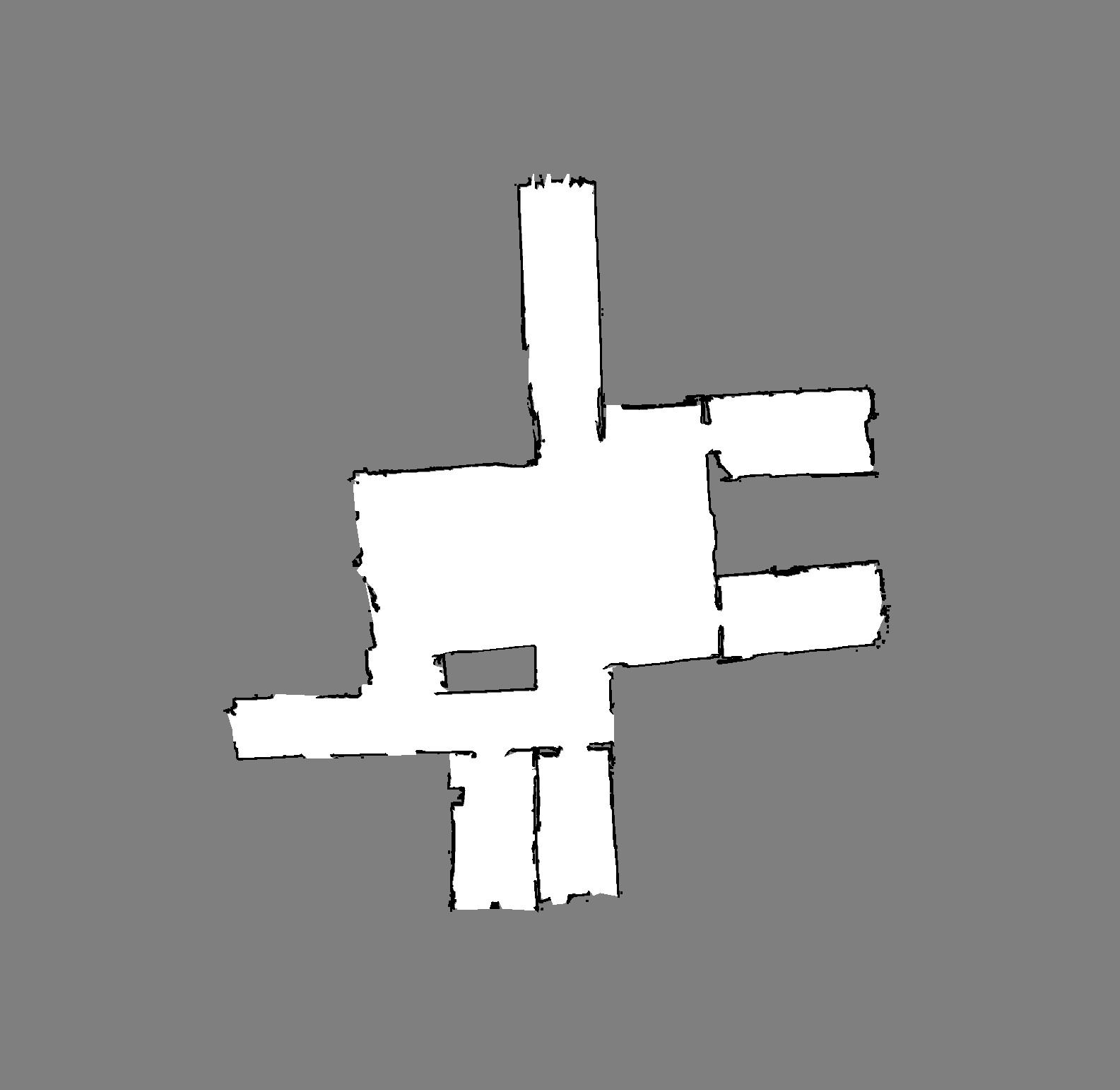}%_20170409114748.jpg}
    % \caption[]{sensor map 9}%% \label{xxx}
  \end{subfigure}%
  ~%
  \begin{subfigure}{.33\linewidth}
    \includegraphics[width=\linewidth]{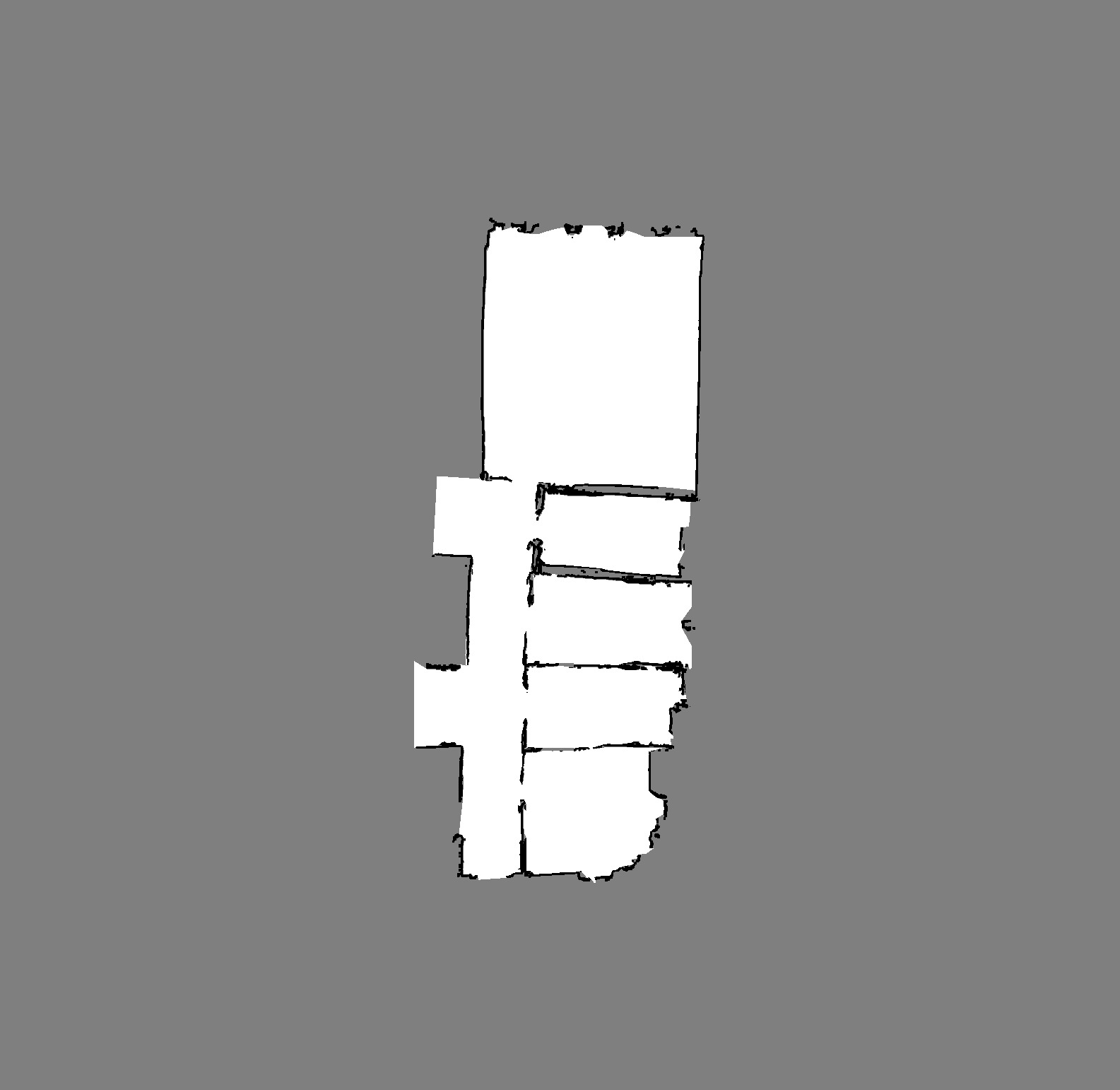}%_20170409115054.jpg}
    % \caption[]{sensor map 10}%% \label{xxx}
  \end{subfigure}%
  ~%
  \begin{subfigure}{.33\linewidth}
    \includegraphics[width=\linewidth]{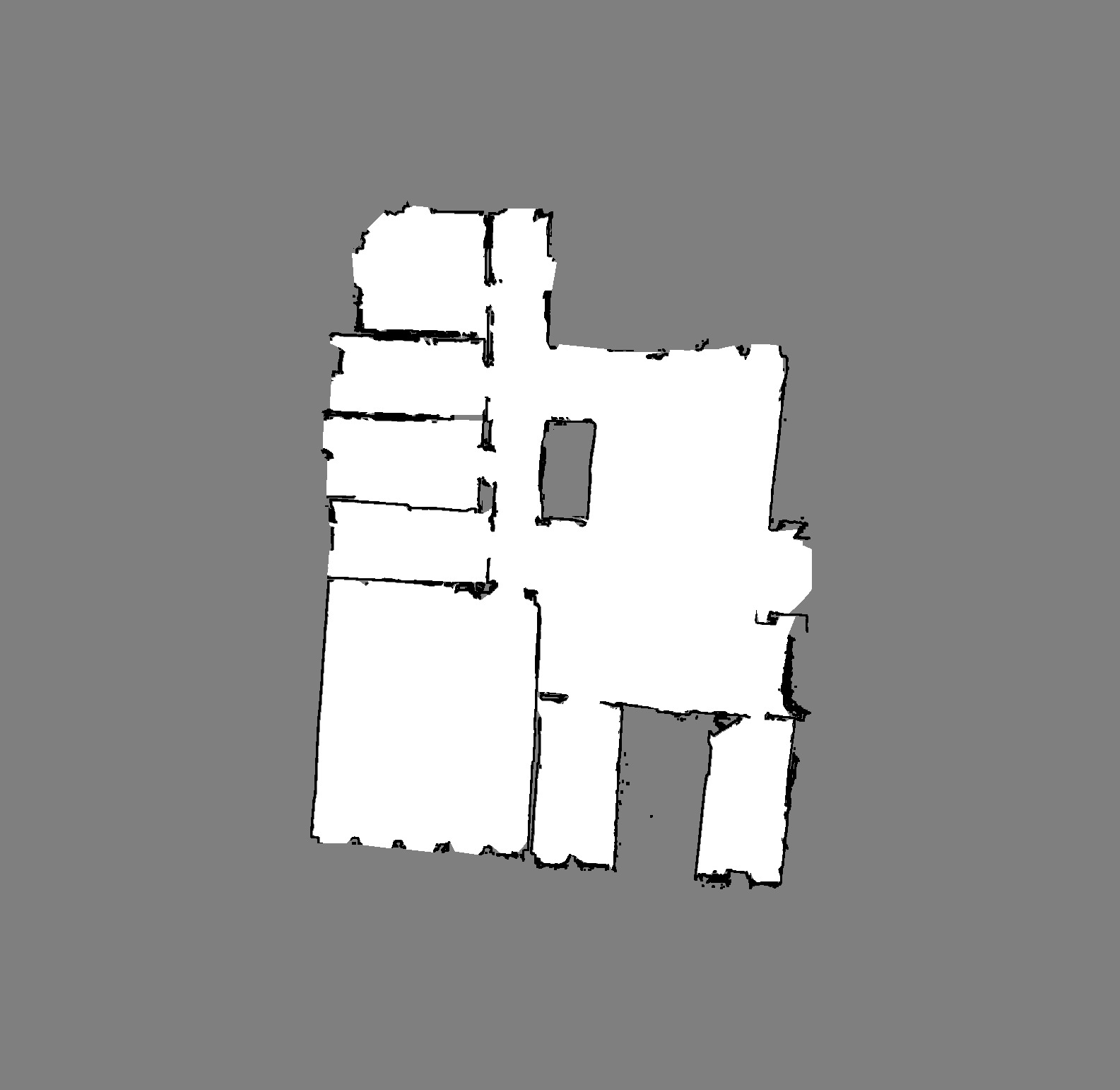}%_20170409115625.jpg}
    % \caption[]{sensor map 11}%% \label{xxx}
  \end{subfigure}%

  \begin{subfigure}{.33\linewidth}
    \includegraphics[width=\linewidth]{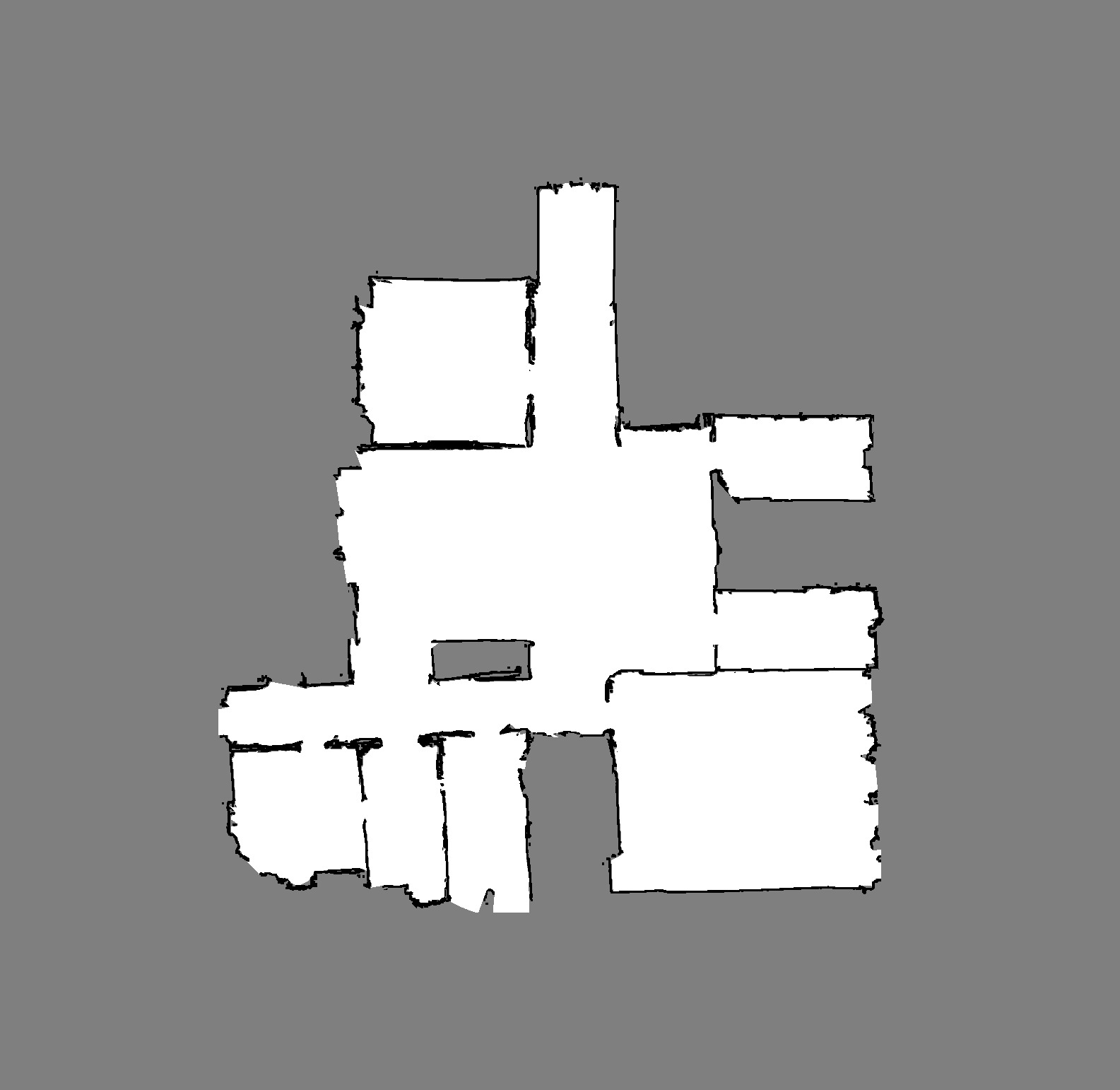}%_20170409120348.jpg}
    % \caption[]{sensor map 12}%% \label{xxx}
  \end{subfigure}%
  ~%
  \begin{subfigure}{.33\linewidth}
    \includegraphics[width=\linewidth]{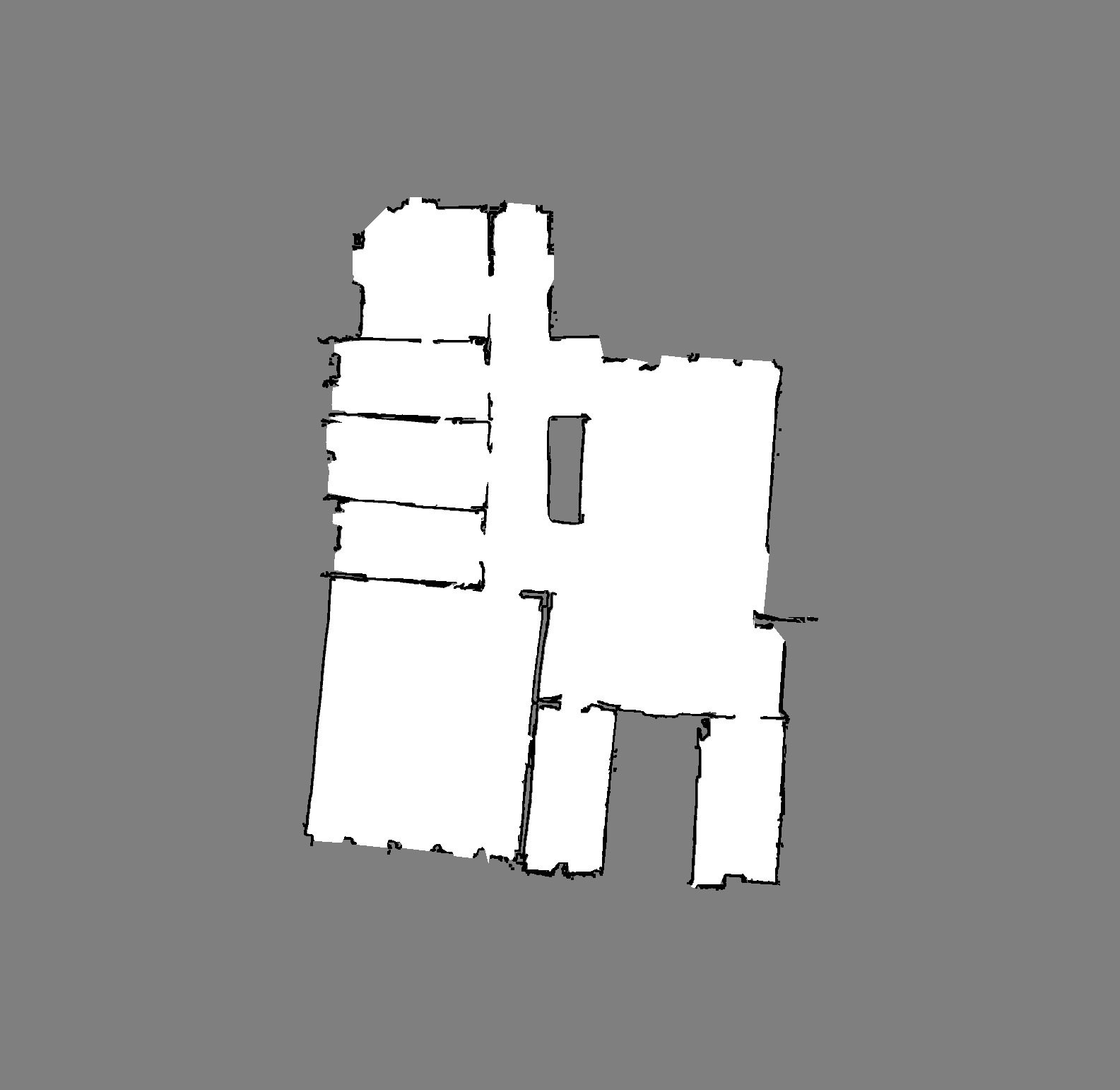}%_20170409120957.jpg}
    % \caption[]{sensor map 13}%% \label{xxx}
  \end{subfigure}%
  ~%
  \begin{subfigure}{.33\linewidth}
    \includegraphics[width=\linewidth]{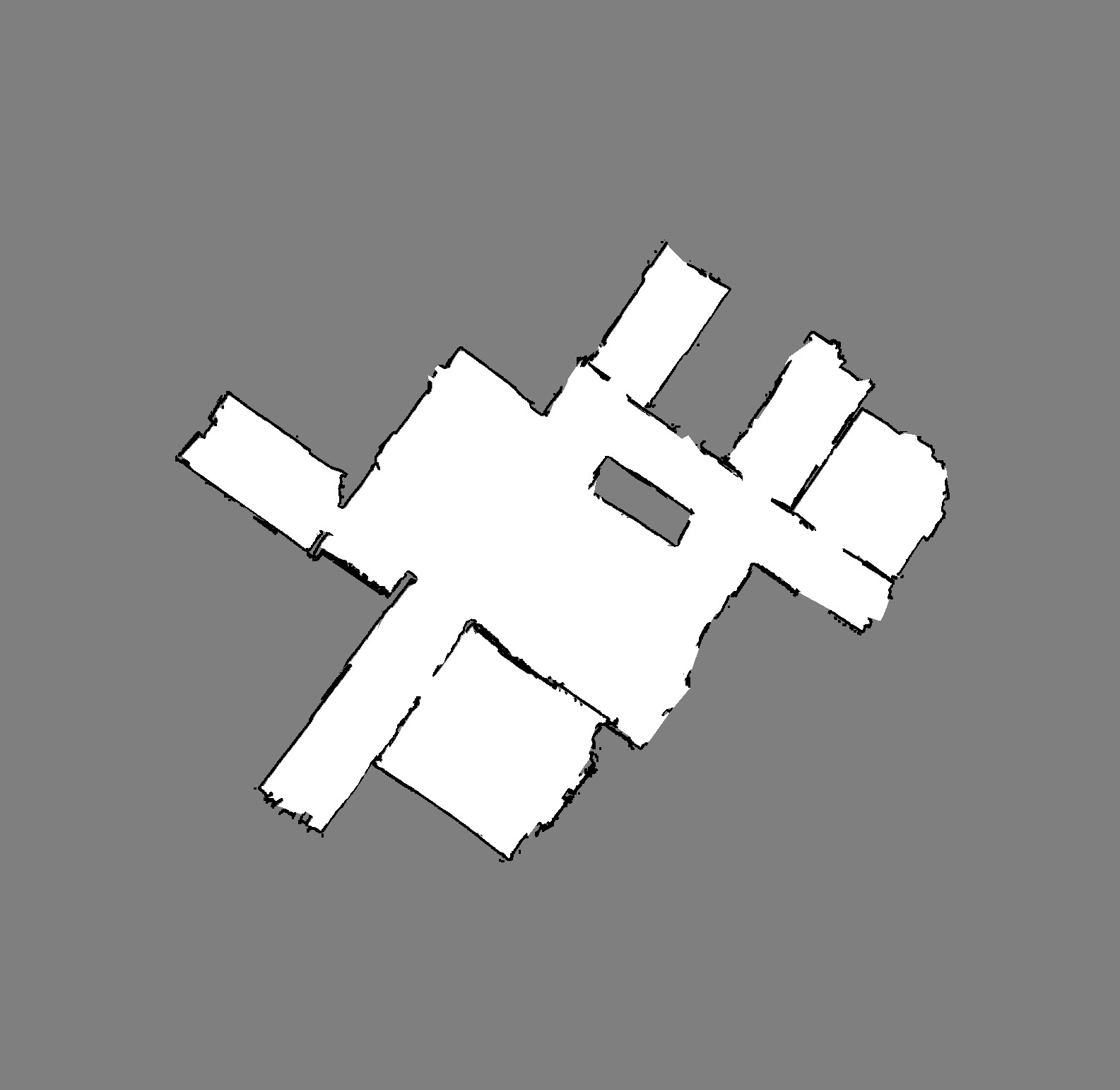}%_20170409121712.jpg}
    % \caption[]{sensor map 14}%% \label{xxx}
  \end{subfigure}%
  \caption[xxx]{HH\_F5 (office building in Halmstad, Sweden)}
  \label{fig:F5_HH}
\end{figure}

%%%%%%%%%%%%%%%%%%%%%%%%%%%%%% kpt4a
\begin{figure}%[!ht]
  \centering
  \begin{subfigure}{.33\linewidth}
    \includegraphics[width=\linewidth]{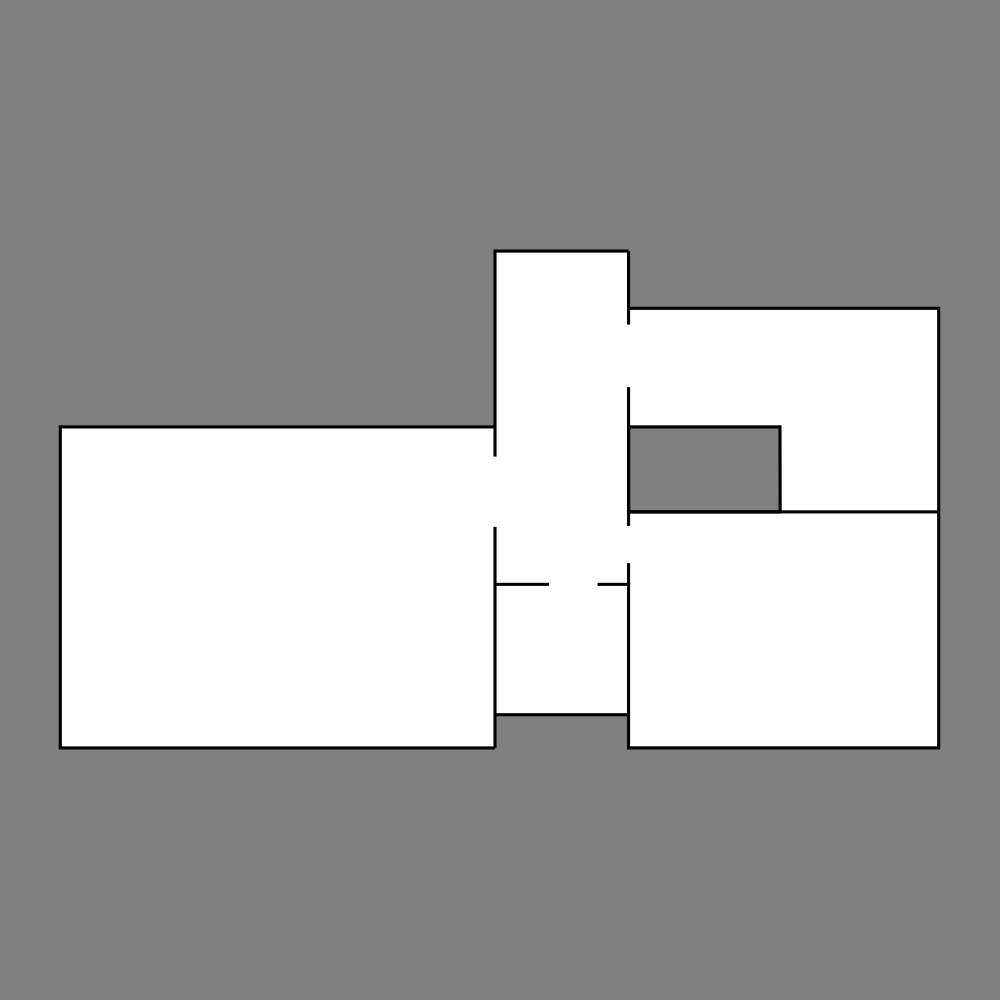}
    % \caption[]{layout}%% \label{xxx}
  \end{subfigure}%
  ~%
  \begin{subfigure}{.33\linewidth}
    \includegraphics[width=\linewidth]{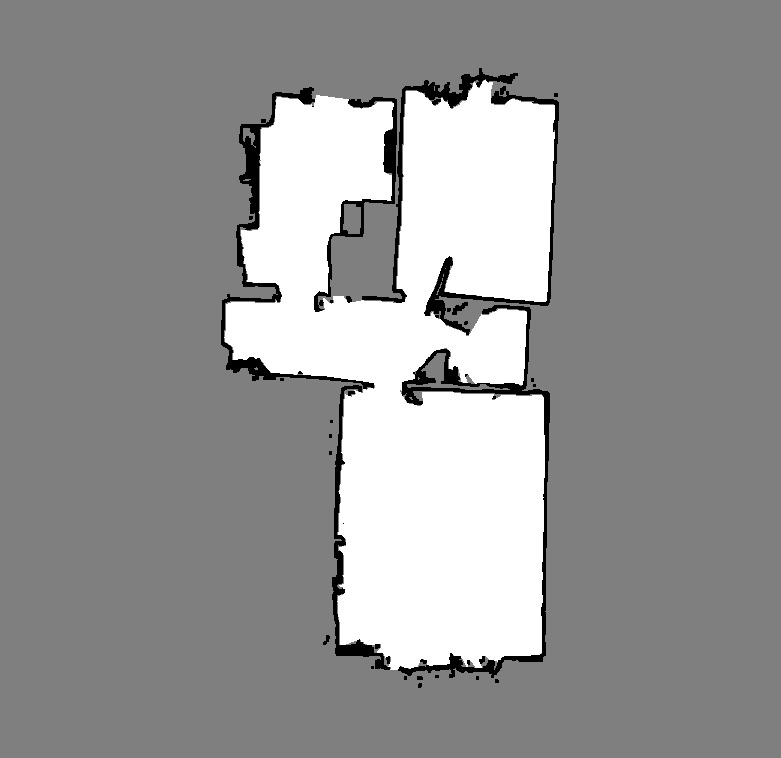}%_20170131163311.jpg}
    % \caption[]{sensor map 1}%% \label{xxx}
  \end{subfigure}%
  ~%
  \begin{subfigure}{.33\linewidth}
    \includegraphics[width=\linewidth]{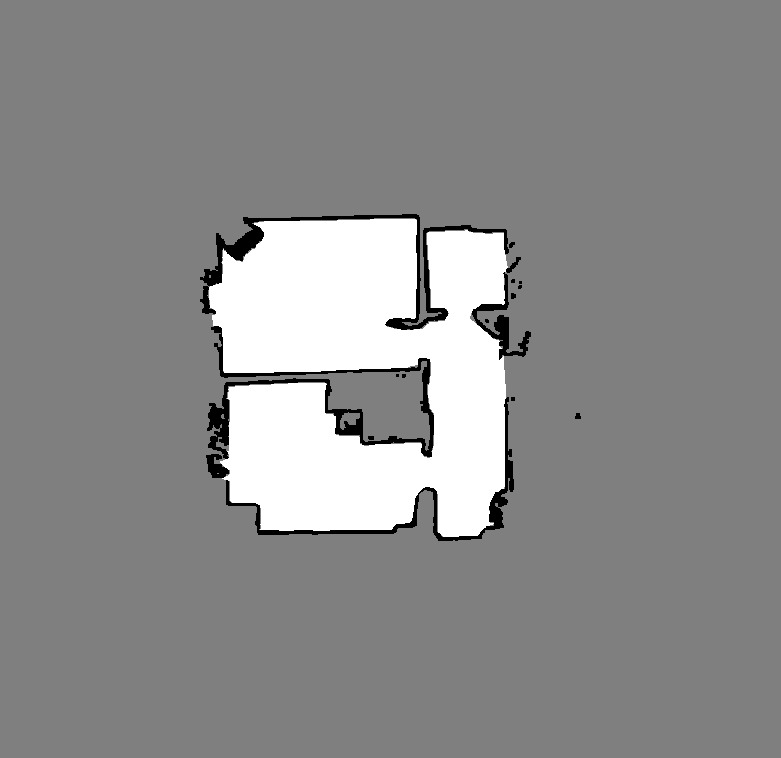}%_20170131163634.jpg}
    % \caption[]{sensor map 2}%% \label{xxx}
  \end{subfigure}%

  \begin{subfigure}{.33\linewidth}
    \includegraphics[width=\linewidth]{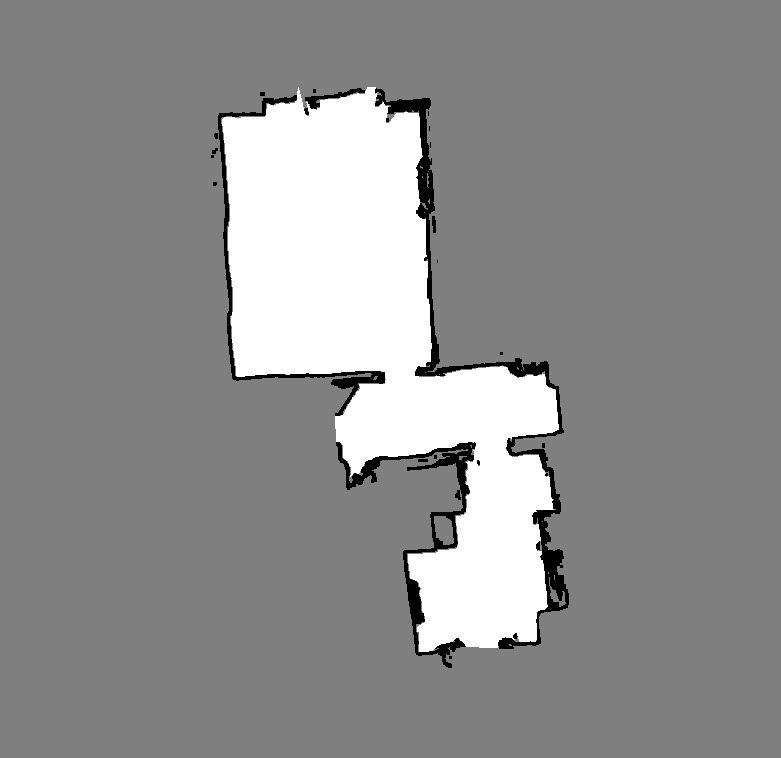}%_20170131162628.jpg}
    % \caption[]{sensor map 3}%% \label{xxx}
  \end{subfigure}%
  ~%
  \begin{subfigure}{.33\linewidth}
    \includegraphics[width=\linewidth]{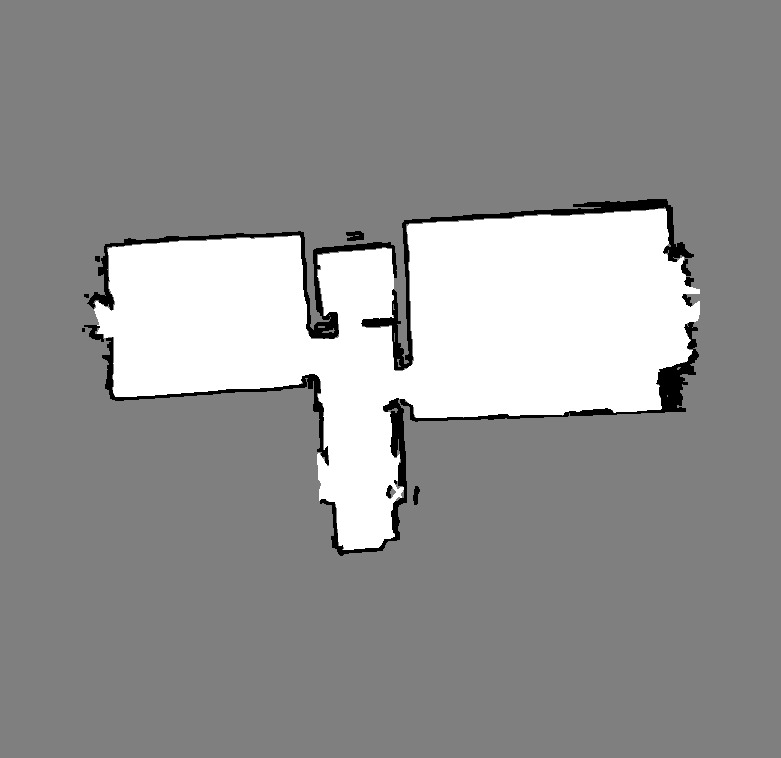}%_20170131164048.jpg}
    % \caption[]{sensor map 4}%% \label{xxx}
  \end{subfigure}
  \caption[xxx]{KPT (apartment in Halmstad, Sweden)}
  \label{fig:KPT}
\end{figure}

\end{document}